\documentclass{article}

\usepackage[main, final, nonatbib]{neurips_2026}

\usepackage[utf8]{inputenc} 
\usepackage[T1]{fontenc}    
\usepackage{hyperref}       
\usepackage{url}            
\usepackage{booktabs}       
\usepackage{amsfonts}       
\usepackage{nicefrac}       
\usepackage{microtype}      
\usepackage{xcolor}         
\usepackage{algorithm}
\usepackage{algpseudocode}

\usepackage{titletoc}       
\titlecontents{part}
  [0em]
  {\vspace{0.8em}\large\bfseries}
  {}
  {}
  {}

\titlecontents{section}
  [1.5em]
  {\small}
  {\contentslabel{1.8em}}
  {}
  {\hfill\contentspage}

\titlecontents{subsection}
  [4.0em]
  {\small\color{black!70}}
  {\contentslabel{2.5em}}
  {}
  {\hfill\contentspage}

\usepackage[sortcites=true,sorting=nyt,backend=biber,natbib=true, giveninits=true]{biblatex}
\usepackage{graphicx} 

\usepackage{amsmath}
\usepackage{bm}
\usepackage{amssymb}
\usepackage{amsfonts}
\usepackage{amsthm}
\usepackage{physics}
\usepackage{todonotes}
\usepackage{mathtools}
\usepackage[labelsep=space]{subcaption}

\usepackage{hyperref}
\usepackage{orcidlink}
\usepackage{cleveref}
\usepackage{multicol}
\usepackage{xcolor}
\usepackage{comment}
\usepackage{soul}
\usepackage{float}
\usepackage{fontawesome}
\usepackage{enumitem}
\newtheorem{result}{Result}

\newcommand{\Expected}[2]{\mathbb{E}_{#2}\left[#1\right]}
\newcommand{\pdf}[3]{\mathcal{P}_{#2}^{#3}\left(#1\right)}
\newcommand{\loss}[2]{\mathcal{E}_{#2}\left(#1\right)}

\newcommand{\Real}{\mathbb{R}}
\newcommand{\Complex}{\mathbb{C}}

\title{
    The Interplay of Data Structure and Imbalance in the Learning Dynamics of Diffusion Models
}

\author{%
    Flavio Nicoletti \\
    DSAI \\
    Chalmers and University of Gothenburg \\
    \texttt{flavion@chalmers.se} \\
    \And
    Chenxiao Ma \\
    DSAI \\
    Chalmers and University of Gothenburg \\
    \texttt{mache@chalmers.se} \\
    \And
    Enrico Ventura \\
    International School of Advanced Studies (SISSA) \\
    \texttt{eventura@sissa.it} \\
    \And
    Luca Saglietti \\
    Department of Computing Sciences \\
    Bocconi University\\
    \texttt{luca.saglietti@unibocconi.it}\\
    \And
    Stefano Sarao Mannelli \\
    DSAI \\
    Chalmers and University of Gothenburg \\
    \texttt{s.saraomannelli@chalmers.se}
}

\usepackage{caption}
\usepackage{subcaption}
\allowdisplaybreaks

\begin{document}

\maketitle

\begin{abstract}
Real-world datasets are inherently heterogeneous, yet how per-class structural differences and sampling imbalance shape the training dynamics of diffusion models—and potentially exacerbate disparities—remains poorly understood. While models typically transition from an initial phase of generalization to memorizing the training set, existing theory assumes homogeneous data, leaving open how class imbalance and heterogeneity reshape these dynamics. In this work, we develop a high-dimensional analytical framework to study class-dependent learning in score-based diffusion models. Analyzing a random-features model trained on Gaussian mixtures, we derive the feature-covariance spectrum to characterize per-class generalization and memorization times. We reveal the explicit hierarchy governing these dynamics: class variance is the primary determinant of learning order—consistently favoring higher-variance classes—while centroid geometry and sampling imbalance play secondary roles. Specifically, sampling imbalance acts as a modulator that can reverse this ordering and, under strong imbalance, forces minority classes to acquire distinct, delayed inference times during backward diffusion. Together, these results suggest that diffusion models can memorize some classes while others remain insufficiently learned. We validate our theoretical predictions empirically using U-Net models trained on Fashion MNIST \href{https://github.com/FlavioNicoletti/Unbalanced_diffusion/tree/main}{\faGithub}.
\end{abstract}

\section{Introduction}
\label{sec:Introduction}

In generative models, unequal learning across classes or sub-populations translates directly into unequal quality or memorization in the generated samples. Recent empirical work has shown that diffusion models can exacerbate such disparities even on datasets that appear balanced at the class level \cite{perera2023analyzing, vice2025exploring, shi2025dissecting}. The mechanisms behind these effects, however, remain poorly understood.

A growing body of theory has begun to characterize the training dynamics of score-based diffusion models. On finite datasets, sufficiently expressive models trained by score matching are naturally driven toward exact reproduction of the training samples, making memorization the asymptotic outcome of training \cite{biroli2024dynamical}. 
The key dynamical phenomenon, however, is that this regime is not necessarily reached immediately: diffusion models can first pass through an initial phase of generalization before eventually memorizing the training set \cite{kadkhodaie_generalization_2023, Biroli_2023, ambrogioni2024thermo, achilli2025memorization}. In overparameterized models, this separation between generalization and memorization is what allows early stopping to act as an implicit regularizer \cite{bonnaire2025diffusion, favero2025bigger, george2025analysis}: there is a window during which the model has learned the distribution but has not yet memorized the training set.

Existing analyses of this timescale separation, however, largely characterize learning at the level of the full dataset, implicitly assuming that all sub-populations are structurally identical.
Real datasets are not homogeneous: classes can differ substantially in intrinsic variance, geometric organization, and sampling frequency \cite{wang2025analytical, merger2025generalization, perera2023analyzing, vice2025exploring}. Under these conditions, there is no reason to expect a single global schedule for learning or memorization. Different sub-populations may be learned at different stages of training, so that a stopping time that still regularizes one class may already leave another overfit or insufficiently learned.

In this work, we develop a high-dimensional theoretical framework for this class-aware setting by studying denoising score matching on Gaussian-mixture data with a random-features model. Our research questions are: \textit{In what order are different classes generalized and memorized during training?} \textit{How do variance, centroid geometry, and sampling imbalance shape that ordering?}

The main contributions of this work are the following:
\begin{itemize}
    \item We characterize the effect of class imbalance on backward diffusion via class-specific inference times (see Result \ref{res:speciation_times}). While in the weak-imbalance regime, all classes remain inferable at the speciation time for exact models \cite{biroli2024dynamical, achilli2026theory}, under strong imbalance, minority classes become inferable at distinct lower diffusion times.
    \item We identify the hierarchy governing class-specific learning dynamics. Class variance is the primary determinant of the order in which classes are generalized and memorized, consistently favoring higher-variance classes, while centroid geometry and sampling ratio play secondary roles. Classes with smaller-norm centroids are favored (see Result \ref{res:learning_hierarchy}) and sampling imbalance acts as a modulating factor that can reverse this ordering in certain regimes (see Result \ref{res:tradeoff}).
    \item We find strong qualitative agreement between U-Net diffusion models trained on Fashion MNIST and Random Feature models trained on a Gaussian mixture parametrized by class-wise means and variances estimated from the data (see Result \ref{res:exp_consist}). This agreement supports, on the one hand, the effectiveness of Gaussian score approximations for diffusion models \cite{wang_vastola24}, and on the other, our theoretical predictions on the roles of variance and sampling imbalance in shaping learning dynamics.   
\end{itemize}

A detailed discussion positioning this work within the related literature is deferred to Appendix~\ref{sec:Related_Works}.

\section{Problem Formulation}
\label{sec:Problem_Formulation}

Score-based diffusion models transport a target data distribution $\mathcal{P}_0(\bm{x})$ to a standard $N$-dimensional normal distribution $\mathcal{N}(\bm{x}; \mathbf{0}, \mathbf{I}_N)$ through a forward process. Data generation is then achieved by reversing this trajectory \cite{song_score-based_2021, anderson1982reverse}. The forward and backward stochastic differential equations are respectively defined as:\\
\begin{minipage}{0.42\textwidth}
    \begin{equation}
    \label{eq:sde_forward}
        d\bm{x}(t) = -\bm{x} dt+\sqrt{2}\,d\bm{B}(t)
    \end{equation}
\end{minipage}%
\hfill
\begin{minipage}{0.54\textwidth}
    \begin{equation}
    \label{eq:sde_backward}
        d\bm{x}(t) = -(\bm{x}+2\mathbf{s}(\bm{x}, t)) dt+\sqrt{2}\,d\bm{B}(t)
    \end{equation}
\end{minipage}

\noindent 
where $\bm{B}(t)$ is the standard Wiener process and $\mathbf{s}(\bm{x}, t)=\nabla_x \log \mathcal{P}_t(\bm{x})$ denotes the score function. The forward process induces a Gaussian transition kernel $\mathcal{P}_t(\bm{x}|\bm{x}_0) = \mathcal{N}(\bm{x}; \bm{x}_0 e^{-t}, \Delta_t \mathbf{I}_N)$, where $\Delta_t = 1 - e^{-2t}$ is the noise variance at time $t$. 

In practice, the exact score is inaccessible and is approximated by a model $\bm{s}(\bm{x}, t | \bm{\Theta})$—in our case, a Random Feature (RF) model \cite{rahimi2007random, george2025denoisingscorematchingrandom}. The parameters $\bm{\Theta}$ are learned by minimizing the score matching risk \cite{ho2020denoising, song_score-based_2021} over a dataset $\bm{X} = \{\bm{x}_\nu\}_{\nu=1}^M$:
\begin{equation}
    \label{eq:empirical_risk_score_denoising}
    \loss{\bm{X}, t|\bm{\Theta}}{train}=\frac{1}{N M}\sum_{\nu=1}^M \Expected{\left\lVert \sqrt{\Delta_t}\,\bm{s}(\bm{x}_\nu(t), t|\bm{\Theta})+\bm{\xi}\right\lVert^2}{\bm{\xi}\sim \mathcal{N}(\mathbf{0}, \mathbf{I}_N)},
\end{equation}
where $\bm{x}_\nu(t) = \bm{x}_\nu e^{-t} + \sqrt{\Delta_t}\bm{\xi}$ represents the noisy samples at diffusion time $t$. The performance is evaluated on fresh samples $\bm{x}\sim \mathcal{P}_0$ through the test error (below $\bm{x}(t) = \bm{x}e^{-t}+\sqrt{\Delta_t}\bm{\xi}$)
\begin{equation}
    \label{eq:test_error_score_denoising}
    \loss{\mathcal{P}_0, t|\bm{\Theta}}{test}=
    \Expected{\left\lVert \sqrt{\Delta_t}\,\bm{s}(\bm{x}(t), t|\bm{\Theta})+\bm{\xi}\right\lVert^2}{\bm{x}\sim \mathcal{P}_0,\bm{\xi}\sim \mathcal{N}(\mathbf{0}, \mathbf{I}_N)},
\end{equation}
or equivalently, through the score MSE in Appendix~\ref{sec:Exact_score_and_score_MSE}.

\paragraph{Score Parameterization.} We model the score function using a RF model with $P$ hidden neurons:
\begin{equation}
    \bm{s}(\bm{x}|\mathbf{A}, \mathbf{W}) =
    \mathbf{A}\,\phi\left(\frac{1}{\sqrt{N}}\mathbf{W}\bm{x}\right),
\end{equation}
where $\mathbf{W} \in \Real^{P\times N}$ is a fixed random projection matrix with i.i.d. standard normal entries $W_{ip} \sim \mathcal{N}(0, 1)$, and $\mathbf{A} \in \Real^{N\times P}$ is the learnable weight matrix.
The non-linear activation function is $\phi(\cdot)=\tanh$ and it is applied element-wise to the pre-activations.
The theoretical framework we develop in the following applies to any activation function that admits a Hermite expansion. In Appendix~\ref{sec:role_of_activation}, we explore the role of the activation function.

\paragraph{Data Distribution.} We assume the training dataset $\bm{X}=\{\bm{x}_{\nu}\}_{\nu=1}^M$ is sampled from a Gaussian Mixture Model (GMM) with $C$ components:
\begin{equation}
\label{eq:pdf_data}
    \pdf{\bm{x}}{0}{} = \sum_{c=1}^C b_c\,\mathcal{N}(\bm{x}; \bm{m}_c, \sigma_c^2 \mathbf{I}_N),
\end{equation}
where $\mathbf{m}_c$ are class centroids, $\sigma_c^2$ are class-conditional variances, and $b_c \geq 0$ are mixing proportions such that $\sum_{c=1}^C b_c = 1$.
Unless stated otherwise, we assume $\forall\ c,\ ||\bm{m}_c||=\mathcal O(\sqrt N)$, $\forall\ c\neq c',\ ||\bm{m}_c-\bm{m}_{c'}||=\mathcal O(\sqrt N)$ and $\sigma_c^2 \simeq \frac{||\bm{m}_c||^2}{N}=\mathcal{O}(1)$.

\paragraph{Training Dynamics.} 
Under Mean Squared Error minimization, the training dynamics of RF models are exactly solvable. 
Indeed, the gradient flow equation under the training loss \eqref{eq:empirical_risk_score_denoising} is a linear ODE for the readout weights $\mathbf{A}$. Assuming an initialization of $\mathbf{A}(0)=\mathbf{0}$, the exact solution is given by:
\begin{equation}
\label{eq:dynamics_readouts_solution}
    \mathbf{A}(\tau)=-\frac{1}{\sqrt{\Delta_t}}\mathbf{V}^\intercal \mathbf{U}^{-1} \left(\mathbf{I}_P-e^{-2\mathbf{U}\,\tau}\right);
\end{equation}
where $\tau$ stands for the training time and $\mathbf{U}$ and $\mathbf{V}$ denote the feature-feature and feature-noise empirical covariance matrices:
\begin{align}
    \label{eq:U_def}
    &\mathbf{U}=\sum_{c=1}^C b_c \mathbf{U}_c, 
    \qquad 
    \mathbf{U}_c=\frac{1}{M_c}\sum_{\nu\in I_c}\Expected{
    \phi\left(\frac{1}{\sqrt{N}}\mathbf{W}\bm{x}_{\nu}(t)\right)
    \phi\left(\frac{1}{\sqrt{N}}\mathbf{W}\bm{x}_{\nu}(t)\right)^\intercal}{
    \bm{\xi}
    },
    \\
    \label{eq:V_def}
    &\mathbf{V}=\sum_{c=1}^C b_c \mathbf{V}_c,
    \qquad
    \mathbf{V}_c=\frac{1}{M_c}\sum_{\nu\in I_c}\Expected{\phi\left(\frac{1}{\sqrt{N}}\mathbf{W}\bm{x}_{\nu}(t)\right)\bm{\xi}^\intercal}{
    \bm{\xi}
    },
\end{align}
and $I_{c}$ is the index set for samples of class $c$. 
The full derivation of the solution of the dynamics is provided in Appendix~\ref{sec:dynamics_solution_derivation}. 
 
\paragraph{High-Dimensional Asymptotics and Gaussian Equivalence.} The learning dynamics are governed by the eigenspectrum of $\mathbf{U}$, which dictates the convergence timescales, while the cross-covariance $\mathbf{V}$ determines the scale of the learned weights. Following \cite{bonnaire2025diffusion}, we analyze these training regimes using Random Matrix Theory to characterize the spectrum of $\mathbf{U}$ in the high-dimensional proportional limit where $N, P, M \rightarrow \infty$ with fixed $\mathcal{O}(1)$ ratios $\chi_{p} = \frac{P}{N}$ and $\chi_{m} = \frac{M}{N}$.


We leverage Gaussian Equivalence (GE) techniques for RF models in a mixture setting \cite{goldt2020modeling,gerace2020generalization,dandi2023universality}, adopting a class-conditional viewpoint (see (Appendix~\ref{sec:GEP_non_centered})).
GE allows to represent a $\Real^{P\times P}$ correlation matrix of non-linear features with a correlation matrix of linear features, perturbed by noise, in the limit $P\rightarrow\infty$, under the following two hypothesis: i) the data distribution has sub-gaussian tails; ii) the operator norm of the data covariance matrix is upper bounded. Although in our work we only consider gaussian components for the mixture, our results can be straightforwardly generalized to any mixture whose components satisfy the hypothesis of GE.
In the high-dimensional limit, the empirical covariance $\mathbf{U}$ admits the asymptotically equivalent formulation $\mathbf{U} \simeq \sum_{c=1}^C b_c \mathbf{U}_{c}^{(gep)}$, where:
\begin{equation} \label{eq:U_GEP}
    \mathbf{U}_{c}^{(\mathrm{gep})} = \frac{1}{M_c}\sum_{\nu\in I_c}\mathbf{G}_t^{(c, \nu)}{\mathbf{G}_t^{(c, \nu)}}^\intercal + \frac{\Delta_t}{N \sigma_c^2 e^{-2t}}\widehat{\mathbf{W}}_{t;c}\widehat{\mathbf{W}}_{t;c}^\intercal + \operatorname{diag}\left(\bm{d}_t^{(c)}\right).
\end{equation}
Here, $\bm{G}_t^{(c, \nu)}$ are linear functions of the samples pre-activations $\bm{u}_\nu=\frac{1}{\sqrt{N}}\mathbf{W}\bm{x}_{\nu}$ with additional
Gaussian noise, $\widehat{\mathbf{W}}_{t;c}\in \Real^{P\times N}$ are cluster-specific RF matrices, and $\operatorname{diag}(\bm{d}_t^{(c)})$ is a diagonal matrix containing the vector $\bm{d}_t^{(c)}\in \Real^P$.
The first term of Eq.~\eqref{eq:U_GEP} is a linearization of Eq.~\eqref{eq:U_def} containing information on the training set and is responsible for the onset of memorization and diffusion-independent generalization; 
the second and third terms, instead, are dataset-independent and represent the $t$-dependent part of generalization. We explain how to derive the GEP in \eqref{eq:U_GEP} and how to single out exactly generalization and memorization terms in Eq.~\eqref{eq:U_GEP} in Appendix~\ref{sec:GEP_feature_correlation_matrices}.

\paragraph{Learning curves.}

Under Eq.~\eqref{eq:dynamics_readouts_solution}, the train and test errors of Eqs.\eqref{eq:empirical_risk_score_denoising},\eqref{eq:test_error_score_denoising} can be written in terms of traces of the empirical correlations in Eqs.\eqref{eq:U_def},\eqref{eq:V_def} and of the population matrix $\widetilde{\mathbf{U}}=\Expected{\mathbf{U}}{\bm{X}}=\sum_{c=1}^C b_c \widetilde{\mathbf{U}}_c$
\begin{equation}
\label{eq:errors_in_terms_of_matrices}
\begin{split}
    \mathcal{E}_{train}(\tau) &= 1 - \frac{1}{N}\operatorname{Tr}\mathbf{V}^{\intercal} \rm{f}_\tau(\mathbf{U})(2\mathbf{I}_P-\mathbf{U}\rm{f}_\tau(\mathbf{U}))\mathbf{V} \\
    \mathcal{E}_{test}^{(c)}(\tau) &= 1 + \frac{1}{N}\operatorname{Tr}\mathbf{V}^{\intercal} \rm{f}_\tau(\mathbf{U})\widetilde{\mathbf{U}}_c\rm{f}_\tau(\mathbf{U})\mathbf{V} -\frac{2}{N}\operatorname{Tr}\mathbf{V}^\intercal \rm{f}_\tau(\mathbf{U})\mathbf{V}.
\end{split}
\end{equation}
where we set $\rm{f}_\tau(\mathbf{U})=\mathbf{U}^{-1}(\mathbf{I}_P-e^{-2 \mathbf{U}\tau})$ and decomposed the test error of Eq.~\eqref{eq:test_error_score_denoising} into class-conditional components $\mathcal{E}_{test}(\tau)=\sum_c b_c \mathcal{E}_{test}^{(c)}(\tau)$.
The formulas in \eqref{eq:errors_in_terms_of_matrices} are exact finite-dimensional $N, P, M<\infty$ learning curves for a given realization of the network and the training dataset encoded in the $\mathbf{U}, \mathbf{V}, \{\widetilde{\mathbf{U}}_c\}$; with these matrices replaced by their GE representations in Appendix~\ref{sec:GEP_feature_correlation_matrices}, they constitute our \emph{semi-analytical} theoretical prediction for the dynamics (more details in Appendix~\ref{sec:analytical_prediction_errors}). In the high-dimensional proportional limit described in the previous paragraph, the traces in Eq.~\eqref{eq:errors_in_terms_of_matrices} concentrate around their mean: we computed the asymptotic expression for the train error in Appendix~\ref{sec:analytical_prediction_errors}, leaving the expressions for the test errors to future work.

\section{Characterization of the Learning Dynamics}
\label{sec:characterization_learning_dynamics}

We characterize how the Random Feature model learns on an imbalanced, structured dataset by minimizing the score-matching empirical risk. We proceed in three steps: First, in Sec.\ref{sec:speciation}, we analyze the backward diffusion process to identify the class-specific inference times, formalizing when distinct data structures become inferable. Next, in Sec.\ref{sec:timescale_separation}, we detail how these structural properties are learned during training, leveraging the eigenspectrum of the feature covariance to define global timescales for generalization and memorization. Finally, in Sec.\ref{sec:class_specific_gen_mem}, we disaggregate these global phases to examine class-specific learning trajectories, demonstrating how the interplay between class variance, centroid norms, and sampling imbalance dictates the exact order in which classes are generalized and memorized.

\subsection{Class Inference in Backward Diffusion}
\label{sec:speciation}
%

\begin{figure}
    \centering
    \includegraphics[width=0.95\linewidth]{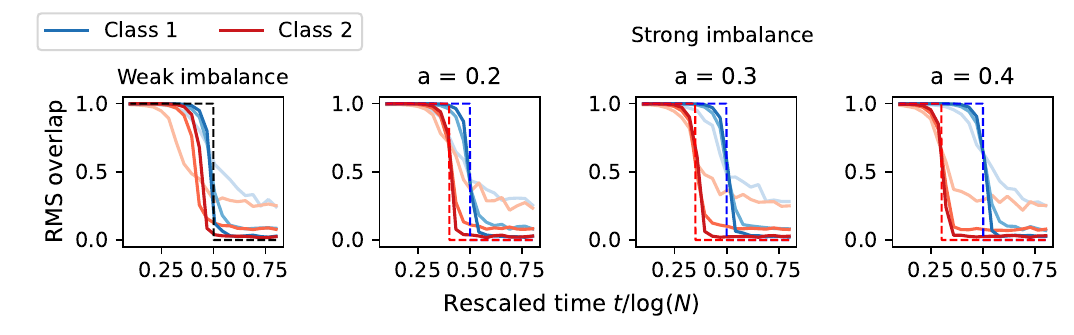}
    \caption{\textbf{Strong imbalance induces separation in inference times.} Root mean square (RMS) overlaps between the projected centroids $\bm{\mu}_c$ (Eq.~\eqref{eq:U_GEP_close_to_spec}) and the top principal components ($\bm{\psi}_{1},\bm{\psi}_{2}$) of $\mathbf{U}_{\mathrm{gep}}$, defined as $\operatorname{RMS}_c\equiv \operatorname{Mean}(\sum_{i=1}^2(\bm{\mu}_c\cdot \bm{\psi}_i)^2/\lVert \bm{\mu}_c \lVert^2)^{1/2}$, as a function of rescaled diffusion time $\widetilde{t}\equiv t/\log N$. \textit{Left} panel corresponds to weak imbalance ($b_1=0.7, b_2=0.3$, Eq.~\eqref{eq:weak_unbalance_speciation}), while the remaining panels illustrate instances of strong imbalance ($b_1=1-N^{-a}$, $b_2=N^{-a}$, Eq.~\eqref{eq:strong_unbalance_speciation}). Dashed lines denote asymptotic theoretical predictions from Eqs.\eqref{eq:weak_unbalance_speciation},\eqref{eq:strong_unbalance_speciation}; continuous lines represent finite-size numerical simulations for $N \in \{50, 500, 5000\}$ and $P \in \{100, 1000, 10000\}$, with increasing system size indicated by darker colors. Simulations show convergence toward the predicted asymptotic limits. Parameter values: $\sigma_1^2=\sigma_2^2=0.5$, $\lVert \bm{m}_1 \lVert= \lVert \bm{m}_2 \lVert = \sqrt{N}$, $\bm{m}_1\cdot \bm{m}_2=0$, $M=2000$, $N_{runs}=50$ for $P=100, 1000$ and $N_{runs}=10$ for $P=10000$.}
    \label{fig:speciation_unbalance}
\end{figure}

Prior work characterizes the \textit{speciation transition} as the diffusion time at which the generative sampling process begins to resolve specific data modes \cite{raya_spontaneous_2023, biroli2024dynamical, achilli2026theory}. Such definitions are model-agnostic, characterizing mode resolution only in terms of the geometric properties of data during backward diffusion.
In our framework, we ask the following question: \emph{when will a learning model infer different classes of a dataset during backward diffusion?}
We define the \textit{class-inference time} as the point along the backward diffusion trajectory where a class centroid becomes inferable through the learning of the score function.

To precisely characterize these times, we analyze the spectrum of the Gaussian equivalent feature covariance $\mathbf{U}_{\mathrm{gep}}$. Expanding Eq.~\eqref{eq:U_GEP} in the large-time regime (small $e^{-t}$), we obtain:
\begin{equation} \label{eq:U_GEP_close_to_spec}
    \mathbf{U}_{\mathrm{gep}} = \frac{\gamma^2}{N}\mathbf{W}\mathbf{W}^\intercal + (\widetilde{\beta}-\gamma^2)\mathbf{I}_P + \gamma^2 e^{-2t}\sum_{c=1}^C b_c \bm{\mu}_c\bm{\mu}_c^\intercal + \mathcal{O}(e^{-4t}),
\end{equation}
where $\widetilde{\beta} = \Expected{\phi^2(z)}{z\sim\mathcal{N}(0, 1)}$, $\gamma = \Expected{z\phi(z)}{z\sim\mathcal{N}(0, 1)}$, and $\bm{\mu}_c=\frac{1}{\sqrt{N}}\mathbf{W}\bm{m}_c$ are the projected class centroids. The derivation of \eqref{eq:U_GEP_close_to_spec} can be found in Appendix~\ref{sec:expansion_of_U_large_t}.

Eq.~\eqref{eq:U_GEP_close_to_spec} shows that, in the high-dimensional limit ($N \rightarrow \infty$), a class becomes inferable when its corresponding rank-1 perturbation $\bm{\mu}_c\bm{\mu}_c^\intercal$ produces an isolated eigenvalue that escapes the continuous bulk of the matrix spectrum. This spectral phase transition occurs when the perturbation norm is non-vanishing, yielding the condition:
\begin{equation} \label{eq:class_speciation_condition}
    \gamma^2 e^{-2t} \lVert \bm{\mu}_c \lVert^2 b_c = \mathcal{O}(1).
\end{equation}
Assuming the initial class centroids are extensive, i.e. $\lVert \bm{m}_c\lVert^2 = \mathrm{m}_c^2 N$, the required diffusion time to allow inference scales as $t = \mathcal{O}(\log N)$. By varying the asymptotic scaling of the sampling weights $b_c$, we can evaluate how class structure dictates the temporal emergence of features:

\begin{result}[Class-Specific Inference Times]
\label{res:speciation_times}
Let $b_c$ denote the sampling weight of class $c$, assume the validity of the GE reduction. The inference time $t_I^{(c)}$ is governed by two distinct data regimes:
\begin{itemize}[leftmargin=*, nosep]
    \item \textup{\textbf{Weak Imbalance:}} When all classes are proportionally represented ($b_c = \mathcal{O}(1)$), class imbalance has no leading-order effect on the backward dynamics. All classes are inferred simultaneously at the speciation time $t_s$:
    \begin{equation} \label{eq:weak_unbalance_speciation}
        t_I^{(c)} = \frac{1}{2}\log N + \mathcal{O}(1) \equiv t_s.
    \end{equation}
    \item \textup{\textbf{Strong Imbalance:}} When a class is severely undersampled ($b_c = \mathcal{O}(N^{-a_c})$ for $0 < a_c < 1$), its transition is delayed. The system exhibits a hierarchy of class-dependent inference times:
    \begin{equation} \label{eq:strong_unbalance_speciation}
        t_I^{(c)} = \frac{1-a_c}{2}\log N + \mathcal{O}(1)\leq t_s.
    \end{equation}
\end{itemize}
\end{result}

Result \ref{res:speciation_times} shows that while for mild data imbalance class inference corresponds asymptotically to class speciation, strong imbalance breaks this symmetry. The generative model must infer the dataset structure sequentially, prioritizing overrepresented classes and suppressing the emergence of minority classes until much later in the backward process (i.e., closer to $t=0$). Fig.\ref{fig:speciation_unbalance} validates these theoretical predictions against finite-size numerical simulations for $C=2$, demonstrating good agreement as $N, P \rightarrow \infty$.

Note that an equivalent hierarchy of inference times emerges even under balanced and weakly imbalanced sampling ($b_c = \mathcal{O}(1)$) if the classes have subleading geometric properties. Specifically, if a centroid is subextensive such that $\lVert \bm{m}_c \lVert^2 = \mathcal{O}(N^{1-a_c})$, Eq.~\eqref{eq:class_speciation_condition} trivially dictates that its inference time will be equivalently delayed by $\frac{1-a_c}{2}\log N$. 

\subsection{Timescale Separation: Generalization vs. Memorization}
\label{sec:timescale_separation}
%

\begin{figure}[htbp]
    \centering
    \begin{subfigure}[b]{0.49\textwidth}
        \centering
        \includegraphics[width=\linewidth]{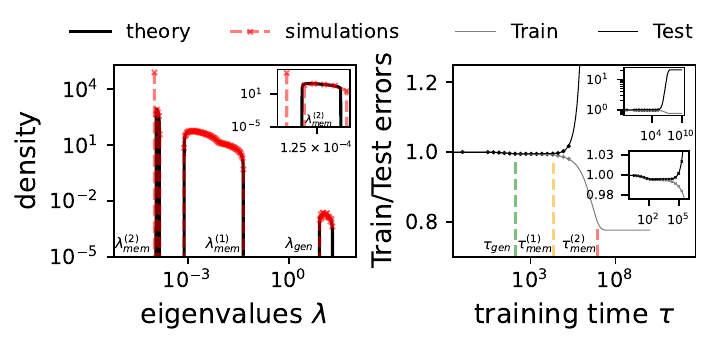}
        \caption{\textbf{Spectral density and error dynamics.} \textit{Left:} Theoretical spectral density of $\mathbf{U}$ (solid line) versus numerical histograms, highlighting the three critical eigenvalues $\lambda_{gen}$, $\lambda_{mem}^{(1)}$, and $\lambda_{mem}^{(2)}$ bounding the primary eigenvalue bulks. \textit{Right:} Train and test losses evaluated over training time $\tau$. Analytical predictions (solid lines) tightly track numerical gradient descent averages (dots). The vertical dashed lines denote the timescales corresponding to the critical eigenvalues.}
        \label{fig:timescale_first}
    \end{subfigure}
    \hfill
    \begin{subfigure}[b]{0.49\textwidth}
        \centering
        \includegraphics[width=\linewidth]{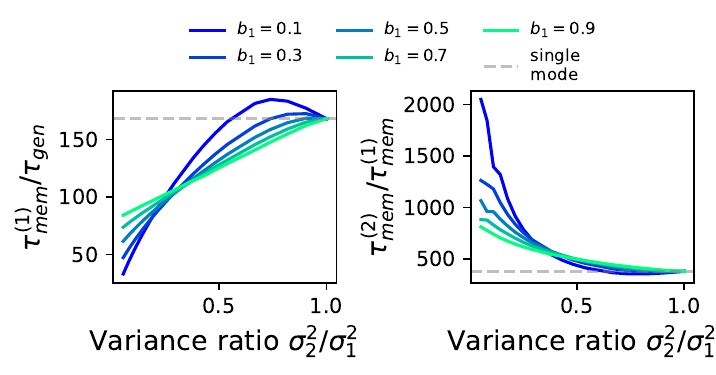}
        \caption{\textbf{Generalization and memorization windows.} Theoretical widths of the generalization window $w_g$ \textit{(Left)} and memorization window $w_m$ \textit{(Right)} as a function of the variance ratio $v = \sigma_2^2/\sigma_1^2$, plotted for varying mixing fraction $b_1$. The dashed baseline represents the standard unimodal setting. Reducing the variance ratio extends memorization but induces a non-monotonic contraction in the generalization window, an effect amplified by undersampling the high-variance class.} 
        \label{fig:timescale_second}
    \end{subfigure}
    
    \caption{\textbf{Strong class diversity can reduce the generalization window}.
    Parameters for theory: $\chi_p=60.0$, $\chi_m=30.0$, $b_1=b_2=0.5$, $\sigma_1^2=0.5, \sigma_2^2=0.25$, $\lVert \bm{m}_1 \lVert=\lVert \bm{m}_2 \lVert=0$, $t=0.001$. Parameters for finite-size simulations: $N=100$, $P=6000$, $M=3000$, $N_{epoch}=2\times 10^6$, learning rate $\eta=5\times 10^{-5} N/\Delta_t$, $N_{run}^{spectrum}=10$, $N_{run}^{train}=20$. Training time in (\subref{fig:timescale_first}) is rescaled as $\tau \rightarrow \tau/\widetilde{\eta}$, where $\widetilde{\eta}=\eta \Delta_t/N$.
    }
    \label{fig:timescale_main}
\end{figure}

Having established how data structure emerges across diffusion time ($t$), we now analyze the network's learning dynamics over training time ($\tau$). Specifically, we investigate how the eigenspectrum of the feature covariance matrix $\mathbf{U}$ dictates the global timescales of generalization and memorization during gradient descent.

For clarity, we restrict our focus to a binary Gaussian mixture ($C=2$) evaluated at small diffusion times ($t \ll 1$). We quantify class diversity via the variance ratio $v \equiv \sigma_2^2/\sigma_1^2$, treating it as a structural perturbation to the unimodal baseline ($v=1$).
Leveraging the heuristic Replica Method from Statistical Physics (detailed in Appendix~\ref{sec:derivation_spectral_equations}), we derive the spectral density of $\mathbf{U}$ in the asymptotic proportional limit described in \ref{sec:Problem_Formulation}. As shown in the left panel of Fig.\ref{fig:timescale_first}, the spectrum exhibits distinct bulk regions whose extremal eigenvalues govern distinct phases of the training trajectory. In particular, the lower edge of the rightmost bulk $\lambda_{gen}$, the upper edge of the central bulk $\lambda_{mem}^{(1)}$, and the lower edge of the leftmost bulk $\lambda_{mem}^{(2)}$, identify three interpretable timescales:
\begin{itemize}[nosep]
    \item $\tau_{gen} = (2\lambda_{gen})^{-1}$: onset of the generalization phase.
    \item $\tau_{mem}^{(1)} = (2\lambda_{mem}^{(1)})^{-1}$: end of generalization and onset of data memorization. 
    \item $\tau_{mem}^{(2)} = (2\lambda_{mem}^{(2)})^{-1}$: completion of the memorization/overfitting phase. 
\end{itemize}

In contrast to the first two timescales, the latter depends on diffusion time, with $\tau_{mem}^{(2)} \propto 1/t$ for $t \ll 1$ (see Appendix~\ref{sec:equations_for_the_edges}), indicating that the final memorization stage occurs only at asymptotically long training times.
Based on a simple counting argument on the eigenvalue density (see Appendix~\ref{sec:equations_for_the_edges}), the training phase corresponding to the central bulk can be associated with the creation of basins of attraction around each training sample, with widths set by the forward noise variance $\Delta_t \approx 2t$.
The remaining training phase, therefore, corresponds to small refinements within these narrow basins. Although this final regime is theoretically relevant to characterize the full memorization window, it is unlikely to be observed empirically: finite learning rates effectively bypass the associated small-eigenvalue directions, and nearest-neighbor metrics will typically deem the data fully memorized before $\tau_{mem}^{(2)}$ is reached.

As illustrated in Fig.~\ref{fig:timescale_first}, the theory characterizes the spectrum and corresponding spectral edges map to the macroscopic transitions in the empirical train and test losses. To evaluate the persistence of these learning phases, we define the generalization window $w_g = \tau_{mem}^{(1)}/\tau_{gen}$ and the memorization window $w_m = \tau_{mem}^{(2)}/\tau_{mem}^{(1)}$.
Fig.~\ref{fig:timescale_second} plots these windows as a function of the variance ratio $v$. We observe that strong class diversity (i.e. $v$ deviating significantly from $1$) reliably contracts the generalization window relative to the unimodal baseline, while strictly expanding the memorization window. 
In regimes of high class diversity ($v\ll 1$), imbalance modulates the width of $w_g$: imbalance toward the high-variance class mitigates the structural effects of class diversity, whereas imbalance toward the low-variance class amplifies the contraction.

\subsubsection{Class-Specific Learning Dynamics}
\label{sec:class_specific_gen_mem}
%

\begin{figure}[t]
    \centering
    \includegraphics[width=0.75\linewidth]{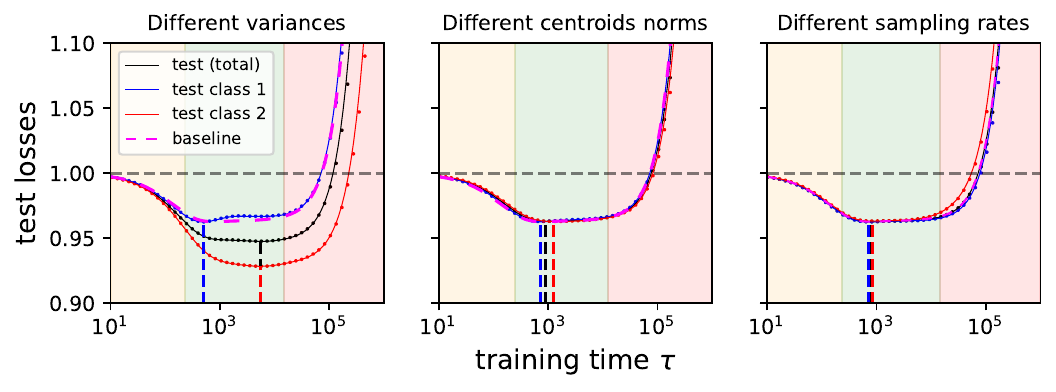}
    \caption{\textbf{Class heterogeneity induces class-specific timescales.} Class-conditional test errors (Eq.~\ref{eq:errors_in_terms_of_matrices}) evaluated over training time across three structural ablations relative to a perfectly symmetric baseline ($\sigma_1^2=\sigma_2^2=0.5$, $\lVert \bm{m}_1 \lVert = \lVert \bm{m}_2 \lVert = \sqrt{N}$, $b_1=b_2=0.5$, $\bm{m}_1\cdot \bm{m}_2=0$). The perturbations are: variance disparity ($\sigma_1^2=0.5, \sigma_2^2=0.25$; \textit{left}), centroid norm disparity ($\lVert \bm{m}_1 \lVert=\sqrt{N}, \lVert \bm{m}_2 \lVert=\frac{3}{2}\sqrt{N}$; \textit{center}), and sampling imbalance ($b_1=0.8, b_2=0.2$; \textit{right}). Solid lines denote theoretical predictions (Appendix~\ref{sec:analytical_prediction_errors}), dots represent empirical averages from gradient descent simulations and vertical dashed lines refer to the positions of the $\tau_{gen}^{(c)},\;\;\;c=1,2$. Background colors characterize the training windows theoretically derived from the spectrum of $\mathbf{U}_{\mathrm{gep}}$: early (orange), generalization (green), and memorization (red).
    Other parameters: $\chi_p=60.0$, $\chi_m=30.0$, $t=0.01$, $N=100$, $N_{epoch}=2\times 10^6$, learning rate $\eta=5\times 10^{-5} N/\Delta_t$, $N_{run}^{train}=50$.
    }
    \label{fig:test_error_ablations}
\end{figure}

To quantify the performance on each class in the dataset, we 
consider the semi-analytic class-specific test errors in Eq.~\eqref{eq:errors_in_terms_of_matrices}.
By tracking the class-specific loss trajectories, we formalize two critical times in the training process. 
The optimal generalization time, $\tau_g^{(c)}$, is defined as the training time that minimizes the class-conditional test error in Eq.~\eqref{eq:errors_in_terms_of_matrices}. 
Following this initial generalization phase, the empirical risk minimization on a finite dataset inevitably induces overfitting. We denote the onset of pure memorization, $\tau_m^{(c)} > \tau_g^{(c)}$, as the threshold where this overfitting causes the test error to degrade back to 1.\footnote{Because the readout weights are initialized at zero ($\mathbf{A}(0)=\mathbf{0}$, see Sec.\ref{sec:Problem_Formulation}), the network outputs zero at initialization. Therefore, the initial test error is trivially equal to the variance of the standard Gaussian noise: $\mathcal{E}_{test}^{(c)}(0) = 1$.} Formally:
\begin{equation} \label{eq:class_specific_gen_mem_time}
    \tau_{g}^{(c)} = \operatorname*{argmin}_{\tau}\, \mathcal{E}_{test}^{(c)}(\tau), \qquad \mathcal{E}_{test}^{(c)}(\tau_{m}^{(c)}) = 1.
\end{equation}
Fig.\ref{fig:test_error_ablations} shows the test error under three structural ablations relative to a balanced baseline with orthogonal, equal-norm centroids: doubling the variance of class 2 (left), multiplying the centroid norm of class 2 by $3/2$ (center), and skewing the sampling weights (right).
These perturbations reveal a hierarchy of influence: 
\begin{result}[Hierarchy of Class Learning]
\label{res:learning_hierarchy}
Suppose the per-class generalization and memorization times are measured from the semi-analytical curves of Eq.\eqref{eq:errors_in_terms_of_matrices}, under the GE reduction and the assumptions in Section~\ref{sec:Problem_Formulation}.
The learning timescales are governed by the following factors:
\begin{itemize}[nosep]
    \item \textup{\textbf{Primary: \underline{Variance}}}. High-variance classes present smoother and larger score targets, leading to significantly faster generalization and memorization compared to low-variance components.
    \item \textup{\textbf{Secondary: \underline{Centroid Geometry}} and \textbf{\underline{Sampling Imbalance}}}. Classes with smaller-norm centroids are prioritized over those with larger-norm centroids. Strong imbalance acts as a modulator capable of inverting the learning order established by variance and geometry.
\end{itemize}
\end{result}
For completeness, Appendix~\ref{sec:Exact_score_and_score_MSE} provides analogous results evaluated on the exact score-matching Mean Squared Error.


\begin{figure}[h]
    \centering
    \begin{subfigure}[b]{0.495\textwidth}
        \centering
        \includegraphics[width=\linewidth]{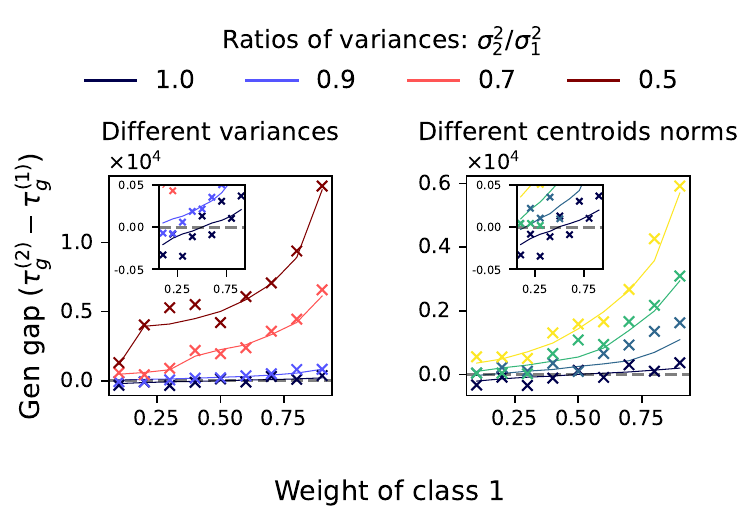}
        \caption{\textbf{Generalization gaps.} Gap between optimal generalization training time of the two classes against the sampling weight of class 1 (Eq.~\eqref{eq:class_specific_gen_mem_time}). Panels isolate the impact of variance disparity (\textit{Left}, with inset zoom) and the centroid norm disparity (\textit{Right}) introduced in Fig.~\ref{fig:test_error_ablations}. 
        }
        \label{fig:gaps_first}
    \end{subfigure}
    \hfill
    \begin{subfigure}[b]{0.495\textwidth}
        \centering
        \includegraphics[width=0.975\linewidth]{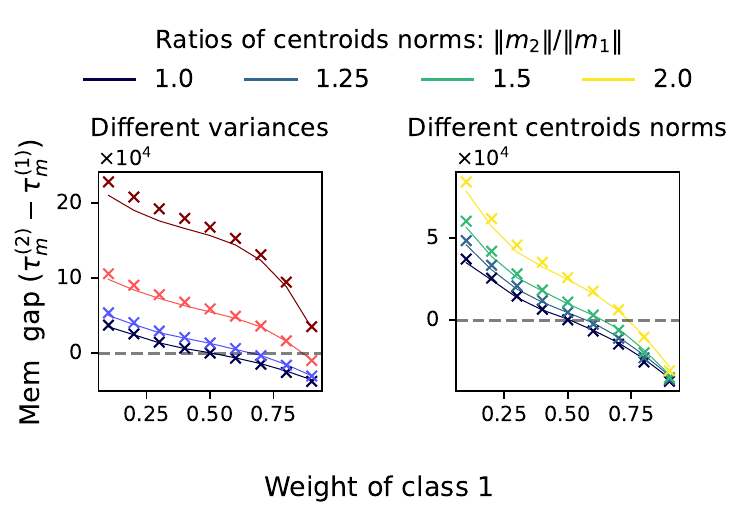}
        \caption{\textbf{Memorization gaps.} Gap between memorization times of the two classes against the sampling weight of class 1 (Eq.~\eqref{eq:class_specific_gen_mem_time}). 
        As in \subref{fig:gaps_first}, the panels show the variance disparity setting (\textit{Left}) and the centroid norm disparity setting (\textit{Right}).
        }
        \label{fig:gaps_second}
    \end{subfigure}
    
    \caption{\textbf{Imbalance can reverse the order of learning for classes}. In both subfigures, 
    solid lines represent theoretical predictions derived in Appendix~\ref{sec:analytical_prediction_errors}, while crosses represent empirical average gaps from gradient descent simulations. 
    Parameters: $\chi_p=60.0$, $\chi_m=30.0$, $t=0.01$, $N=100$, $N_{epoch}=2\times 10^6$, learning rate $\eta=5\times 10^{-5} N/\Delta_t$, $N_{run}^{train}=50$.}
    \label{fig:gaps_main}
\end{figure}

As illustrated in Fig.~\ref{fig:gaps_main}, class imbalance acts as a modulator capable of inverting the learning order observed in the balanced setting. By studying the memorization and generalization gaps between classes as functions of the mixing proportion $b_1$, we identify a fundamental trade-off:
\begin{result}[The Generalization-Memorization Trade-off]
\label{res:tradeoff}
Sample imbalance impacts generalization and memorization in opposite directions:
\begin{enumerate}[nosep]
    \item \textup{\textbf{Scarcity delays generalization:}} Undersampling the high-variance class forces the model to generalize the low-variance class first.
    \item \textup{\textbf{Scarcity accelerates memorization:}} Conversely, oversampling the high-variance class deprives the low-variance class of data, causing it to be memorized first.
\end{enumerate}
Consequently, adjusting sampling weights to close the generalization gap between classes inevitably widens the memorization gap, and vice versa.
\end{result}

\section{Numerical Experiments on Fashion MNIST}
\label{sec:Numerical_Experiments_Fashion_MNIST}

\paragraph{Model, Training, and Generation.} 
A standard DDPM \cite{ho2020denoising} was employed on the Fashion MNIST dataset \cite{xiao2017fashion}, with a U-Net \cite{ronneberger2015u} serving as the backbone. Each model is trained on a binary subset of Fashion MNIST, pairing the Sneaker class—which exhibits the lowest pixel-wise variance—with the remaining nine classes to yield nine distinct class pairs. Each pair is trained at three Sneaker proportions (25\%, 50\%, 75\%) on 8,000 images. To capture both early and late training dynamics, 32 log-spaced checkpoints are saved per run. At each checkpoint, the DDPM ancestral sampler \cite{ho2020denoising} is used to generate 500 samples per pair. Further implementation details are deferred to Appendix~\ref{sec:more_numerical_info}.

\paragraph{Evaluation.} Following \cite{yoon2023diffusion,gu2023memorization,bonnaire2025diffusion}, a generated sample $\bm{x}_c$ is considered memorized if the distance ratio to its nearest training neighbors $\mathrm{nn}_1,\,\mathrm{nn}_2$ satisfies
$\frac{\|\bm{x}_c - \mathrm{nn}_1(\bm{x}_c)\|_2}{\|\bm{x}_c - \mathrm{nn}_2(\bm{x}_c)\|_2} < \frac{1}{3}$.
The quantity $f_\mathrm{mem}^c(t)$ is the expected memorization fraction under the data distribution at diffusion time $t$. We define the threshold-crossing time $t_{\mathrm{mem}}^c = \inf\{t:f_{\mathrm{mem}}^c(t)\ge 1/3\}$, yielding the Sneaker-partner memorization gap $\Delta t_{\mathrm{mem}}(c_p)=t_{\mathrm{mem}}^{\mathrm{snk}}-t_{\mathrm{mem}}^{c_p}$. 
Samples are assigned to classes using dedicated binary ResNet-18 classifiers \cite{he2016deep} (macro F1 $>96.0\%$, balanced accuracy $>97.3\%$).

\paragraph{Results and Discussion.} 

\begin{figure}[t]
    \centering
    \includegraphics[width=\linewidth]{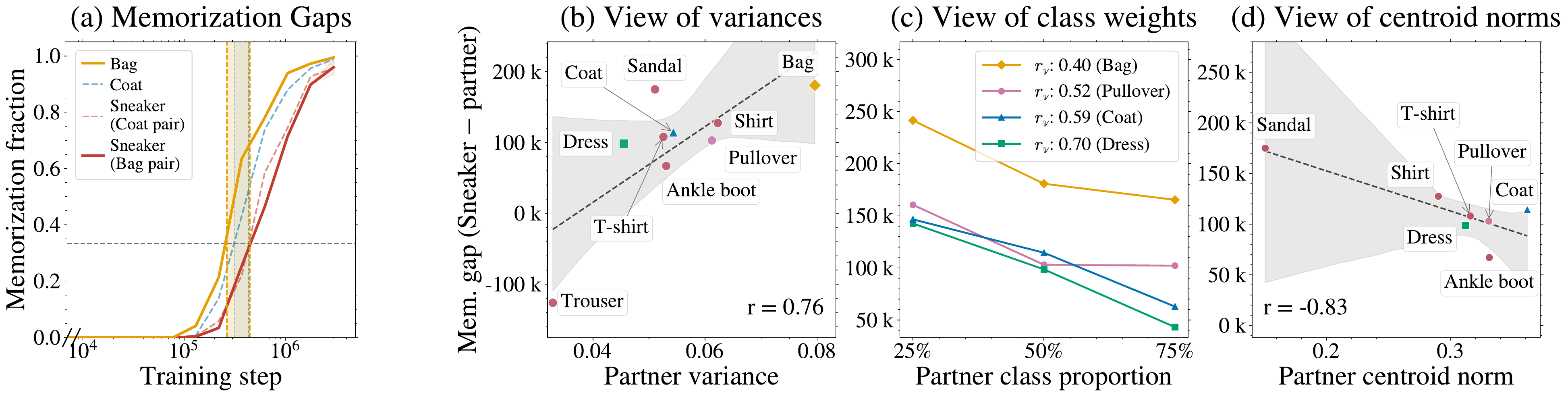}
    \caption{\textbf{DDPM memorization gaps on Fashion MNIST.}
    (\textbf{a}) \textbf{Per-class memorization fractions for Sneaker-Coat and Sneaker-Bag DDPMs.} The horizontal dashed line marks the threshold $f_{\mathrm{mem}}^c(t)=1/3$ used to define $t_{\mathrm{mem}}^c$; vertical shaded regions show the resulting Sneaker-partner time gap $\Delta t_{\mathrm{mem}}(c_p)$. Within each pair, the higher-variance partner class is memorized before Sneaker. Across pairs, the Bag-Sneaker DDPM exhibits a larger memorization gap than the Coat-Sneaker DDPM.
    (\textbf{b}) \textbf{Memorization gap versus the empirical mean pixel variance of the balanced partner class.} The dashed line is a linear fit, and the gray band is a $95\%$ bootstrap confidence interval obtained by resampling class points and refitting.
    (\textbf{c}) \textbf{Memorization gap across partner mixing proportions $b_{c_p}$.} The legend reports $r_{v}=\widehat{\sigma}_{\mathrm{snk}}^2/\widehat{\sigma}_{c_p}^2$. The selected classes are ordered in roughly 0.1 increments of $r_{v}$; apart from Bag, they have comparable empirical variances. The trend is consistent with the one in Fig.~\ref{fig:gaps_second}. 
    (\textbf{d}) \textbf{Memorization gap versus partner centroid norm.} To isolate centroid effects, we exclude relatively extreme variance cases, computed on the balanced subset; the fit and bootstrap confidence interval are computed as in (b).}
    \label{fig:Fashion MNIST}
\end{figure}

We first verify whether the structural hierarchy predicted by Result~\ref{res:learning_hierarchy} and shown in Fig.\ref{fig:test_error_ablations} is observed in Fashion MNIST DDPMs. As shown in Fig.\ref{fig:Fashion MNIST}\textbf{a}, in the Sneaker-Bag pairs, the higher-variance class attains the memorization threshold before its lower-variance counterpart. This relationship between the largest and smallest variance classes is also corroborated on CIFAR in Appendix~\ref{sec:more_figures}. Moreover, in Fig.\ref{fig:Fashion MNIST}\textbf{b}, the memorization time gap $\Delta t_{\mathrm{mem}}(c_p)$ is positively correlated with the empirical mean pixel variance of different partner classes. Additionally, imbalance acts as a modulator: Fig.\ref{fig:Fashion MNIST}\textbf{c} illustrates that increasing the partner mixing proportion $b_{c_p}$ generally reduces the memorization gap. The curves in Fig.\ref{fig:Fashion MNIST}\textbf{c} maintain a vertical ordering consistent with the variance ratio $r_v = \widehat{\sigma}_{\mathrm{snk}}^2/\widehat{\sigma}_{c_p}^2$, indicating that variance remains a dominant factor across the tested imbalance regimes. Finally, Fig.\ref{fig:Fashion MNIST}\textbf{d} isolates the effect of centroid geometry by excluding extreme-variance cases
. 
It confirms the theoretical prediction that partner classes with a lower centroid norm relative to the Sneaker class exhibit larger memorization gaps, an observation consistent with the comparatively large gap recorded for the Sneaker-Sandal pair.

\begin{figure}[t]
    \centering
    \includegraphics[width=\linewidth]{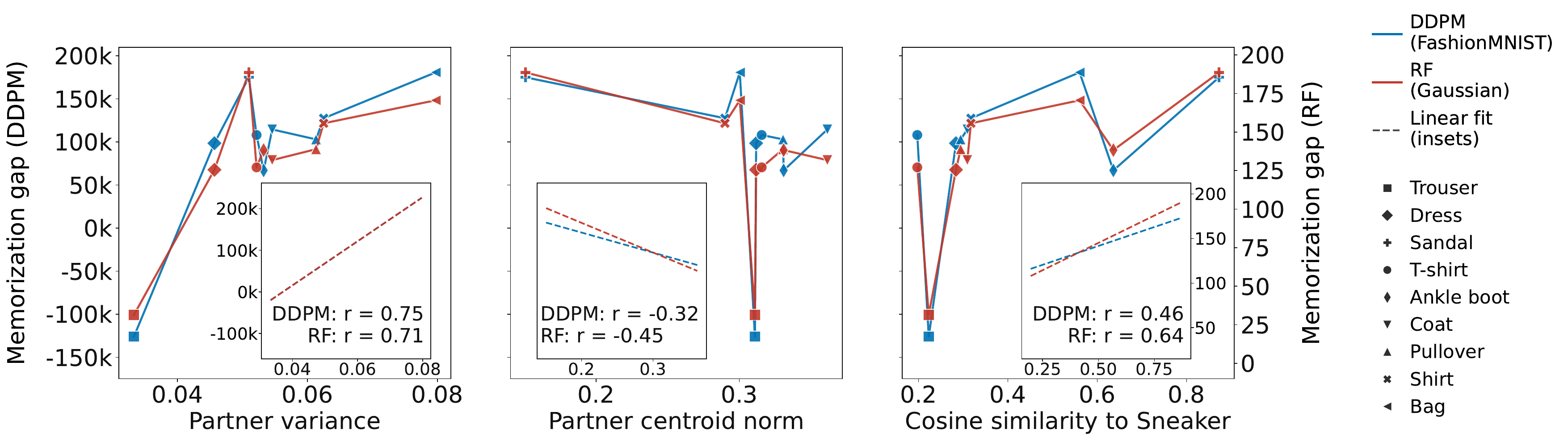}
    \caption{\textbf{DDPM--RF comparison across three descriptors.} For each Sneaker-partner pair $(c_{\mathrm{snk}},c_p)$, we compare the DDPM memorization gap on Fashion MNIST with the corresponding RF memorization gap on Gaussian synthetic data. The RF model uses pair descriptors extracted from the data: partner variance, normalized partner centroid norm, and centroid cosine similarity to Sneaker. The panels plot both gaps as functions of each descriptor; 
    connecting lines are visual guides only. Insets report linear fits and Pearson correlations in the same raw units as the outer panels.}
    \label{fig:ddpm-rf}
\end{figure}

While previous sections isolated specific data properties on synthetic data, in these numerical experiments we observe their combined effect. Additionally, each class has its own non-null overlap with the reference Sneaker class, reflected by the cosine similarity. 
To generalize the theory to this case, we 
construct a synthetic Gaussian mixture model (GMM) using the centroid norms, variances, and cosine similarities extracted from Fashion MNIST.


\begin{result}[Consistency between RF learning and Real Architectures]   
\label{res:exp_consist}
Using only these pair-level statistics, the RF model evaluated on Gaussian synthetic data qualitatively matches the memorization-gap trends of the full DDPM (Fig.\ref{fig:ddpm-rf}). Despite operating on distinct gap scales, the models share similar descriptor-dependent patterns:
\begin{itemize}[leftmargin=*, nosep]
    \item \textup{\textbf{Qualitative agreement:}} Both models exhibit larger gaps for partner classes with larger variances and smaller centroid norms, matching the theoretical predictions. This similarity emerges despite comparing distinct architectures (RF vs. U-Net) and datasets (synthetic GMM vs. Fashion MNIST).
    \item \textup{\textbf{Pairwise structure:}} The RF model captures specific geometric relationships precisely. For instance, when the pairs have similar variances and cosine similarities, the one with a smaller centroid norm results in a larger memorization gap in both models (e.g., see the Shirt pair compared to the Pullover pair). Likewise, a partner class exhibiting moderate variance and a very small centroid norm results in a large memorization gap in both settings (e.g., see the Sandal pair).
\end{itemize}
\end{result}

These results indicate that a Gaussian structural representation captures the leading-order effects of the memorization hierarchy in real image data. While first- and second-order moments do not fully resolve the complexity of memorization---as evidenced by residual discrepancies---they successfully bridge the high-dimensional theory with empirical practice.

\section{Conclusions}
\label{sec:Conclusions}

We developed a high-dimensional analytical framework for class-aware learning dynamics in score-based diffusion models, characterizing the interplay of data structure and sampling imbalance. We showed that strong imbalance induces distinct class-specific inference times, and that during score learning---close to sampling time---the statistical properties of the data, especially class variances, are the leading factors governing the order in which classes are learned. Centroid geometry and sampling imbalance play secondary roles, with the latter acting as a modulator that can reverse this ordering. We further found that U-Net DDPMs trained on Fashion MNIST qualitatively track these predictions when an RF model is parametrized by empirical class moments, supporting the relevance of Gaussian score approximations to real architectures. A natural practical implication of our results is that a single early-stopping time need not regularize all classes equally under heterogeneous data. Since classes with higher variance or smaller centroid norm are learned earlier, reweighting or rebalancing toward lower-signal classes may offer a way to mitigate per-class disparities. We leave a systematic study of such interventions to future work.

Our analysis is limited in two main respects. First, our presentation focused on binary Gaussian mixtures with isotropic covariances, even though the framework naturally extends to multi-class and non-isotropic settings. 
Despite the simplicity of such distributions, influent works about the theory of generative diffusion \cite{biroli2024dynamical, raya_spontaneous_2023} already relied on Gaussian distributions. Their motivation is, together with a higher interpretability of the model, also the fact that at high and intermediate noise levels diffusion models effectively behave like Gaussian mixtures \cite{wang_vastola24}. The same hypothesis seems transposable to training time, as supported by our experimental analysis in Section \ref{sec:Numerical_Experiments_Fashion_MNIST}, where we find a remarkable agreement between measures performed on U-Nets and Gaussian mixture models across learning. 
Furthermore, our analysis is fundamentally restricted to capturing first- and second-order structures. This leaves open the question of how more complex heterogeneities affect generative performance. For example, dataset outliers may act as strong attractors in diffusion dynamics \cite{jeon2024understanding}, and real data classes often have internal multimodal structure that cannot be reduced to a single Gaussian component. 
Moreover, future work could study data with power-law spectra. Such covariance eigenvalue distributions are common in natural images \cite{bialek_natural} and they are suspected to bring interesting implications for training diffusion models \cite{merger2025generalization}.  
Since weaker intra-class modes can also be suppressed by conditional sampling and guidance \cite{jin2025stage}, extending the theory to more realistic class structure and conditional settings would be an important next step.

\newpage
\section*{Acknowledgments and Disclosure of Fundings}
The authors thank Giulio Biroli, Sebastian Goldt, and Bruno Loureiro for their precious feedback. F.N., C.M., and S.S.M. were supported by the Wallenberg AI, Autonomous Systems, and Software Program (WASP). The computations were enabled by resources provided by the National Academic Infrastructure for Supercomputing in Sweden (NAISS) at Alvis and Tetralith (Chalmers Centre for Computational Science and Engineering, C3SE) partially funded by the Swedish Research Council through grant agreement no. 2022-06725, under project NAISS 2024/22-1082.
\printbibliography
\newpage
\appendix
\startcontents[appendix]

{\Huge \bfseries Appendices}

\vspace{1em}

\begingroup
\setcounter{tocdepth}{2}
\printcontents[appendix]{}{0}{}
\endgroup

\vspace{1.5em}

\section{Notation}
\label{sec:mathematical_notations}
We summarize the nontrivial notation used throughout the paper.

\begin{table}[h]
    \centering
    \renewcommand{\arraystretch}{1.25}
    \begin{tabular}{p{0.22\linewidth} p{0.70\linewidth}}
        \toprule
        \textbf{Notation} & \textbf{Meaning} \\
        \midrule
        $x^2$ vs. $x^{(2)}$
        & Superscripts without parentheses denote powers, while superscripts in parentheses index elements of a list or sequence. \\

        $\bm{x}, \mathbf{X}$
        & Bold lowercase symbols denote vectors and bold uppercase symbols denote matrices, unless stated otherwise. For example, if $\bm{x}\in\Real^N$, then $\bm{x}=(x_1,\ldots,x_N)^\intercal$. \\

        $\phi(\bm{x})$
        & For a scalar function $\phi:\Real\to\Real$, unless specified otherwise, $\phi(\bm{x})$ denotes elementwise application, i.e.,
        $\phi(\bm{x})=(\phi(x_1),\ldots,\phi(x_N))^\intercal$. \\

        $\odot$
        & Hadamard, or elementwise, product between vectors or matrices. \\

        $\bm{x}^{\odot a}$
        & Elementwise power: $\bm{x}^{\odot a}=(x_1^a,\ldots,x_N^a)^\intercal$. \\

        $\otimes$
        & Kronecker product between matrices. \\

        $\bm{x}\bm{x}^\intercal$
        & Outer product of a vector with itself. \\

        $\Expected{\cdot}{x\sim\mathcal{P}}$
        & Expectation with respect to $x\sim\mathcal{P}$. When the distribution is omitted, the expectation is taken over the standard normal distribution $\mathcal{N}(0,1)$. \\
        \bottomrule
    \end{tabular}
\end{table}

For the Hadamard product, given $\bm{x},\bm{y}\in\Real^N$,
\begin{equation}
    \label{eq:hadamard_vec}
    \bm{x}\odot\bm{y}
    =
    (x_1y_1,\ldots,x_Ny_N)^\intercal .
\end{equation}
For matrices $\mathbf{X},\mathbf{Y}\in\Real^{N\times N}$, the product is applied elementwise:
\begin{equation}
    \label{eq:hadamard_matrix}
    (\mathbf{X}\odot\mathbf{Y})_{ij}
    =
    X_{ij}Y_{ij}.
\end{equation}

For matrices $\mathbf{X},\mathbf{Y}\in\Real^{N\times N}$, the Kronecker product is
\begin{equation}
    \label{eq:kronecker_product}
    \mathbf{X}\otimes\mathbf{Y}
    =
    \begin{pmatrix}
        X_{11}\mathbf{Y} & \cdots & X_{1N}\mathbf{Y} \\
        \vdots & \ddots & \vdots \\
        X_{N1}\mathbf{Y} & \cdots & X_{NN}\mathbf{Y}
    \end{pmatrix}.
\end{equation}

For $\bm{x}\in\Real^N$, the outer product is
\begin{equation}
    \label{eq:external_vectors}
    \bm{x}\bm{x}^\intercal
    =
    \begin{pmatrix}
        x_1^2 & x_1x_2 & \cdots & x_1x_N \\
        x_2x_1 & x_2^2 & \cdots & x_2x_N \\
        \vdots & \vdots & \ddots & \vdots \\
        x_Nx_1 & x_Nx_2 & \cdots & x_N^2
    \end{pmatrix}.
\end{equation}

For notational brevity, when no confusion arises, we occasionally suppress full parameter dependence. For example, $\Delta_t\equiv 1-e^{-2t}$ is sometimes written simply as $\Delta$.

The range of indices in summation and product symbols is omitted when its extremes are clear from the context, i.e. they were specified in previous equations.

\section{Further related works}
\label{sec:Related_Works}

\textbf{Dynamical transitions in diffusion models.} The recent success of generative diffusion models has promoted a growing effort to interpret their behavior through the lens of statistical mechanics, particularly in the high-dimensional limit where mean-field techniques become informative. Early works in this direction have analyzed how diffusion models reconstruct the target data distribution through a sequence of dynamical transitions that coincide with ergodicity-breaking events in an effective diffusion potential \cite{Biroli_2023, ambrogioni2024thermo, achilli2025memorization}. \cite{biroli2024dynamical} characterizes the dynamical phases of diffusion models---namely speciation and collapse---while \cite{achilli2026theory} highlighted speciation timescale separations for centered groups whose difference stems from higher-order moments.
Still within the study of sampling dynamics in diffusion models, \cite{ventura2024spectral, achilli2024losing, wang2024diffusionmodelslearnlowdimensional, stanczuk_your_2023, ross2024geometric} investigated the evolution of the score function’s geometry in space and time. In particular, \cite{ventura2024spectral, achilli2024losing} demonstrated that: directions of higher variance in the data distribution are sampled earlier, as the score function progressively reconstructs the latent data structure; at later stages, data points that are strongly aligned with these high-variance directions are preferentially memorized, reflecting a temporal progression that parallels the earlier emergence of generalization.   

\textbf{Memorization and generalization in score learning.} Another line of research has applied similar tools to study how diffusion models generalize the target distribution and memorize data points when parametrized by simple neural network architectures. In particular, \cite{bonnaire2025diffusion, george2025analysis} study how a RF model \cite{rahimi2007random} fits the score function of a diffusion model when the data distribution is a unimodal multivariate Gaussian. 
Memorization is connected to the local density of training points in data space in \cite{merger2026localcoveragegovernsmemorization}, with low-density regions shown to be more likely to be memorized.
The transition from memorization to generalization as dataset size increases has been widely documented \cite{kadkhodaie_generalization_2023}, with early stopping acting as a crucial implicit regularisation mechanism driven by time-scale separation in overparameterized networks \cite{bonnaire2025diffusion, favero2025bigger, urfin2026learnmemorizespectralbias}: in particular, \cite{urfin2026learnmemorizespectralbias} identifies the mechanism behind the time-scale separation, analysing the spectrum of the Gram matrix of the Neural Tangent Kernel. 
Notably, the universal time-scale separation between feature learning and overfitting demonstrated in two-layer regression networks \cite{montanari2025dynamical} closely parallels these memorization dynamics and transitions, with the crucial difference that overfitting in diffusion models is always malign \cite{urfin2026doubledescentmalignoverfitting, latourellevigeant2026generalizationmemorizationoverfittingdiffusion}.
Memorization has also been explicitly detected by analysing the Hessian of the score in models with outliers \cite{jeon2024understanding}, conceptually echoing earlier findings on spurious correlations \cite{sagawa2020investigation}. Unlike these studies, which primarily focus on unimodal distributions or isolated microscopic outliers, our work provides a solvable analytical framework for macroscopic data mixtures, explicitly deriving how the combination of class imbalance and cluster variance dictates distinct, class-specific critical times for both learning and memorization.

Furthermore, multiple studies address score learning when the model is parametrized by a linear or shallow denoiser. \cite{mendes2026solvable} showed that while PCA fails to recover signals in a spiked cumulant model, shallow non-linear autoencoders succeed. 
Other works \cite{wang2025analytical, merger2025generalization} focus on the dynamical manner in which linear denoisers learn the data covariance. \cite{wang2025analytical} shows that directions of large variance are learned first during training, consistent with the fact that such directions are sampled earlier in the diffusion dynamics \cite{ventura2024spectral}.
Previous studies characterized linear auto-encoders as correlation machines utilizing power iteration for generative modeling \cite{weitzner2025diffusion}. Within more complex architectures, U-Net diffusion models yield multiplicative compositional representations \cite{liang2024diffusion}, whereas exact ODEs defined the online SGD dynamics of diffusion models utilizing two-layer autoencoders \cite{cui2023analysis, cui2025solvable}. Although these frameworks identify variance as a driver of learning speed, they omit the role of heterogeneous datasets and their contribution to learning. We extend this understanding by demonstrating that while variance is the dominant force, other contributing factors, such as centroid norm and class imbalance, play a crucial role, becoming the decisive factors that dictates the learning sequence when class variances are comparable.


\textbf{Data imbalance and algorithmic bias.} Despite recent theoretical advances \cite{sagawa2020investigation,jain2024bias,sarao2025bias,pezzicoli2025class}, analytical literature on the effect of heterogeneous datasets and data imbalance remains extremely sparse. Empirically, it has been shown that generative models can exacerbate biases relating to demographic traits such as race, gender, and age, even when trained on balanced datasets \cite{perera2023analyzing}. While the evolution of bias across modern diffusion models is highly dataset-dependent \cite{vice2025exploring}, mechanistic interpretability approaches suggest that these biases often stem from neurons encoding for a multitude of semantic concepts—termed polysemantic—within the architectures \cite{shi2025dissecting, kwon2022diffusion}. However, a rigorous theoretical understanding of \textit{why} and \textit{when} these models prioritize majority over minority classes during training is lacking. Our work bridges this critical gap,
providing exact analytical predictions demonstrating that data structure and severe class imbalance force the features of undersampled modes to emerge at distinct, lower inference times.

\clearpage
\phantomsection
\addcontentsline{toc}{part}{Appendices for theory}
{\Large \bfseries Appendices for theory}

\vspace{0.5em}

\section{The solution of the dynamics}
\label{sec:dynamics_solution_derivation}

We consider the gradient flow (GF) limit of the gradient descent of the readout weights:
\begin{equation}
    \label{eq:GD_readouts}
    \mathbf{A}_{k+1} = \mathbf{A}_k-\eta \frac{\partial \mathcal{E}_{train}}{\partial \mathbf{A}}|_{\mathbf{A}=\mathbf{A}_k}.
\end{equation}
where the training loss is defined in \eqref{eq:empirical_risk_score_denoising}.
We consider the small learning rate limit: we set $\eta \approx \widetilde{\eta}\,d\tau$, so that $\mathbf{A}_{k+1}-\mathbf{A}_{k}\approx d\mathbf{A}$ and
\begin{equation*}
    d\mathbf{A} = -\widetilde{\eta}\frac{\partial \mathcal{E}_{train}}{\partial \mathbf{A}}\Biggl|_{\mathbf{A}=\mathbf{A}_\tau}\,d\tau.
\end{equation*}
The GF equation then reads
\begin{equation}
    \label{eq:GF_readouts}
    \dot{\mathbf{A}} = -\widetilde{\eta}\frac{\partial \mathcal{E}_{train}}{\partial \mathbf{A}}\Biggl|_{\mathbf{A}=\mathbf{A}(\tau)}\equiv -2\widetilde{\eta}\frac{\Delta_t}{N}\left[\mathbf{A}\mathbf{U}+\frac{1}{\sqrt{\Delta_t}}\mathbf{V}^\intercal\right].
\end{equation}
with $\mathbf{U}$ and $\mathbf{V}$ defined in \eqref{eq:U_def}, \eqref{eq:V_def}.
The GF equation is an ODE for the matrix $\mathbf{A}$, and thus admits an explicit solution. Let us consider the Singular Value Decomposition (SVD) of $\mathbf{A}=\mathbf{M}_L^\intercal \mathbf{S}_A \mathbf{M}_R$, with $\mathbf{M}_L\in \Real^{N\times N}$ and $\mathbf{M}_R\in \Real^{P\times P}$ two orthogonal matrices: then the GF equation becomes
\begin{equation}
    \dot{\mathbf{S}}_A = -2\widetilde{\eta}\frac{\Delta_t}{N} \left[\mathbf{S}_A\widehat{\mathbf{U}}+\frac{1}{\sqrt{\Delta_t}}\widehat{\mathbf{V}}^\intercal\right],
\end{equation}
with $\widehat{\mathbf{U}}=\mathbf{M}_R \mathbf{U} \mathbf{M}_R^\intercal$ and $\widehat{\mathbf{V}}=\mathbf{M}_R \mathbf{V} \mathbf{M}_L^\intercal$.
Since the matrix $\mathbf{S}_A$ is non-zero only on the diagonal
with indices $i=1,\dots, N$, we have the set of scalar equations for the singular values $\sigma_A^{(i)}$
\begin{equation}
    \dot{\sigma}_A^{(i)} = -2\widetilde{\eta}\frac{\Delta_t}{N} \left[\sigma_A^{(i)} \widehat{\lambda}_i+\frac{1}{\sqrt{\Delta_t}}\widehat{v}_i\right]
\end{equation}
where $\widehat{\lambda}_i\equiv \widehat{U}_{ii}$ and $\widehat{v}_i=\widehat{V}_{ii}$.
They admit the solutions
\begin{equation}
    \sigma_A^{(i)} = -\frac{1}{\sqrt{\Delta}_t}\frac{\widehat{v}_i}{\widehat{\lambda}_i}(1-e^{-2 \widetilde{\eta}\frac{\Delta_t}{N} \widehat{\lambda}_i\tau})+\sigma_A^{(i)}(0)e^{-2 \widetilde{\eta}\frac{\Delta_t}{N} \widehat{\lambda}_i\tau}\qquad i=1,\dots, N
\end{equation}
Taking the matrix form of this last equation and projecting back, we obtain
\begin{equation}
\label{eq:dyn}
    \mathbf{A}(\tau) = -\frac{1}{\sqrt{\Delta_t}}\mathbf{V}^\intercal \mathbf{U}^{-1}\left[\mathbf{I}_P-e^{-\frac{2\widetilde{\eta}\Delta_t}{N}\mathbf{U}\tau}\right]+\mathbf{A}(0)e^{-\frac{2\widetilde{\eta}\Delta_t}{N}\mathbf{U}\tau}.
\end{equation}
By setting $\mathbf{A}(0)=\mathbf{0}$ and $\widetilde{\eta}=N/\Delta_t$, we obtain Eq.~\eqref{eq:dynamics_readouts_solution}.

\section{The GEP for non-centered data}
\label{sec:GEP_non_centered}

Let us consider a vector field $\bm{\varphi}$ as follows:
\begin{equation}
    \bm{\varphi} = \phi\left(\frac{1}{\sqrt{N}}\mathbf{W}\bm{x}\right)
\end{equation}
where $\phi$ is a non-linear function applied entrywise to the vector $\bm{u}=\frac{1}{\sqrt{N}}\mathbf{W}\bm{x}$, $\bm{x}\in \Real^N$ is a variable with mean $\bm{m}$ and covariance $\mathbf{\Sigma}$, while $\mathbf{W}\in\Real^{P\times N}$ is a random matrix with normal entries. By the central limit theorem, for $N\rightarrow \infty$ the vector $\bm{u}$ is a Gaussian vector with the following mean and covariance matrix:
\begin{equation}
    \bm{\mu}_u=\frac{1}{\sqrt{N}}\mathbf{W} \bm{m}
\end{equation}
\begin{equation}
    \boldsymbol{\Sigma}_{u} = \frac{1}{N}\mathbf{W}\mathbf{\Sigma}\mathbf{W}^\intercal
\end{equation}
We now evaluate the first two cumulants of $\bm{\varphi}$: the first reads
\begin{equation}
    \boldsymbol{\alpha}\equiv
    \Expected{\boldsymbol{\varphi}}{\bm{x}\sim \mathcal{P}_x}=
    \Expected{\phi(\boldsymbol{u})}{\boldsymbol{u}\sim \mathcal{N}(\boldsymbol{\mu}_{u}, \mathbf{\Sigma}_{u})}= \Expected{\phi(\sigma z\bm{1}_P+\bm{\mu}_u)}{z}+\mathcal{O}_N(N^{-1/2})
\end{equation}
where we indicated with $\mathcal{P}_x$ the distribution of $\bm{x}$ and
\begin{equation}
    \sigma^2=\frac{1}{N}\operatorname{Tr}\mathbf{\Sigma}
\end{equation}
and $\bm{1}_P\equiv (1,\dots, 1)^\intercal\in \Real^{P}$. We used the fact that for $N\gg 1$ one has $\Sigma_{p p}^{(u)}\simeq \frac{1}{N}\operatorname{Tr}\mathbf{\Sigma}=\sigma^2$.
For the non-diagonal entries of the second cumulant, we use the Mehler kernel formula and obtain
\begin{equation}
\label{eq:mk_for_phi_cov}
\begin{gathered}
    \Expected{\varphi_p \varphi_q}{\bm{x}\sim \mathcal{P}_x}-\Expected{\varphi_p}{\bm{x}\sim \mathcal{P}_x}\,\Expected{\varphi_q}{\bm{x}\sim \mathcal{P}_x}= \\
    \, \\
    \sum_{n=0}^{\infty}\frac{1}{n!}\left(\frac{\Sigma_{pq}^{(u)}}{\sigma^2}\right)^n
    \Expected{\operatorname{He}_n(z)\phi(z\sigma+\mu_u^{(p)})}{z}
    \Expected{\operatorname{He}_n(z)\phi(z\sigma+\mu_u^{(q)})}{z}-\alpha_p\alpha_q \\
    \, \\
    =\frac{1}{\sigma^2}\gamma_p\gamma_q\Sigma_{pq}^{(u)}+\mathcal{O}_N(N^{-1}),\qquad \bm{\gamma}=\Expected{z\,\phi(z\sigma\bm{1}_P+\bm{\mu}_u)}{z}
\end{gathered}
\end{equation}
Note that the off-diagonal elements of the covariance vanish in the high-dimensional limit $\Sigma_{pq}^{(u)}=\mathcal{O}(1/\sqrt{N})$.
The second moment of the diagonal entries reads
\begin{equation}
    \bm{\beta}\equiv\Expected{\boldsymbol{\varphi}^{\odot 2}}{\bm{x}\sim \mathcal{P}_x}=
    \Expected{\phi^{\odot 2}(\boldsymbol{u})}{\boldsymbol{u}\sim \mathcal{N}(\boldsymbol{\mu}_{u}, \mathbf{\Sigma}_{u})}=
    \Expected{\phi^2(\sigma z\bm{1}_P+\bm{\mu}_{u})}{z}+\mathcal{O}_N(N^{-1/2}).
\end{equation}
The second order cumulant is then
\begin{equation}
    \Expected{\boldsymbol{\varphi}\boldsymbol{\varphi}^\intercal}{\bm{x}\sim \mathcal{P}_x}-\Expected{\boldsymbol{\varphi}}{\bm{x}\sim \mathcal{P}_x}\Expected{\boldsymbol{\varphi}}{\bm{x}\sim \mathcal{P}_x}^\intercal \simeq \frac{1}{\sigma^2}\bm{\gamma}\bm{\gamma}^\intercal \odot \bm{\Sigma}_u+\operatorname{diag}(\bm{\beta}-\bm{\gamma}^{\odot 2}-\bm{\alpha}^{\odot 2}).
\end{equation}
The first two cumulants of the non-linear field $\bm{\varphi}$ are equivalent to those of a linear field $\bm{\varphi}_{gep}$ with Gaussian noise:
\begin{equation}
\label{eq:GEP_non_centered}
\boxed{
    \,\boldsymbol{\varphi}_{gep}(\bm{u})=\boldsymbol{\alpha}+\frac{1}{\sigma}(\boldsymbol{u}-\boldsymbol{\mu}_{u})\odot\, \boldsymbol{\gamma}+\boldsymbol{\eta}\,\odot\,(\boldsymbol{\beta}-\boldsymbol{\gamma}^{\odot 2}-\boldsymbol{\alpha}^{\odot 2})^{\odot \frac{1}{2}}\qquad \boldsymbol{\eta}\sim \mathcal{N}(\mathbf{0}, \mathbf{I}_P)\,
}
\end{equation}
Note that if we set $\boldsymbol{m}=\bm{0}$ and take an odd activation $\phi$, then $\bm{\alpha}=\bm{0}$, $\boldsymbol{\mu}_{u}=\bm{0}$ and we obtain the known formula for the GEP for centered variables:
\begin{equation}
    \boldsymbol{\varphi}_{gep}(\bm{u})=\frac{1}{\sigma}\gamma\,\boldsymbol{u}+\sqrt{\beta-\gamma^2}\,\boldsymbol{\eta}
\end{equation}

We can apply the previous GEP formula to a generic multimodal distribution that admits a pure-density decomposition \cite{achilli2026theory}. 
Indeed, one has
\begin{equation}
    \Expected{\bm{\varphi}\boldsymbol{\varphi}^\intercal}{\boldsymbol{x}\sim \sum_{c=1}^C b_c \mathcal{P}_c(\cdot)}
    =
    \sum_{c=1}^C b_c\,\Expected{\boldsymbol{\varphi}_{gep}^{(c)}{\boldsymbol{\varphi}_{gep}^{(c)}}^\intercal}{\bm{x} \sim \mathcal{N}(\mathbf{m}_c, \boldsymbol{\Sigma}_c),\, \boldsymbol{\eta}_c \sim \mathcal{N}(\mathbf{0}, \mathbf{I}_P)},
\end{equation}
where we apply the GEP to each component of the mixture:
\begin{equation}
\label{eq:GEP_non_centered_conditional}
   \boldsymbol{\varphi}_{gep}^{(c)}(\bm{u})= \boldsymbol{\alpha}_c+\frac{1}{\sigma_c}(\boldsymbol{u}-\boldsymbol{\mu}_{u}^{(c)})\odot\, \boldsymbol{\gamma}_c+\boldsymbol{\eta}_c\,\odot\,(\boldsymbol{\beta}_c-\boldsymbol{\gamma}_c^{\odot 2}-\boldsymbol{\alpha}_c^{\odot 2})^{\odot \frac{1}{2}}\qquad \boldsymbol{\eta}_c\sim \mathcal{N}(\mathbf{0}, \mathbf{I}_P).
\end{equation}

\section{The GEP of feature correlation matrices}
\label{sec:GEP_feature_correlation_matrices}

We derive the GEP expressions for the feature-feature and feature-noise correlation matrices in Eqs.~\eqref{eq:U_def}, \eqref{eq:V_def}, that we rewrite here for reader's convenience:
\begin{equation*}
\begin{gathered}
    \mathbf{U}=\sum_{c=1}^C b_c \mathbf{U}_c, \qquad 
    \mathbf{U}_c=\frac{1}{M_c}\sum_{\nu\in I_c}\Expected{
    \phi\left(\frac{1}{\sqrt{N}}\mathbf{W}\bm{x}_{\nu}(t)\right)
    \phi\left(\frac{1}{\sqrt{N}}\mathbf{W}\bm{x}_{\nu}(t)\right)^\intercal}{
    \bm{\xi}
    }, \\
    \, \\
    \mathbf{V}=\sum_{c=1}^C b_c \mathbf{V}_c,
    \qquad
    \mathbf{V}_c=\frac{1}{M_c}\sum_{\nu\in I_c}\Expected{\phi\left(\frac{1}{\sqrt{N}}\mathbf{W}\bm{x}_{\nu}(t)\right)\bm{\xi}^\intercal}{
    \bm{\xi}
    }, \\
    \, \\
    \bm{x}_{\nu}(t)=\bm{x}_{\nu}e^{-t}+\sqrt{\Delta_t}\bm{\xi}.
\end{gathered}
\end{equation*}
We also derive the GEP of the corresponding population matrices, 
\begin{equation}
    \label{eq:U_pop_def}
    \widetilde{\mathbf{U}}=\sum_{c=1}^C b_c \widetilde{\mathbf{U}}_c,
    \qquad
    \widetilde{\mathbf{U}}_c=\Expected{
    \phi\left(\frac{1}{\sqrt{N}}\mathbf{W}\bm{x}(t)\right)
    \phi\left(\frac{1}{\sqrt{N}}\mathbf{W}\bm{x}(t)\right)^\intercal}{
    \bm{\xi}, \bm{x}(0)\sim \mathcal{N}_c}
\end{equation}
\begin{equation}
    \label{eq:V_pop_def}
    \widetilde{\mathbf{V}}=\sum_{c=1}^C b_c \widetilde{\mathbf{V}}_c,
    \qquad
    \widetilde{\mathbf{V}}_c=\Expected{\phi\left(\frac{1}{\sqrt{N}}\mathbf{W}\bm{x}(t)\right)\bm{\xi}^\intercal}{
    \bm{\xi},\bm{x}(0)\sim\mathcal{N}_c}
\end{equation}
that will be used in Appendix~\ref{sec:analytical_prediction_errors}. Our derivation follows the steps of those in \cite{bonnaire2025diffusion, george2025denoisingscorematchingrandom}, focusing on mixtures of isotropic modes, i.e., with $\Sigma_c \equiv \sigma_c^2 \mathbf{I}_N$.

\subsection{The GEP of $\mathbf{U}$}

We begin by deriving the diagonal elements of $\mathbf{U_c}$: they can be obtained straightforwardly from the concentration of the empirical mean in \eqref{eq:U_def}
\begin{equation}
\label{eq:beta_tilde}
\begin{gathered}
    U_\mathrm{c}^{(qq)} = \frac{1}{M_c}\sum_{\nu\in I_c}\Expected{\phi^2\left(\frac{1}{\sqrt{N}}\sum_{j=1}^NW_{q j}x_j^{(\nu)}(t)\right)}{\boldsymbol{\xi}}=\widetilde{\beta}_c^{(q)}(t)+\mathcal{O}_M(M^{-1/2}) \\
    \, \\
    \widetilde{\bm{\beta}}_c(t) = \Expected{\phi^{\odot 2}\left(z\,\Gamma_c(t)\bm{1}_P+e^{-t}\bm{\mu}_c\right)}{z}.
\end{gathered}
\end{equation}
We introduced the variance of class $c$ at time $t$:
\begin{equation}
    \Gamma_c^2(t)=\sigma_c^2e^{-2t}+\Delta_t.
\end{equation}

As for the off-diagonal elements of $\mathbf{U}_c$, one first introduces the diffusion noise latent vector
\begin{equation}
    \mathrm{u} = \frac{1}{\sqrt{N}}\mathbf{W}\boldsymbol{\xi}
\end{equation}
which has zero mean and a covariance with unit diagonal entries and off-diagonal elements $\Sigma_u^{(p q)}=\frac{1}{N}\sum_{i=1}^N W_{pi}W_{qi}$. Then, by applying the Mehler kernel formula to \eqref{eq:U_def}, we get
\begin{equation}
\begin{gathered}
\label{eq:U_Kernel_Mehler}
    U_c^{(p q)} = \frac{1}{M_c}\sum_{\nu\in I_c} \phi_0^{(t)}\left(
    \frac{1}{\sqrt{N}}\sum_{i=1}^N\,W_{pi}x_i^{(\nu)}
    \right)\phi_0^{(t)}\left(
    \frac{1}{\sqrt{N}}\sum_{i=1}^N\,W_{qi}x_i^{(\nu)}
    \right) \\
    \, \\
    +\frac{\Sigma_{u}^{(p q)}}{M_c}\sum_{\nu\in I_c} \phi_1^{(t)}\left(
    \frac{1}{\sqrt{N}}\sum_{i=1}^N\,W_{pi}x_i^{(\nu)}
    \right)\phi_1^{(t)}\left(
    \frac{1}{\sqrt{N}}\sum_{i=1}^N\,W_{qi}x_i^{(\nu)}
    \right)+\mathcal{O}_N(N^{-1}) \\
    \, \\
    \text{with} \\
    \, \\
    \phi_0^{(t)}(x)=\Expected{\phi\left(e^{-t}x+\sqrt{\Delta_t}z\right)}{z}\qquad \phi_1^{(t)}(x)=\Expected{z\,\phi\left(e^{-t}x+\sqrt{\Delta_t}z\right)}{z}       
\end{gathered}
\end{equation}

The truncation of the expansion was possible because $\Sigma_u^{(pq)}=\frac{1}{N}\sum_{i}W_{pi}W_{qi}=\mathcal{O}(N^{-1/2})$. Let us first evaluate the second term in \eqref{eq:U_Kernel_Mehler}: 
we use the following  Hermite expansion of $\phi$ for $c_0>0$ and $c_1\in\Real$
\begin{equation}
    \phi(c_0 x+c_1)=\sum_{n=0}^{\infty}\frac{h_n}{n!}\operatorname{He}_n(x),\qquad h_n=\Expected{\operatorname{He}_n(z)\,\phi(c_0z+c_1)}{z}.
\end{equation}
With all this, we get
\begin{equation}
\label{eq:gamma_tilde}
\begin{gathered}
    \frac{1}{M_c}\sum_{\nu\in I_c} \phi_1^{(t)}\left(
        \frac{1}{\sqrt{N}}\sum_{i=1}^N\,W_{pi}x_i^{(\nu)}
        \right)\phi_1^{(t)}\left(
        \frac{1}{\sqrt{N}}\sum_{i=1}^N\,W_{pi}x_i^{(\nu)}
        \right)^\intercal \\
    \, \\
    \simeq 
    \Expected{y\,\phi\left(e^{-t}u_p+\sqrt{\Delta_t}y\right)}{u\sim \mathcal{N}(\mu_c^{(p)}, \sigma_c^2),\, y}
    \Expected{y\,\phi\left(e^{-t}u_q+\sqrt{\Delta_t}y\right)}{u\sim \mathcal{N}(\mu_c^{(q)}, \sigma_c^2),\, y}
    \, \\
    \, \\
    \text{then}\, \\
    \, \\
    \Expected{y\,\phi\left(e^{-t}u_p+\sqrt{\Delta(t)}y\right)}{u\sim \mathcal{N}(\mu_c^{(p)}, \sigma_c^2),\, y}
    = \sum_{n=0}^{\infty}\frac{h_n^{(p)}}{n!}\Expected{
    y \operatorname{He}_n
    \left(\frac{e^{-t}\sigma_c\widetilde{u}+\sqrt{\Delta_t} y}{\Gamma_t^c}
    \right)
    }{y,\,\widetilde{u}}
    \\
    \, \\
    =\frac{\sqrt{\Delta(t)}}{\Gamma_c(t)}\widetilde{\gamma}_c^{(p)}(t), \\
    \, \\
    \widetilde{\bm{\gamma}}_c(t) = \Expected{z\,\phi\left(z\,\Gamma_c(t)\bm{1}_P+e^{-t}\bm{\mu}_c\right)}{z},
\end{gathered}
\end{equation}
and thus \eqref{eq:U_Kernel_Mehler} becomes
\begin{equation*}
    \begin{gathered}
        U_c^{(p q)} = \frac{1}{M_c}\sum_{\nu\in I_c} \phi_0^{(t)}\left(
        \frac{1}{\sqrt{N}}\sum_{i=1}^N\,W_{pi}x_i^{(\nu)}
        \right)\phi_0^{(t)}\left(
        \frac{1}{\sqrt{N}}\sum_{i=1}^N\,W_{qi}x_i^{(\nu)}
        \right)
        +\frac{\Delta_t}{\Gamma_c^2(t)}\widetilde{\gamma}_c^{(p)}\widetilde{\gamma}_c^{(q)}\Sigma_{u}^{(p q)}.     
    \end{gathered}
\end{equation*}
To evaluate the first term in \eqref{eq:U_Kernel_Mehler}, we use the GEP of Eq.~\eqref{eq:GEP_non_centered_conditional}:
\begin{equation}
\begin{gathered}
    \phi_0\left(\frac{1}{\sqrt{N}}\mathbf{W}\bm{x}^{(\nu)}\right)\overset{\text{GEP}} {\rightarrow} \bm{G}_c^{(\nu)} \\
    \, \\
    \bm{G}_c^{(\nu)}\equiv \boldsymbol{\alpha}_c+\frac{1}{\sigma_c\sqrt{N}}\mathbf{W}(\bm{x}^{(\nu)}-\boldsymbol{m}_c)\odot \boldsymbol{\gamma}_c+(\boldsymbol{\beta}_c-\boldsymbol{\alpha}_c^{\odot2}-\boldsymbol{\gamma}_c^{\odot 2})^{\odot\frac{1}{2}}\odot \boldsymbol{\eta}_\nu^c,
\end{gathered}
\end{equation}
where the vectors of Hermite coefficients can be computed using their definition in Appendix~\ref{sec:GEP_non_centered} and the definition of $\phi_0$ in \eqref{eq:U_Kernel_Mehler}:
\begin{equation}
\label{eq:alpha}
    \boldsymbol{\alpha}_c(t) = 
    \Expected{\phi_0(z\,\sigma_c\bm{1}_P+\bm{m}_c)}{z}
    =\Expected{\phi(z\,\Gamma_c(t)\,\mathbf{1}_P+e^{-t}\boldsymbol{\mu}_c)}{z}
\end{equation}
\begin{equation}
\label{eq:beta}
\begin{gathered}
    \, \\
    \boldsymbol{\beta}_c(t) = 
    \Expected{\phi_0^{\odot 2}(z\,\sigma_c\,\mathbf{1}+\boldsymbol{m}_c)}{z} \\
    \, \\
    =\Expected{\phi(\Gamma_c(t)u\,\mathbf{1}_P+e^{-t}\boldsymbol{\mu}_c)\odot
    \phi(\Gamma_c(t)v\,\mathbf{1}_P+e^{-t}\boldsymbol{\mu}_c)}{u, v\sim\mathcal{N}(0, \boldsymbol{\Sigma}_c^{(2)}(t))} \\
    \, \\
    \text{with} \\
    \, \\
    \boldsymbol{\Sigma}_c^{(2)}(t) =
    \begin{pmatrix}
        1 & \frac{e^{-2 t}\sigma_c^2}{(\Gamma_c(t))^2} \\
         & \\
        \frac{e^{-2 t}\sigma_c^2}{(\Gamma_c(t))^2} & 1
    \end{pmatrix} \\
    \,
\end{gathered}
\end{equation}
\begin{equation}
\label{eq:gamma}
    \boldsymbol{\gamma}_c(t) = 
    \Expected{z\,\phi_0(z\,\sigma_c\bm{1}_P+\bm{m}_c)}{z}
    =\frac{\sigma_c e^{-t}}{\Gamma_c(t)}\widetilde{\boldsymbol{\gamma}}_c(t)
\end{equation}

\begin{figure*}
    \centering
    \includegraphics[width=0.9\linewidth]{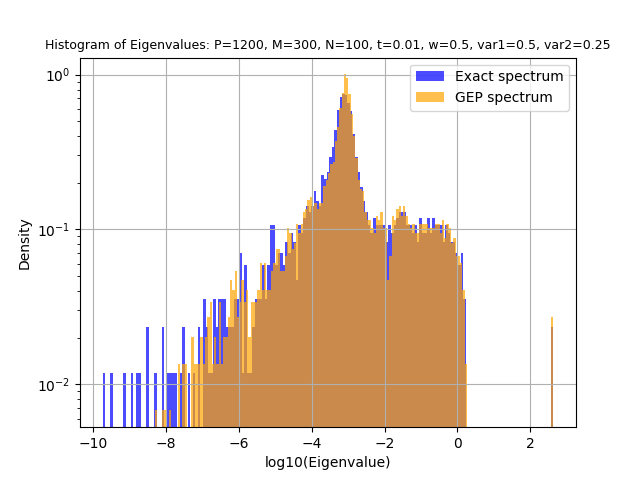}
    \caption{Comparison of the GEP derived in Eq.~\eqref{eq:GEP_U_matrix_form} with the spectrum of $U$ obtained through a Monte Carlo estimation of \eqref{eq:U_def}.}
    \label{fig:test_GEP_U_with_exactMC}
\end{figure*}

Putting everything together, the GEP form of the matrix $\mathbf{U}$ is equal to

\begin{equation}
\boxed{
    \label{eq:GEP_U_matrix_form}
\begin{gathered}
    \mathbf{U}\sim \sum_{c=1}^C b_c \mathbf{U}_c^{(gep)}, 
    \, \\
    \,\mathbf{U}_{c}^{(gep)}=\frac{1}{M_c}\sum_{\nu\in I_c}\bm{G}_c^{(\nu)}(t)\bm{G}_c^{(\nu)}(t)^\intercal + \frac{\Delta(t)}{N \sigma_c^2 e^{-2 t}}\boldsymbol{\gamma}_c(t)\boldsymbol{\gamma}_c(t)^\intercal\odot\mathbf{W}\mathbf{W}^\intercal\, \\
    +\operatorname{diag}
    \left(
    \widetilde{\boldsymbol{\beta}}_c(t)-\boldsymbol{\beta}_c(t)
    -\frac{\Delta_t}{\sigma_c^2 e^{-2 t}}\boldsymbol{\gamma}_c(t)^{\odot 2}
    \right), \\
    \, \\
    \,\bm{G}_c^{(\nu)}= \boldsymbol{\alpha}_c+\frac{1}{\sigma_c\sqrt{N}}\mathbf{W}(\bm{x}^{(\nu)}-\boldsymbol{m}_c)\odot \boldsymbol{\gamma}_c+(\boldsymbol{\beta}_c-\boldsymbol{\alpha}_c^{\odot2}-\boldsymbol{\gamma}_c^{\odot 2})^{\odot\frac{1}{2}}\odot \boldsymbol{\eta}_\nu^c\, \\
    \, \\
\end{gathered}
}
\end{equation}
In \eqref{eq:U_GEP} we set
\begin{equation}
    \widehat{\mathbf{W}}_c \equiv \bm{\gamma}_c\bm{1}_N^\intercal \odot \mathbf{W},\qquad \bm{d}_c=\widetilde{\boldsymbol{\beta}}_c(t)-\boldsymbol{\beta}_c(t)
    -\frac{\Delta_t}{\sigma_c^2 e^{-2 t}}\boldsymbol{\gamma}_c(t)^{\odot 2}.
\end{equation}
Note that the GEP matrix can be decomposed in terms of a finite-norm matrix $\widehat{\mathbf{U}}_c^{(gep)}$ plus a low rank extensive perturbation
\begin{equation}
\label{eq:U_gep_highlight_perturbations}
    \mathbf{U}_c^{(gep)} = \widehat{\mathbf{U}}_c^{(gep)}+ \boldsymbol{\alpha}_c\boldsymbol{\alpha}_c^\intercal
    +\left[\frac{\boldsymbol{\alpha}_c}{M_c}\sum_{\nu\in I_c}(\mathbf{G}_{c}^{(\nu)}-\boldsymbol{\alpha}_c)^\intercal+\frac{1}{M_c}\sum_{\nu\in I_c}(\mathbf{G}_{c}^{(\nu)}-\boldsymbol{\alpha}_c)\boldsymbol{\alpha}_c^\intercal\right]
\end{equation}
The perturbation is extensive because its first term is a rank-one matrix of norm $\mathcal{O}_P(P)$, while the other is $\mathcal{O}_P(1)$. The first term in particular is related to the projection of the centroids in the space of features: this can be seen explicitly in the large $t$ limit:
\begin{equation}
    \bm{\alpha}_c \approx \gamma \,e^{-t}\bm{\mu}_c.
\end{equation}
where in the last equation $\gamma=\Expected{z\phi(z)}{z}$.
We tested the goodness of the GEP in \eqref{eq:GEP_U_matrix_form}: in Fig.~\ref{fig:test_GEP_U_with_exactMC}, we compare the spectrum of $\mathbf{U}_{\mathrm{gep}}$ with that of the exact $\mathbf{U}$, which was estimated with Monte-Carlo.

\subsection{The GEP of $\widetilde{\mathbf{U}}$}

To compute the GEP of the corresponding population matrix, we simply use the GEP on the vector field
\begin{equation*}
    \bm{\varphi}=\phi\left(\frac{1}{\sqrt{N}}\mathbf{W}\bm{x}(t)\right)
\end{equation*}
with $\bm{x}(t)$ sampled from $\mathcal{N}(\cdot, \bm{m}_c e^{-t}, \Gamma_c^2(t)\mathbf{I}_N)$. We get
\begin{equation}
    \bm{\varphi}_{gep}^{(c)}=\boldsymbol{\alpha}_c+\frac{1}{\sigma_c\sqrt{N}}\mathbf{W}(\bm{x}^{(\nu)}-\boldsymbol{m}_c)\odot \boldsymbol{\gamma}_c+(\widetilde{\boldsymbol{\beta}}_c-\boldsymbol{\alpha}_c^{\odot2}-\widetilde{\boldsymbol{\gamma}}_c^{\odot 2})^{\odot\frac{1}{2}}\odot \boldsymbol{\eta}_\nu^c\qquad \boldsymbol{\eta}_\nu^c\sim \mathcal{N}(\mathbf{0}, \mathbf{I}_P)
\end{equation}
and thus the GEP of $\widetilde{\mathbf{U}}$ is obtained by averaging the $\bm{\varphi}_{gep}^{(c)}{\bm{\varphi}_{gep}^{(c)}}\,^\intercal$ over the $\bm{x}\sim \mathcal{N}_c$ and the $\bm{\eta}$, obtaining
\begin{equation}
\boxed{
    \label{eq:U_tilde_GEP}
    \begin{gathered}
    \widetilde{\mathbf{U}}\sim \sum_{c=1}^C b_c \widetilde{\mathbf{U}}_c^{(gep)},
    \, \\
    \, \\
    \widetilde{\mathbf{U}}_c^{(gep)}=\boldsymbol{\alpha}_c \boldsymbol{\alpha}_c^\intercal + \frac{1}{N}\mathbf{W}\mathbf{W}^\intercal\odot \widetilde{\boldsymbol{\gamma}}_c \widetilde{\boldsymbol{\gamma}}_c^\intercal+\operatorname{diag}(\widetilde{\boldsymbol{\beta}}_c-\boldsymbol{\alpha}_c^{\odot2}-\widetilde{\boldsymbol{\gamma}}_c^{\odot 2}).
    \end{gathered}
}
\end{equation}
One can easily verify that by averaging $\mathbf{U}_c^{(gep)}$ over the dataset and the GEP noise, one directly obtains \eqref{eq:U_tilde_GEP}.

It is insightful to relate $\mathbf{U}$ to its population version $\widetilde{\mathbf{U}}$: with a few straightforward manipulations, one can show that
\begin{equation}
\label{eq:isolate_gen_mem}
\begin{gathered}
    \mathbf{U}_c^{gep} = \widetilde{\mathbf{U}}_c^{gep}+\mathbf{\delta U}_c^{(gep)} \\
    \, \\
    \mathbf{\delta U}_c^{(gep)} = \frac{1}{M_c}\sum_{\nu\in I_c}\bm{G}_c^{(\nu)}(t)\bm{G}_c^{(\nu)}(t)^\intercal
    -\frac{1}{N}\mathbf{W}\mathbf{W}^\intercal\odot \boldsymbol{\gamma}_c \boldsymbol{\gamma}_c^\intercal \\
    +\operatorname{diag}
    \left(
    \widetilde{\boldsymbol{\beta}}_c-\boldsymbol{\beta}_c
    -\frac{\Delta_t}{\sigma_c^2 e^{-2 t}}\boldsymbol{\gamma}_c^{\odot 2}
    \right) 
    -\operatorname{diag}(\widetilde{\boldsymbol{\beta}}_c-\boldsymbol{\alpha}_c^{\odot2}-\widetilde{\boldsymbol{\gamma}}_c^{\odot 2})
    -\bm{\alpha}_c\bm{\alpha}_c^\intercal
\end{gathered}
\end{equation}
The generalization during training is due to the population matrix $\widetilde{\mathbf{U}}\sim \sum_c b_c \widetilde{\mathbf{U}}_c^{gep}$, while memorization originates from the matrix $\delta\mathbf{U}\sim \sum_c b_c \delta\mathbf{U}_c^{gep}$.
Proceeding similarly to \cite{bonnaire2025diffusion}, one can verify that $\lVert \widetilde{\mathbf{U}}-\sum_c b_c \bm{\alpha}_c\bm{\alpha}_c^\intercal \lVert_F = \mathcal{O}(\chi_p)$ and $\lVert \delta\mathbf{U} \lVert_F = \mathcal{O}(\chi_p/\sqrt{\chi_m})$, where $\lVert \cdot \lVert_F$ is the Frobenius norm.

\subsection{The GEP of $\mathbf{V}$}

We take formula \eqref{eq:V_def} and rewrite the latent vector $\bm{u}_{\nu}(t)=\frac{1}{\sqrt{N}}\mathbf{W}\bm{x}_\nu(t)$, where $\nu\in I_c$, as
\begin{equation}
    \bm{u}_\nu(t)=\bm{u}_{\nu}+e^{-t}\bm{\mu}_c
\end{equation}
The vector $\bm{u}_\nu$ is a Gaussian vector with zero mean and covariance $\mathbf{\Sigma}_{uu}=\frac{\Gamma_c^2(t)}{N}\mathbf{W}\mathbf{W}^\intercal$. The cross covariance with the noise reads $\mathbf{\Sigma}_{u\xi}=\frac{\sqrt{\Delta}_t}{\sqrt{N}}\mathbf{W}$. Then the Mehler kernel formula applied to \eqref{eq:V_def} yields
\begin{equation}
\label{eq:V_Kernel_Mehler}
\begin{gathered}
    V_c^{(p j)} = \frac{1}{M_c}\sum_{\nu\in I_c} \sum_{n=0}^{\infty} \frac{(\Sigma_{u\xi}^{(p j)})^n}{n!}\Expected{
    \phi\left(
    z \Gamma_c(t)+e^{-t}\mu_c^{(p)}
    \right)
    \operatorname{He}_n(z)
    }{z}\Expected{z\operatorname{He}_n(z)}{z} \\
    \, \\
    = \sqrt{\frac{\Delta_t}{N}}\frac{\widetilde{\gamma}_c^{(p)}}{\Gamma_c(t)}W_{p j}+\mathcal{O}_N(N^{-1})
\end{gathered}
\end{equation}
Then the GEP for $\mathbf{V}$ reads
\begin{equation}
    \label{eq:V_GEP}
    \boxed{
    \begin{gathered}
        \mathbf{V}\sim \sum_{c=1}^C b_c\mathbf{V}_c^{(gep)} \\
        \, \\
        \mathbf{V}_c^{(gep)}=\sqrt{\frac{\Delta_t}{N}}\frac{\widetilde{\bm{\gamma}}_c \bm{1}_N^\intercal}{\Gamma_c(t)}\odot \mathbf{W}.
    \end{gathered} 
    }
\end{equation}

\subsection{The GEP of $\widetilde{\mathbf{V}}$}

It is easy to see from \eqref{eq:V_Kernel_Mehler} that the GEP of $\widetilde{\mathbf{V}}$ is the same as that of $\mathbf{V}$, up to finite-size corrections: so we write
\begin{equation}
    \label{eq:V_tilde_GEP}
    \boxed{
    \begin{gathered}
        \widetilde{\mathbf{V}}\sim \sum_{c=1}^C b_c\widetilde{\mathbf{V}}_c^{(gep)} \\
        \, \\
        \widetilde{\mathbf{V}}_c^{(gep)}=\mathbf{V}_c^{(gep)}.
    \end{gathered} 
    }
\end{equation}

\section{The matrix $\mathbf{U}$ at large $t$}
\label{sec:expansion_of_U_large_t}

We derive Eq.~\eqref{eq:U_GEP_close_to_spec} through a large $t$ expansion—or equivalently, a small $e^{-t}$ expansion. We start with the Hermite coefficients: for any component $p=1,\dots, P$
\begin{equation}
    \label{eq:alpha_large_t}
\begin{split}
    \alpha_c^{(p)} &= \Expected{\phi(\Gamma_c z+\mu_c^{(p)} e^{-t})}{z}=\int \frac{d z}{(2\pi \Gamma_c^2)^{1/2}}e^{-\frac{(z-\mu_c^{(p)} e^{-t})^2}{2 \Gamma_c^2}}\phi(z) \\
    &\, \\
    &= \Expected{\phi(z)}{z}+\mu_c^{(p)}\Expected{z\phi(z)}{z} e^{-t} + \mathcal{O}(e^{-2 t})
\end{split}
\end{equation}

\begin{equation}
    \label{eq:beta_tilde_large_t}
\begin{split}
    \widetilde{\beta}_c^{(p)} &= \Expected{\phi^2(\Gamma_c z+\mu_c^{(p)} e^{-t})}{z}=\int \frac{d z}{(2\pi \Gamma_c^2)^{1/2}}e^{-\frac{(z-\mu_c^{(p)} e^{-t})^2}{2 \Gamma_c^2}}\phi^2(z) \\
    &\, \\
    &= \Expected{\phi^2(z)}{z}+\mu_c^{(p)}\Expected{z\phi^2(z)}{z} e^{-t} + \mathcal{O}(e^{-2 t})
\end{split}
\end{equation}

\begin{equation}
    \label{eq:beta_large_t}
\begin{split}
    \beta_c^{(p)} &= \Expected{\phi^2(\Gamma_c z+\mu_c^{(p)} e^{-t})}{z} \\
    &=\int \frac{d z_1 dz_2}{2\pi \Gamma_c^2\sqrt{\det \bm{\Sigma}_c^{(2)}}}e^{-\frac{(z_1-\mu_c^{(p)} e^{-t})^2}{2 \Gamma_c^2 \det \bm{\Sigma}_c^{(2)}}
    -\frac{(z_2-\mu_c^{(p)} e^{-t})^2}{2 \Gamma_c^2 \det \bm{\Sigma}_c^{(2)}}
    +\frac{(z_1-\mu_c^{(p)} e^{-t})(z_2-\mu_c^{(p)} e^{-t})\sigma_c^2 e^{-2 t}}{\Gamma_c^4 \det \bm{\Sigma}_c^{(2)}}
    }\phi(z_1)\phi(z_2) \\
    &\, \\
    &= \Expected{\phi(z)}{z}^2 + 2\mu_c^{(p)}\Expected{\phi(z)}{z}\Expected{z\phi(z)}{z}e^{-t}+ \mathcal{O}(e^{-2 t})
\end{split}
\end{equation}

\begin{equation}
    \label{eq:gamma_tilde_large_t}
\begin{split}
    \widetilde{\gamma}_c^{(p)} &= \Expected{z\phi(\Gamma_c z+\mu_c^{(p)} e^{-t})}{z}=\int \frac{d z}{(2\pi \Gamma_c^2)^{1/2}}e^{-\frac{(z-\mu_c^{(p)} e^{-t})^2}{2 \Gamma_c^2}}z\phi(z) \\
    &\, \\
    &= \Expected{z\phi(z)}{z}+\mu_c^{(p)}\Expected{z^2\phi(z)}{z} e^{-t} + \mathcal{O}(e^{-2 t})
\end{split}
\end{equation}

\begin{equation}
    \label{eq:gamma_large_t}
    \gamma_c^{(p)} = \frac{\sigma_c e^{-t}}{\Gamma_c}\widetilde{\gamma}_c^{(p)}=\sigma_c e^{-t}\Expected{z\phi(z)}{z}+\mathcal{O}(e^{-2 t})
\end{equation}

where we also expanded $\Gamma_c(t)=\sqrt{(\sigma_c^2-1) e^{-2 t}+1}\approx 1+\mathcal{O}(e^{-2 t})$. Let us define $\alpha=\Expected{\phi(z)}{z}$, $\beta=\Expected{\phi^2(z)}{z}$, $\gamma=\Expected{z\phi(z)}{z}$.
At leading order in $e^{-t}$, the GEP features read
\begin{equation}
    \label{eq:GEP_features_large_t}
\begin{split}
    \bm{G}_{c,\nu} &= \alpha\bm{1}_P+\gamma e^{-t}\bm{\mu}_c
+\frac{e^{-t}\gamma}{\sqrt{N}}\mathbf{W}(\bm{x}_\nu-\bm{m}_c)+\mathcal{O}(e^{-2 t})
\end{split}
\end{equation}
The other terms in Eq.~\eqref{eq:GEP_U_matrix_form} expand as follows:
\begin{equation}
    \frac{\Delta(t)}{N \sigma_c^2 e^{-2 t}}\boldsymbol{\gamma}_c(t)\boldsymbol{\gamma}_c(t)^\intercal\odot\mathbf{W}\mathbf{W}^\intercal
    =
    \frac{\gamma^2}{N}\mathbf{W}\mathbf{W}^\intercal + \mathcal{O}(e^{-2t})
\end{equation}
\begin{equation}
    \operatorname{diag}\left(\widetilde{\bm{\beta}}_c-\bm{\beta}_c-\frac{\Delta}{\sigma_c^2 e^{-2t}}\bm{\gamma}_c^{\odot 2}\right) = (\beta-\alpha^2-\gamma^2)\mathbf{I}_P + \mathcal{O}(e^{-2 t})
\end{equation}
We now assume that $t=\mathcal{O}(\log N)$, so that $e^{-t}=\mathcal{O}(N^{-1})$: then, we can neglect all corrections but those in the finite-rank perturbation in Eq.~\eqref{eq:GEP_U_matrix_form}, and the matrix $\mathbf{U}$ becomes
\begin{equation}
    \label{eq:U_large_t_leading_order}
    \mathbf{U} = (\alpha\bm{1}_P+\gamma e^{-t}\bm{\mu}_c)(\alpha\bm{1}_P+\gamma e^{-t}\bm{\mu}_c)^\intercal + \frac{\gamma^2}{N}\mathbf{W}\mathbf{W}^\intercal+(\beta-\alpha^2-\gamma^2)\mathbf{I}_P
\end{equation}
For $\phi=\tanh$, $\alpha=0$ and we obtain Eq.~\eqref{eq:U_GEP_close_to_spec}. 

\section{The role of the activation function}
\label{sec:role_of_activation}
\begin{figure}[h!]
    \centering
    \includegraphics[width=0.95\linewidth]{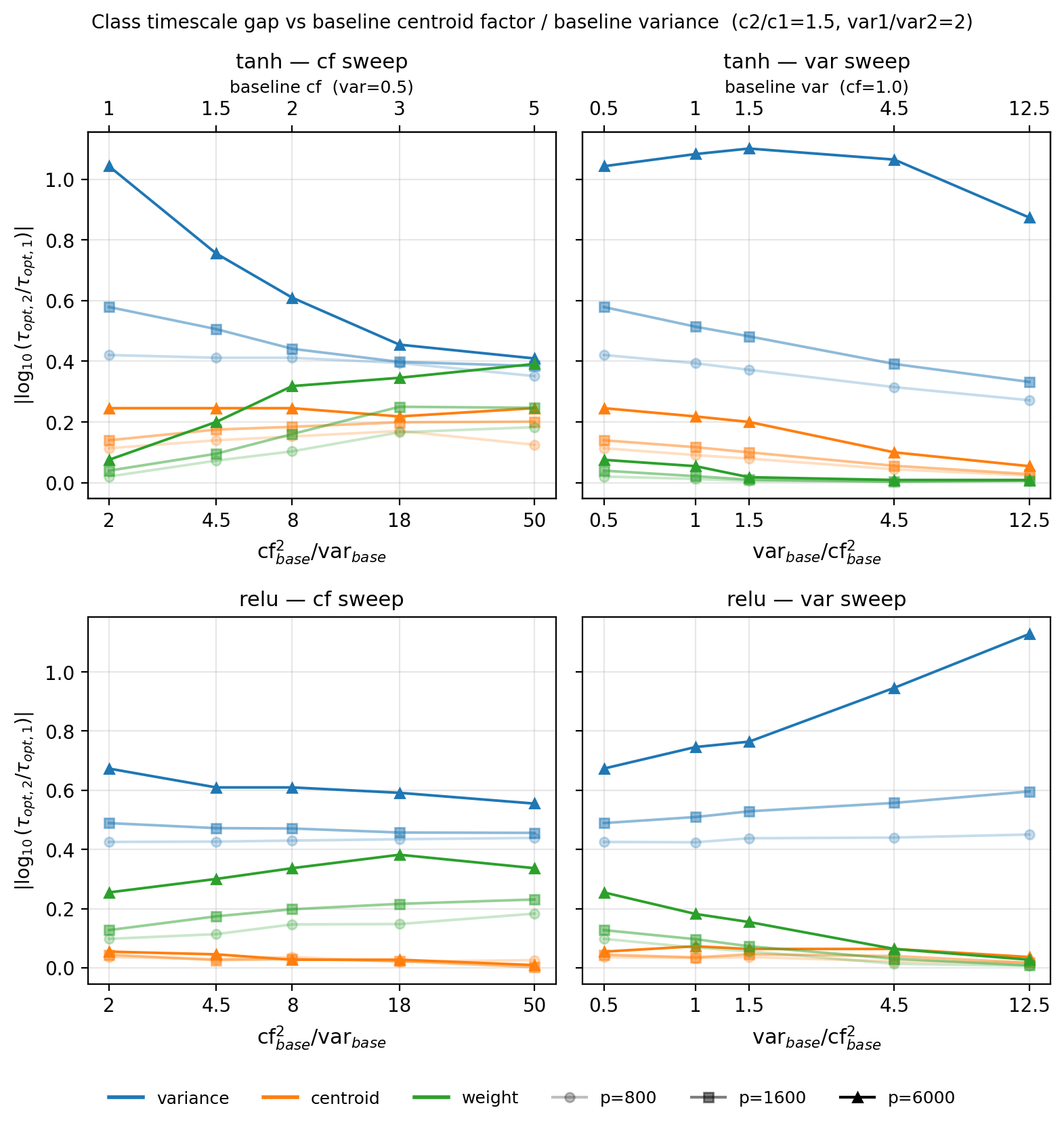}
    \caption{Generalization log-gaps versus baseline ratios: top is for a $\tanh$ activation, bottom for a $\mathrm{ReLU}$ activation; left panels refer to a sweep in baseline centroid factors $\mathrm{cf}_b=1.0, 1.5, 2.0, 3.0, 5.0$, right panels to a sweep in baseline variance $\sigma_b^2=0.5, 1.0, 1.5, 4.5, 12.5$. Curves are obtained from the semi-analytical curves of Eqs.\eqref{eq:errors_semi_analytical}. Parameters are $N=100$, $M=3000$, $P=800, 1600, 6000$.
    }
    \label{fig:relu_vs_tanh_gen}
\end{figure}
\begin{figure}[h!]
    \centering
    \includegraphics[width=0.95\linewidth]{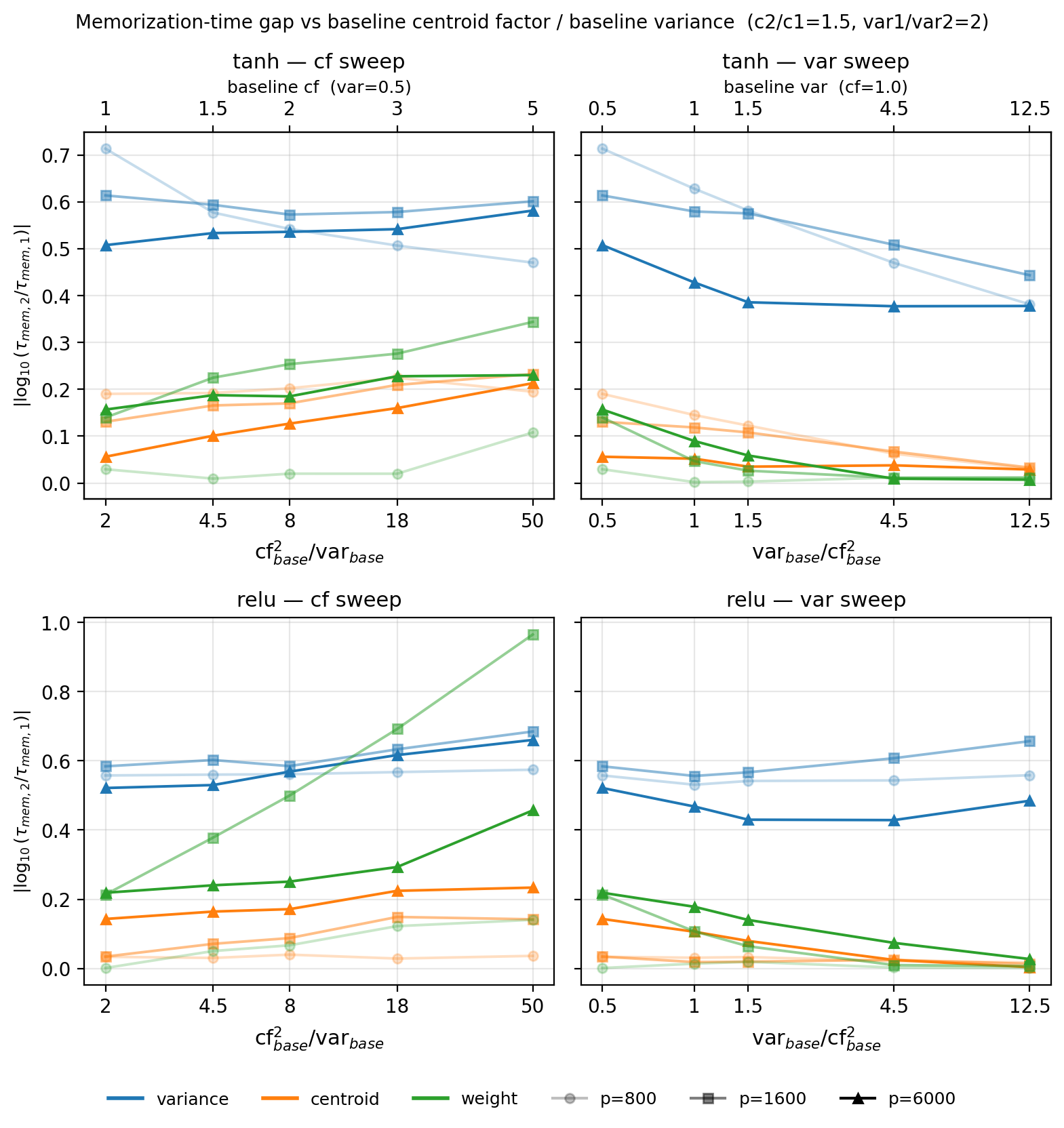}
    \caption{Memorization log-gaps versus baseline ratios: top is for a $\tanh$ activation, bottom for a $\mathrm{ReLU}$ activation; left panels refer to a sweep in baseline centroid factors $\mathrm{cf}_b=1.0, 1.5, 2.0, 3.0, 5.0$, right panels to a sweep in baseline variance $\sigma_b^2=0.5, 1.0, 1.5, 4.5, 12.5$. Curves are obtained from the semi-analytical curves of Eqs.\eqref{eq:errors_semi_analytical}. Parameters are $N=100$, $M=3000$, $P=800, 1600, 6000$.}
    \label{fig:relu_vs_tanh_mem}
\end{figure}

In this Appendix we explain how the choice of the activation function affects the learning dynamics in the DSM of Eq.~\eqref{eq:empirical_risk_score_denoising}: first, we consider the simple case of a linear activation; then, we characterize how the generalization and memorization gaps of sections~\ref{sec:class_specific_gen_mem} change when using a $\mathrm{ReLU}$ activation instead of $\tanh$.

\paragraph{Linear activation}

The role of the activation function is encoded in the Hermite coefficients $\{\bm{\alpha}_c,\bm{\beta}_c,\widetilde{\bm{\beta}}_c, \bm{\gamma}_c\}_{c=1}^C$, defined in eqs~\eqref{eq:alpha},\eqref{eq:beta},\eqref{eq:beta_tilde}, \eqref{eq:gamma}. For a linear activation, the Hermite coefficients simplify to
\begin{equation}
    \label{eq:alpha_simplified}
    \bm{\alpha}_c = \Expected{\phi(\Gamma_c z\bm{1}+\bm{\mu}_c e^{-t})}{z} \rightarrow \bm{\mu}_c e^{-t}
\end{equation}
\begin{equation}
    \label{eq:beta_simplified}
    \bm{\beta}_c = \Expected{\phi(\Gamma_c u\bm{1}+\bm{\mu}_c e^{-t})\odot \phi(\Gamma_c v\bm{1}+\bm{\mu}_c e^{-t})}{u,v\sim \mathcal{N}(\bm{0},\bm{\Sigma}_c^{(2)}(t))} \rightarrow \bm{\mu}_c^{\odot 2} e^{-2t}+\sigma_c^2e^{-2t} \bm{1}
\end{equation}
\begin{equation}
    \label{eq:beta_tilde_simplified}
    \widetilde{\bm{\beta}}_c = \Expected{\phi^2(\Gamma_c z\bm{1}+\bm{\mu}_c e^{-t})}{z} \rightarrow \bm{\mu}_c^{\odot 2} e^{-2t}+\Gamma_c^2 \bm{1}
\end{equation}
\begin{equation}
    \label{eq:gamma_tilde_simplified}
    \widetilde{\bm{\gamma}}_c = \Expected{z\phi(\Gamma_c z\bm{1}+\bm{\mu}_c e^{-t})}{z} \rightarrow \Gamma_c \bm{1}\Longrightarrow \bm{\gamma}_c \rightarrow \sigma_c e^{-t}\bm{1}
\end{equation}

and the GEP correlation matrices $\mathbf{U}_{\mathrm{gep}}$ and $\widetilde{\mathbf{U}}_c^{(\mathrm{gep})},\mathbf{V}_c,\,c=1,\dots,C$ simply reduce respectively to the empirical and population data correlation matrices at diffusion time $t$:
\begin{equation}
    \label{eq:U_linear_activation}
    \mathbf{U}_{\mathrm{gep}} \rightarrow \sum_{c} b_c \bm{\mu}_c\bm{\mu}_c^\intercal e^{-2 t}+\frac{1}{M N}\sum_{\nu=1}^M \mathbf{W}\left[e^{-2 t}(\bm{x}_\nu-\bm{m}_c)(\bm{x}_\nu-\bm{m}_c)^\intercal+\Delta \mathbf{I}_N\right]\mathbf{W}^\intercal
\end{equation}
\begin{equation}
    \label{eq:V_linear_activation}
    \mathbf{V}_c^{(\mathrm{gep})} \rightarrow \sqrt{\frac{\Delta}{N}}\mathbf{W}
\end{equation}
\begin{equation}
    \label{eq:U_tilde_linear_activation}
    \widetilde{\mathbf{U}}_c^{(\mathrm{gep})} \rightarrow 
    \bm{\mu}_c\bm{\mu}_c^\intercal e^{-2 t}
    +\frac{\Gamma_c(t)}{N}\mathbf{W}\mathbf{W}^\intercal
\end{equation}
For linear activations, the bulks of $\mathbf{U}$ and $\widetilde{\mathbf{U}}_c$ are not affected by either the magnitude or the alignment of the centroids, nor by class sampling ratios $b_c$: these parameters only affect the finite-rank perturbation term $\sum_c b_c \bm{\mu}_c\bm{\mu}_c^\intercal$ of the projected centroids $\bm{\mu}_c \equiv \frac{1}{\sqrt{N}}\mathbf{W}\bm{m}_c$. According to the learning curves formula of Eq.~\eqref{eq:errors_in_terms_of_matrices}, only the bulks of the matrices $\mathbf{U},\widetilde{\mathbf{U}}_c,\,c=1,\dots,C$ bear finite contributions to these curves, as any finite-rank perturbation yields sub-leading $\mathcal{O}(1/N)$ corrections: we conclude that generalization and memorization gaps induced by differences in centroid geometry and sampling imbalance are a non-linear effect in the activation function.

\paragraph{Generalization and memorization gaps}

In this paragraph, we compute the generalization and memorization gaps of section \ref{sec:class_specific_gen_mem} for an RF model with a $\mathrm{ReLU}$ activation function and compare them with the $\tanh$ activation used in the rest of the work. We consider the three ablations of Fig.~\ref{fig:test_error_ablations} with ratios between variances and centroid factors $\sigma_1^2/\sigma_2^2=2$, $\mathrm{cf}_2/\mathrm{cf}_1$ and evaluate the gaps first for different baseline centroids $\mathrm{cf}_b\equiv \mathrm{cf}_1=1.0, 1.5, 2.0, 3.0, 5.0$ and then for different baseline variances $\sigma_b^2\equiv \sigma_1^2=0.5, 1.0, 1.5, 4.5, 12.5$. In Figs.~\ref{fig:relu_vs_tanh_gen},\ref{fig:relu_vs_tanh_mem} we show respectively the generalization and memorization log-gaps, $\log(\tau_{g}^{(2)}/\tau_{g}^{(1)})$ and $\log(\tau_{m}^{(2)}/\tau_{m}^{(1)})$, versus $\mathrm{cf}_b^2/\sigma_b^2$ (left panels) and $\sigma_b^2/\mathrm{cf}^2$ (right panels): for each of the ablations of Fig.~\ref{fig:test_error_ablations}, we see how the gaps are affected by varying the baseline centroid factor $\rm{cf}_b$ (left panels) and the baseline variance $\sigma_b^2$ (right panels). For $\mathrm{cf}_b\underset{\approx}{<} \sigma_b$, both plots confirm result \ref{res:learning_hierarchy}: variance is the dominating factor for the learning gaps, with centroid norm and sampling rates playing a secondary role. Only for $\mathrm{cf}_b\gg \sigma_b$ does the role of centroid norm become comparable to that of variance. This limit does not correspond to realistic datasets statistics.

\section{Spectral equations of $\mathbf{U}$}
\label{sec:derivation_spectral_equations}

The spectrum of the matrix $\mathbf{U}$ is fully characterized by the resolvent matrix
\begin{equation}
    \label{eq:resolvent_matrix}
    \mathbf{\mathcal{G}}_U(z)=(\mathbf{U}-z \mathbf{I}_P)^{-1}\equiv \sum_{q=1}^P\frac{\bm{\psi}_q\bm{\psi}_q^\intercal}{\lambda_q-z},\qquad z\in\Complex\setminus \operatorname{Spec}(\mathbf{U})
\end{equation}
where $\operatorname{Spec}(\cdot)$ is the set of eigenvalues of a matrix.
Since the matrix $\mathbf{U}$ is random through the dataset $\mathbf{X}$ and the random features matrix $\mathbf{W}$, the resolvent is a random matrix itself. However, its first matrix moment concentrates in the high-dimensional limit:
\begin{equation}
    \label{eq:resolvent_trace}
\begin{gathered}
    \lim_{P\rightarrow \infty}\left|\frac{1}{P}\operatorname{Tr}\mathbf{\mathcal{G}}_U(z)-\Expected{\frac{1}{P}\operatorname{Tr}\mathbf{\mathcal{G}}_U(z)}{\mathbf{X},\mathbf{W}}\right|=0
    \, \\
    \\ g_U(z)=\frac{1}{P}\operatorname{Tr}\mathbf{\mathcal{G}}_U(z)\overset{P\rightarrow\infty}{\Longrightarrow} g(z)=\int \frac{d\lambda\,\rho(\lambda)}{\lambda-z}
\end{gathered}
\end{equation}
We shall call the asymptotic limit of the resolvent first moment the \emph{resolvent function}: it is the Stieltjes transform \cite{potters2020first} of the density of eigenvalues $\rho(\lambda)$.
From the statistics of the resolvent matrix, we can evaluate the statistical properties of the eigenvalues and the eigenvectors of $\mathbf{U}$: in particular, from the resolvent function in \eqref{eq:resolvent_trace} we can compute the density of eigenvalues through the Sokhotski-Plemelj formula, also known as the inverse Stieltjes transform:
\begin{equation}
    \label{eq:pdf_eigenvalues_theor}
    \rho(\lambda)=\frac{1}{\pi}\lim_{\epsilon\rightarrow 0_+}\operatorname{Im} g(\lambda+i\epsilon)
\end{equation}
To evaluate the resolvent function, we use computational techniques of statistical physics. We consider the Edwards-Jones formula \cite{Edwards_1975}:
\begin{equation}
\label{eq:Edwards_Jones}
\begin{gathered}
    g(z)=\lim_{P\rightarrow\infty}\frac{1}{P}\Expected{\operatorname{Tr}(\mathbf{U}-z\mathbf{I}_P)^{-1}}{\mathbf{X},\mathbf{W}}=-\frac{\partial}{\partial z}\lim_{P\rightarrow\infty}\frac{1}{P}\Expected{\operatorname{Tr}\log(\mathbf{U}-z\mathbf{I}_P)}{\mathbf{X}, \mathbf{W}}  \\
    \, \\
    = -\frac{\partial}{\partial z}\lim_{P\rightarrow\infty}\frac{1}{P}\Expected{\log\det(\mathbf{U}-z\mathbf{I}_P)}{\mathbf{X}, \mathbf{W}}
    = 2\frac{\partial}{\partial z}\lim_{P\rightarrow\infty}\frac{1}{P}\Expected{\log \mathcal{Z}_U}{\mathbf{X}, \mathbf{W}}, \\
    \, \\
    \, \\
    \mathcal{Z}_U=\int d\boldsymbol{\psi}\,e^{-\frac{1}{2}\boldsymbol{\psi}^\intercal (\mathbf{U}-z\mathbf{I}_P)\boldsymbol{\psi}}.
\end{gathered}
\end{equation}

\subsection{Replica computation of the resolvent function}

The goal of a statistical physics computation is to represent high-dimensional systems in terms of few, low-dimensional parameters: in the case of random matrix calculations, these are resolvent functions of random matrices. Depending on the calculation, the final result can be characterized by several different resolvents. In particular, the main one and objective of this calculation is the resolvent function \eqref{eq:resolvent_trace}, which it is easy to see from \eqref{eq:Edwards_Jones} that it is given by
\begin{equation}
\label{eq:g_from_EJ}
    g(z)=\lim_{P\rightarrow\infty}\frac{1}{P}\Expected{\int d\boldsymbol{\psi}\,\bm{\psi}^\intercal \bm{\psi}\,e^{-\frac{1}{2}\boldsymbol{\psi}^\intercal (\mathbf{U}-z\mathbf{I}_P)\boldsymbol{\psi}}}{\mathbf{X},\mathbf{W}}.
\end{equation}

To evaluate the average over the log in \eqref{eq:Edwards_Jones}, we use the Replica trick \cite{charbonneau2023spin}:
\begin{equation}
    \label{eq:replic_trick}
    \begin{gathered}
        \log \mathcal{Z}_U=\lim_{n\rightarrow 0}\frac{\mathcal{Z}_U^n-1}{n}\Longrightarrow \Expected{\log \mathcal{Z}_U}{\mathbf{X}, \mathbf{W}}=\lim_{n\rightarrow 0}\frac{\Expected{\mathcal{Z}_U^n}{\mathbf{X}, \mathbf{W}}-1}{n} \\
        \, \\
        \text{with} \\
        \, \\
        \Expected{\mathcal{Z}_U^n}{\mathbf{X}, \mathbf{W}} = 
        \Expected{\left(\int d\boldsymbol{\psi}\,e^{-\frac{1}{2}\boldsymbol{\psi}^\intercal (\mathbf{U}-z\mathbf{I}_P)\boldsymbol{\psi}}\right)^n}{\mathbf{X}, \mathbf{W}} \\
        \equiv \Expected{\prod_{a=1}^n\int d\boldsymbol{\psi}_a\,e^{-\frac{1}{2}\boldsymbol{\psi}_a^\intercal (\mathbf{U}-z\mathbf{I}_P)\boldsymbol{\psi}_a}}{\mathbf{X}, \mathbf{W}}.
    \end{gathered}
\end{equation}
Then, we replace $\mathbf{U}$ with its GEP in \eqref{eq:GEP_U_matrix_form}:
\begin{equation*}
\begin{gathered}
    \Expected{\prod_{a=1}^n\int d\boldsymbol{\psi}_a\,e^{-\frac{1}{2}\boldsymbol{\psi}_a^\intercal (\mathbf{U}-z\mathbf{I}_P)\boldsymbol{\psi}_a}}{\mathbf{X}, \mathbf{W}}
    = \Expected{\prod_{a=1}^n\int d\boldsymbol{\psi}_a\,e^{-\frac{1}{2}\boldsymbol{\psi}_a^\intercal (\sum_{c=1}^C b_c\mathbf{U}_{c}^{(gep)}-z\mathbf{I}_P)\boldsymbol{\psi}_a}}{\{\mathbf{X}_c, \mathbf{H}_c\}_{c=1}^C, \mathbf{W}}
\end{gathered}
\end{equation*}
where we indicate with $\mathbf{H}_c\in \Real^{P\times M_c}$ the matrix having as columns the noise vectors of the GEP of the different classes.
Finally, we consider Eq.~\eqref{eq:U_gep_highlight_perturbations}: since we are mostly interested in the absolutely continuous part of the spectrum of $\mathbf{U}^{(gep)}$, we perform the computation for the matrix $\widehat{\mathbf{U}}^{(gep)}=\sum_{c=1}^C b_c \widehat{\mathbf{U}}^{(gep)}_c$, given that this matrix does not have a low-rank perturbation. Indeed, the two matrices have the same absolutely continuous spectrum, differing only by the presence of outliers in $\mathbf{U}_{gep}$.
Thus, in what follows, we proceed with the computation of the following expectation:
\begin{equation}
\label{eq:target_integral}
    \overline{\mathcal{Z}^n}=\Expected{\prod_{a=1}^n\int d\boldsymbol{\psi}_a\,e^{-\frac{1}{2}\boldsymbol{\psi}_a^\intercal (\sum_{c=1}^C b_c\widehat{\mathbf{U}}_{c}^{(gep)}-z\mathbf{I}_P)\boldsymbol{\psi}_a}}{\{\mathbf{X}_c, \mathbf{H}_c\}_{c=1}^C, \mathbf{W}}.
\end{equation}
We set $\widetilde{\bm{G}}_c^{(\nu)}=\bm{G}_c^{(\nu)}-\bm{\alpha}_c$ (compare with Eq.~\eqref{eq:GEP_U_matrix_form}) and expand:
\begin{equation*}
    \begin{gathered}
        \, \\
        \overline{\mathcal{Z}^n}=\mathbb{E}_{{\{\mathbf{X}_c, \mathbf{H}_c\}_{c=1}^C, \mathbf{W}}}\Biggl[\int d\boldsymbol{\Psi}\,\prod_{a=1}^n e^{
        -\frac{1}{2}\sum_{c=1}^C b_c\boldsymbol{\psi}_a^\intercal (\operatorname{diag}\boldsymbol{\varsigma}_c-z\mathbf{I}_P)\boldsymbol{\psi}_a
        -\frac{1}{2\,M}\boldsymbol{\psi}_a^\intercal\sum_{c=1}^C\widetilde{\mathbf{G}}_c\widetilde{\mathbf{G}}_c^\intercal \boldsymbol{\psi}_a} \\
        \, \\
        \times e^{-\frac{\Delta}{2N}\boldsymbol{\psi}_a^\intercal\left(\sum_{c=1}^C \frac{b_c}{\Gamma_c^2}\widetilde{\boldsymbol{\gamma}}_c\widetilde{\boldsymbol{\gamma}}_c^\intercal\right)\odot\,\mathbf{W}\mathbf{W}^\intercal\boldsymbol{\psi}_a} \Biggr]
    \end{gathered}
\end{equation*}
where we set $d\bm{\psi}_1\cdots d\bm{\psi}_n = d\bm{\Psi}$ and defined the vectors
\begin{equation}
    \boldsymbol{\varsigma}_c = \widetilde{\boldsymbol{\beta}}_c-\boldsymbol{\beta}_c-\frac{\Delta_t}{\sigma_c^2 e^{-2 t}}\boldsymbol{\gamma}_c^{\odot 2}.
\end{equation}
We do the disorder averages over the data, the GEP noise, and the random features step by step: to this end, it is convenient to compact the terms in the GEP features:
\begin{equation}
    \label{eq:more_compact_GEP_non_cent}
    \widetilde{G}_c^{p\nu}=\frac{\gamma_c^{(p)}}{\sigma_c\sqrt{N}}\sum_{j=1}^N\,W_{pj}(x_j^{(c,\nu)}-m_j^{(c)})+\widetilde{\eta}_{p\nu}^{(c)}\,\text{,}
    \qquad \widetilde{\boldsymbol{\eta}}_{\nu}^{(c)}\sim\mathcal{N}(\cdot\,;\,\bm{0},\operatorname{diag}(\boldsymbol{h}_c^{\odot 2}))
\end{equation}
where we set $\boldsymbol{h}_c^{\odot 2}=\boldsymbol{\beta}_c-\boldsymbol{\alpha}_c^{\odot2}-\boldsymbol{\gamma}_c^{\odot 2}$.
We then group terms in the GEP covariances in terms of their disorder dependence:
\begin{align*}
&\frac{1}{M}\big(\widetilde{\bm{G}}_c \widetilde{\bm{G}}_c^{\intercal}\big)_{pq}
= 
\Biggl\{\frac{\gamma_{c}^{(p)}\,\gamma_{c}^{(q)}}{M\,\sigma_c^{2}\,N}
    \sum_{i=1}^{N}\sum_{j=1}^N\sum_{\nu=1}^{M_c}
      W_{p i}\,W_{q j}\;
      \,
      \big(x_i^{(c,\nu)}-m_i^{(c)}\big)\,\big(x_j^{(c,\nu)}-m_j^{(c)}\big)
      \\[6pt]
&
+ \left[
    \frac{\gamma_{c}^{(p)}}{M\sqrt{N}\,\sigma_c}
      \sum_{i=1}^{N} W_{q i}\,\sum_{\nu=1}^{M_c}\big(x_i^{(c,\nu)}-m_i^{(c)}\big)\widetilde{\eta}_{q\nu}^{(c)}+\frac{\gamma_{c}^{(q)}}{M\sqrt{N}\,\sigma_c}
      \sum_{i=1}^{N} W_{p i}\,\sum_{\nu=1}^{M_c}\big(x_i^{(c,\nu)}-m_i^{(c)}\big)\widetilde{\eta}_{p\nu}^{(c)}
    \right]\Biggr\}
\\[6pt]
&
+ \Biggl\{\frac{1}{M}
    \sum_{v=1}^{M_c} \widetilde{\eta}^{(c)}_{p\nu}\, \widetilde{\eta}^{(c)}_{q\nu}\Biggr\}\equiv A_{pq}^{(X_c, \widetilde{H}_c, W)}+A_{pq}^{(\widetilde{H}_c, W)}.
\end{align*}
Note that each of the vector GEP coefficients $\boldsymbol{\alpha}_c, \boldsymbol{\beta}_c, \boldsymbol{\gamma}_c, \boldsymbol{\widetilde{\beta}}_c$ depends on the random features $\mathbf{W}$ through the projected centroids $\boldsymbol{\mu}_c\equiv (1/\sqrt{N})\mathbf{W}\mathbf{m}_c$; see their definitions in Appendix~\ref{sec:GEP_feature_correlation_matrices}.
We now proceed with the computations of the disorder averages.
\paragraph{Average over the data.} 
The average over the data $\{\bm{X}_c\}_{c=1}^C$ can be factorized across the different classes' data. The generic $c$ integral is a standard multi-variate integral
\begin{equation*}
\begin{gathered}
    \Expected{e^{-\frac{1}{2}\sum_{a}\boldsymbol{\psi}_a^\intercal\mathbf{A}^{(X_c, \widetilde{H}_c, W)}\boldsymbol{\psi}_a}}{\mathbf{X}_c}\,\,= \\
    \, \\
    \prod_{\nu=1}^{M_c}
    \int
    \frac{d \bm{x}^{(c)}_{\nu}}{(2\pi \sigma_c^{2})^{N/2}}
    \;
    e^{
    -\,\frac{1}{2\sigma_c^{2}}\,
    \big(\bm{x}^{(c)}_{\nu}-\mathbf{m}_{c}\big)^{\intercal}
    \Big(\mathbf{I}_{N} + \frac{1}{N M}\,\mathbf{W}^\intercal\sum_a (\boldsymbol{\gamma}_c\odot \boldsymbol{\psi}_a)(\boldsymbol{\gamma}_c\odot \boldsymbol{\psi}_a)^\intercal \mathbf{W}\Big)
    \big(\bm{x}^{(c)}_{\nu}-\bm{m}_{c}\big)
    } \\
    \, \\
    \times e^{\frac{1}{\sqrt{N}\sigma_c M}\sum_a \boldsymbol{\psi}_a^\intercal \widetilde{\boldsymbol{\eta}}_{\nu}^{(c)}[(\boldsymbol{\psi}_a\odot \boldsymbol{\gamma}_c)^\intercal\mathbf{W}](\bm{x}_\nu^{(c)}-\mathbf{m}_c)} \\
    \, \\
    = e^{\frac{1}{2 N M^2}\sum_{\nu=1}^{M_c}\sum_{a b}(\boldsymbol{\psi}_a^\intercal \widetilde{\boldsymbol{\eta}}_{\nu}^{(c)})(\boldsymbol{\psi}_b^\intercal \widetilde{\boldsymbol{\eta}}_{\nu}^{(c)})(\mathbf{W}^\intercal \boldsymbol{\psi}_a\odot \boldsymbol{\gamma}_c)^\intercal(\mathbf{I}_N+\frac{1}{N M}\boldsymbol{J}_{c})^{-1}(\mathbf{W}^\intercal \boldsymbol{\psi}_b\odot \boldsymbol{\gamma}_c)-\frac{M_c}{2}\log\det(\mathbf{I}_N+\frac{1}{N M}\boldsymbol{J}_{c})}, \\
    \, \\
    \mathbf{J}_c\equiv \mathbf{W}^\intercal\sum_a (\boldsymbol{\gamma}_c\odot \boldsymbol{\psi}_a)(\boldsymbol{\gamma}_c\odot \boldsymbol{\psi}_a)^\intercal \mathbf{W}.
\end{gathered}
\end{equation*}
where we recall that $\chi_p \equiv P/N$. To make the $\mathbf{J}_c$ independent of the fields $\{\bm{\psi}_a\}$, we introduce the auxiliary fields $\bm{\omega}_{a; c}=\frac{1}{\sqrt{P}}\mathbf{W}^\intercal (\boldsymbol{\psi}_a\odot \boldsymbol{\gamma}_c))$ through delta functions
\begin{equation}
\label{eq:omega}
\begin{gathered}
    1=\int \prod_{c=1}^C d\bm{\Omega}_c \delta(\sqrt{P}\boldsymbol{\omega}_{a; c}-\mathbf{W}^\intercal (\boldsymbol{\psi}_a\odot \boldsymbol{\gamma}_c))
    \\
    \\
    =\int \prod_{c=1}^C d\bm{\Omega}_c \frac{d\widehat{\bm{\Omega}}_c}{(2\pi)^{n N}}e^{i\sqrt{P}\boldsymbol{\omega}_{a; c}^\intercal\widehat{\bm{\omega}}_{a; c}-i\widehat{\bm{\omega}}_{a; c}^\intercal\mathbf{W}^\intercal (\boldsymbol{\psi}_a\odot \boldsymbol{\gamma}_c)}
\end{gathered}
\end{equation}
where we set $d\bm{\Omega}_c=d\bm{\omega}_{1,c}\cdots d\bm{\omega}_{n,c}$ and $d\widehat{\bm{\Omega}}_c=d\widehat{\bm{\omega}}_{1,c}\cdots d\widehat{\bm{\omega}}_{n,c}$.
After averaging over the data, Eq.~\eqref{eq:target_integral} becomes
\begin{equation*}
    \begin{gathered}
        \, \\
        \mathbb{E}_{{\{\mathbf{X}_c, \mathbf{H}_c\}_{c=1}^C, \mathbf{W}}}
        \Biggl[\int d\boldsymbol{\Psi}\,\prod_{a=1}^n 
        e^{
        -\frac{1}{2}\sum_{c=1}^C b_c\boldsymbol{\psi}_a^\intercal (\operatorname{diag}\boldsymbol{\varsigma}_c-z\mathbf{I}_P)\boldsymbol{\psi}_a
        -\frac{1}{2\,M}\boldsymbol{\psi}_a^\intercal\sum_{c=1}^C\widetilde{\mathbf{G}}_c\widetilde{\mathbf{G}}_c^\intercal \boldsymbol{\psi}_a
        } \\
        \, \\
        \times e^{
        -\frac{\Delta}{2N}\boldsymbol{\psi}_a^\intercal\left(\sum_{c=1}^C \frac{b_c}{\Gamma_c^2}\widetilde{\boldsymbol{\gamma}}_c\widetilde{\boldsymbol{\gamma}}_c^\intercal\right)\odot\,\mathbf{W}\mathbf{W}^\intercal\boldsymbol{\psi}_a
        }\Biggr] \\
        \, \\
        =\mathbb{E}_{\{\mathbf{H}_c\}_{c=1}^C, \mathbf{W}}\Biggl[
        \int d\boldsymbol{\Psi}\int \prod_{c} d\boldsymbol{\Omega}_c \frac{d\widehat{\boldsymbol{\Omega}}_c}{(2\pi)^{N n}}
        e^{
        i\sqrt{P}\sum_{c} \sum_{a}\boldsymbol{\omega}_{a; c}^\intercal\widehat{\bm{\omega}}_{a; c}
        -\frac{1}{2}\sum_{c} b_c\sum_{a}\boldsymbol{\psi}_a^\intercal (\operatorname{diag}\boldsymbol{\varsigma}_c-z\mathbf{I}_P)\boldsymbol{\psi}_a
        } \\
        \, \\
        \times e^{
        -\frac{P \Delta}{2 N e^{-2t}}\sum_c \sum_a \frac{b_c}{\sigma_c^2}\boldsymbol{\omega}_{a; c}^\intercal \boldsymbol{\omega}_{a; c}
        +\frac{\chi_p}{2 M^2}\sum_c\sum_{\nu=1}^{M_c}\sum_{a b}(\boldsymbol{\psi}_a^\intercal \widetilde{\boldsymbol{\eta}}_{\nu}^{(c)})(\boldsymbol{\psi}_b^\intercal \widetilde{\boldsymbol{\eta}}_{\nu}^{(c)})\bm{\omega}_{a; c}^\intercal(\mathbf{I}_N+\frac{\chi_p}{ M}\boldsymbol{J}_{c})^{-1}\bm{\omega}_{a; c}
        } \\
        \, \\
        \times e^{
        -\frac{M}{2}\sum_c b_c\log\det(\mathbf{I}_N+\frac{\chi_p}{M}\boldsymbol{J}_{c})
        -i\sum_c\sum_a\widehat{\bm{\omega}}_{a; c}^\intercal\mathbf{W}^\intercal (\boldsymbol{\psi}_a\odot \boldsymbol{\gamma}_c))
        -\frac{1}{2 M}\sum_c\sum_{a}\boldsymbol{\psi}_{a}^{\intercal} \widetilde{\boldsymbol{\eta}}_{\nu}^{(c)} (\widetilde{\boldsymbol{\eta}}_{\nu}^{(c)})^\intercal \boldsymbol{\psi}_{a}
        }
        \Biggr].
    \end{gathered}
\end{equation*}
\paragraph{Average over the GEP noise.} Again, the average over the GEP noise $\{\mathbf{\widetilde{H}}_c\}_{c=1}^C$ can be factorized into the different classes
\begin{equation*}
\begin{gathered}
\mathbb{E}_{\mathbf{\widetilde{H}}_c}\Biggl[e^{- \tfrac{1}{2}\sum_{a}\boldsymbol{\psi}_{a}^{\intercal} \mathbf{A}^{(\widetilde{H}_c,W)} \boldsymbol{\psi}_{a}
+
\frac{\chi_p}{2 M^2}\sum_{\nu=1}^{M_c}\sum_{a b}
\big(\widetilde{\boldsymbol{\eta}}_{\nu}^{(c)}\big)^{\intercal}\boldsymbol{\psi}_a\,\boldsymbol{\psi}_b^{\intercal}
\widetilde{\boldsymbol{\eta}}_{\nu}^{(c)}\,\boldsymbol{\omega}_{a; c}^\intercal[\mathbf{I}_N+\frac{\chi_p}{M}\boldsymbol{J}_{c}]^{-1}\boldsymbol{\omega}_{b, c}}\Biggr] = \\
\, \\
\prod_{\nu=1}^{M_c}\int \frac{d\widetilde{\boldsymbol{\eta}}_{\nu}^{(c)}\,e^{-\frac{1}{2 }(\widetilde{\boldsymbol{\eta}}_{\nu}^c)^\intercal \operatorname{diag}(\boldsymbol{h}_c^{\odot 2})^{-1} \widetilde{\boldsymbol{\eta}}_{\nu}^c
}}{(2 \pi)^{P/2}\sqrt{\det(\operatorname{diag}(\boldsymbol{h}_c^{\odot 2}))}}
e^{-\frac{1}{2}
\big(\widetilde{\boldsymbol{\eta}}_{\nu}^{(c)}\big)^{\intercal}\sum_{a b}\left[\frac{\delta_{a b}}{M}\bm{\psi}_a\bm{\psi}_a^\intercal-\frac{\chi_p}{2 M^2}\,\boldsymbol{\psi}_a\,\boldsymbol{\omega}_{a; c}^\intercal[\mathbf{I}_N+\frac{\chi_p}{S}\boldsymbol{J}_{c}]^{-1}\boldsymbol{\omega}_{b, c}
\boldsymbol{\psi}_b^{\intercal}\right]
\widetilde{\boldsymbol{\eta}}_{\nu}^{(c)}\,} = \\
\, \\
e^{-\frac{M_c}{2}\log\det\left(\mathbf{I}_P+\frac{1}{M}\operatorname{diag}(\boldsymbol{h}_c)\boldsymbol{\mathcal{J}}_c\operatorname{diag}(\boldsymbol{h}_c)\right)} \\
\, \\
\text{where} \\
\, \\
\bm{\mathcal{J}}_c(\boldsymbol{\psi}, \boldsymbol{\omega}_c)=\sum_{ab}\boldsymbol{\psi}_a\boldsymbol{\psi}_b^\intercal\left(\delta_{ab}-\frac{\chi_p}{M}\boldsymbol{\omega}_{a; c}^{\intercal}\left(\boldsymbol{I}_N+\frac{\chi_p}{M}\boldsymbol{J}_c'(\boldsymbol{\omega}_c)\right)^{-1}\boldsymbol{\omega}_{b, c}\right).
\end{gathered}
\end{equation*}
where we set $\boldsymbol{J}_c' = N\boldsymbol{J}_c$.
We introduce the projected class centroids $\bm{\mu}_c = \frac{1}{\sqrt{N}}\mathbf{W}\bm{m}_c$ with a Dirac delta:
\begin{equation}
    \label{eq:class_centroids}
    1=\int \prod_{c=1}^C d\bm{\mu}_c \delta\left(\bm{\mu}_c-\frac{1}{\sqrt{N}}\mathbf{W}\bm{m}_c\right)
    = \int\prod_{c=1}^C d\bm{\mu}_c \frac{d\bm{\mu}_c}{(2\pi)^P}e^{i \widehat{\bm{\mu}}_c^\intercal \bm{\mu}_c-\frac{i}{\sqrt{N}}\widehat{\bm{\mu}}_c^\intercal\mathbf{W}\bm{m}_c}
\end{equation}
After averaging over the data and the GEP noise, Eq.~\eqref{eq:target_integral} becomes
\begin{equation*}
    \begin{gathered}
        \, \\
        \mathbb{E}_{{\{\mathbf{X}_c, \mathbf{H}_c\}_{c=1}^C, \mathbf{W}}}
        \Biggl[\int d\boldsymbol{\Psi}\,\prod_{a=1}^n 
        e^{
        -\frac{1}{2}\sum_{c=1}^C b_c\boldsymbol{\psi}_a^\intercal (\operatorname{diag}\boldsymbol{\varsigma}_c-z\mathbf{I}_P)\boldsymbol{\psi}_a
        -\frac{1}{2\,M}\boldsymbol{\psi}_a^\intercal\sum_{c=1}^C\widetilde{\mathbf{G}}_c\widetilde{\mathbf{G}}_c^\intercal \boldsymbol{\psi}_a
        } \\
        \, \\
        \times e^{
        -\frac{\Delta}{2N}\boldsymbol{\psi}_a^\intercal\left(\sum_{c=1}^C \frac{b_c}{\Gamma_c^2}\widetilde{\boldsymbol{\gamma}}_c\widetilde{\boldsymbol{\gamma}}_c^\intercal\right)\odot\,\mathbf{W}\mathbf{W}^\intercal\boldsymbol{\psi}_a
        }\Biggr] \\
        \, \\
        =\mathbb{E}_{\mathbf{W}}\Biggl[
        \int d\boldsymbol{\Psi}\int \prod_{c} d\boldsymbol{\Omega}_c  \frac{d\widehat{\boldsymbol{\Omega}}_c}{(2\pi)^{n N}}d\bm{\mu}_c \frac{d\bm{\mu}_c}{(2\pi)^P}
        e^{
        i\sqrt{P}\sum_{c} \sum_{a}\boldsymbol{\omega}_{a; c}^\intercal\widehat{\bm{\omega}}_{a; c}
        +i\sum_c \widehat{\bm{\mu}}_c^\intercal \bm{\mu}_c
        } \\
        \, \\
        \times e^{
        -\frac{1}{2}\sum_{c} b_c\sum_{a}\boldsymbol{\psi}_a^\intercal (\operatorname{diag}\boldsymbol{\varsigma}_c-z\mathbf{I}_P)\boldsymbol{\psi}_a
        -\frac{P \Delta}{2 N e^{-2t}}\sum_c \sum_a \frac{b_c}{\sigma_c^2}\boldsymbol{\omega}_{a; c}^\intercal \boldsymbol{\omega}_{a; c}
        -\frac{M}{2}\sum_c b_c\log\det\left(\mathbf{I}_P+\frac{1}{M}\operatorname{diag}(\boldsymbol{h}_c)\boldsymbol{\mathcal{J}}_c\operatorname{diag}(\boldsymbol{h}_c)\right)
        } \\
        \, \\
        \times e^{
        -\frac{M}{2}\sum_c b_c\log\det(\mathbf{I}_N+\frac{\chi_p}{M}\boldsymbol{J}_{c})
        -i\sum_c\sum_a\widehat{\bm{\omega}}_{a; c}^\intercal\mathbf{W}^\intercal (\boldsymbol{\psi}_a\odot \boldsymbol{\gamma}_c)
        -\frac{i}{\sqrt{N}}\sum_c\widehat{\bm{\mu}}_c^\intercal\mathbf{W}\bm{m}_c
        }
        \Biggr].
    \end{gathered}
\end{equation*}
\paragraph{Average over the random features.} The average over the random features $\mathbf{W}$ is straightforward:
\begin{equation}
\label{eq:avg_random_feat}
    \begin{gathered}
        \Expected{e^{-i\sum_{c}\left[\frac{1}{\sqrt{N}}\widehat{\boldsymbol{\mu}}_c^\intercal \boldsymbol{W}\boldsymbol{m}_c+\sum_a (\boldsymbol{\gamma}_c\odot \boldsymbol{\psi}_a)^\intercal \boldsymbol{W}\widehat{\boldsymbol{\omega}}_{a; c}\right]}}{W} = \\
        \, \\
        \int \prod_{p j}\frac{dW_{p j}}{\sqrt{2 \pi}}e^{-\frac{1}{2}\sum_{p j}W_{p j}^2
        -i\sum_{c}\left[\frac{1}{\sqrt{N}}\widehat{\boldsymbol{\mu}}_c^\intercal \boldsymbol{W}\boldsymbol{m}_c+\sum_a (\boldsymbol{\gamma}_c\odot \boldsymbol{\psi}_a)^\intercal \boldsymbol{W}\widehat{\boldsymbol{\omega}}_{a; c}\right]
        }  = \\
        \, \\
        e^{-\frac{1}{2 N}\sum_{c,c'}\boldsymbol{m}_c^\intercal \boldsymbol{m}_{c'}
        \, \widehat{\boldsymbol{\mu}}_c^\intercal \widehat{\boldsymbol{\mu}}_{c'}-\frac{1}{2}\sum_{c, c'} \sum_{a b} (\boldsymbol{\gamma}_c\odot \boldsymbol{\psi}_a)^\intercal (\boldsymbol{\gamma}_{c'}\odot \boldsymbol{\psi}_b)\, \widehat{\boldsymbol{\omega}}_{a; c}^\intercal\widehat{\boldsymbol{\omega}}_{b, c'}
        +o_P(P) 
        }
    \end{gathered}
\end{equation}
The term that is $o_P(P)$ is $-\frac{1}{\sqrt{N}}\sum_{c, c'}\sum_a (\widehat{\boldsymbol{\omega}}_{a; c}^\intercal \boldsymbol{m}_c)\,(\boldsymbol{\gamma}_c\odot \boldsymbol{\psi}_a)^\intercal \widehat{\boldsymbol{\mu}}_{c'}$: the first dot yields an $O_P(\sqrt{P})$ contribution, the second an $\mathcal{O}_P(1)$ one, given that $\lVert \widehat{\bm{\omega}}_{a; c} \lVert = \mathcal{O}_P(1)$; then, the overall scaling of the last term is $\mathcal{O}_P(1)$, and thus can be neglected, since the first two terms on the right-hand side of \eqref{eq:avg_random_feat} are $\mathcal{O}_P(P)$.

The average of the replicated partition function, after the three disorder averages over data, GEP noise and random features, reads

\begin{equation}
    \begin{gathered}
        \overline{\mathcal{Z}^n}=
        \int 
        \prod_{c} \left(d\boldsymbol{\mu}_c\,\frac{d\widehat{\boldsymbol{\mu}}_c}{(2\pi)^P} \right)
        \,e^{-\frac{1}{2 N}\sum_{c, c'} \boldsymbol{m}_c^\intercal \boldsymbol{m}_{c'}\, \widehat{\boldsymbol{\mu}}_c^\intercal \widehat{\boldsymbol{\mu}}_{c'}+i \sum_{c} \widehat{\boldsymbol{\mu}}_c^\intercal \boldsymbol{\mu}_{c}}
         \\
        \, \\
        \times \int
        \prod_{c} \left(d\boldsymbol{\Omega}_c\,\frac{d\widehat{\boldsymbol{\Omega}}_c}{(2\pi)^{n N}} \right) e^{-\frac{P \Delta}{2 N e^{-2t}}\sum_c \sum_a \frac{b_c}{\sigma_c^2}\boldsymbol{\omega}_{a; c}^\intercal \boldsymbol{\omega}_{a; c}+i\sqrt{P} \sum_c\sum_a \widehat{\boldsymbol{\omega}}_{a; c}^\intercal \boldsymbol{\omega}_{a; c} 
        }
         \\
        \, \\
        \times\int d\boldsymbol{\Psi} e^{-\frac{1}{2}\sum_c b_c \sum_a \boldsymbol{\psi}_a^\intercal (\operatorname{diag}(\boldsymbol{\varsigma}_c)-z\boldsymbol{I}_P) \boldsymbol{\psi}_a}
        e^{-\frac{M}{2}\sum_c b_c \log \det (\boldsymbol{I}_N
        +\frac{\chi_p}{M}\boldsymbol{J}_c')
        } \\
        \, \\
        \times e^{
        -\frac{M}{2}\sum_c b_c \log\det\left(I_P+\frac{1}{M}\operatorname{diag}(\boldsymbol{h}_c)\boldsymbol{\mathcal{J}}_c\operatorname{diag}(\boldsymbol{h}_c)\right)
        -\frac{1}{2}\sum_{c, c'} \sum_{a b} (\boldsymbol{\gamma}_c\odot \boldsymbol{\psi}_a)^\intercal (\boldsymbol{\gamma}_{c'}\odot \boldsymbol{\psi}_b)\, \widehat{\boldsymbol{\omega}}_{a; c}^\intercal\widehat{\boldsymbol{\omega}}_{b, c'}
        }
    \end{gathered}
\end{equation}
\paragraph{Average over the auxiliary fields.} We can perform the Gaussian integrations with respect to the $\widehat{\bm{\omega}}_{a; c}$ and the $\widehat{\bm{\mu}}_c$: we start with the integral over the $\widehat{\boldsymbol{\omega}}_{a; c}$:
\begin{equation}
    \begin{gathered}
        \int \frac{d\widehat{\mathbf{\Omega}}_c}{(2\pi)^{C n N}}\,
        e^{
        -\frac{1}{2}\sum_{c c'}\sum_{a b}K_{a b}^{c c'}\widehat{\boldsymbol{\omega}}_{a;c}^\intercal \widehat{\boldsymbol{\omega}}_{b;c'}-\sum_a \sum_c
        i \boldsymbol{\omega}_{a; c}^\intercal \widehat{\boldsymbol{\omega}}_{a; c}
        } = \\
        \, \\
        =
        \frac{1}{(2\pi)^{C n N/2}}
        e^{
        -\frac{1}{2}\sum_{c c'}\sum_{a b}(\mathbf{K}^{-1})_{a b}^{c c'}\boldsymbol{\omega}_{a;c}^\intercal \boldsymbol{\omega}_{b;c'}
        -\frac{N}{2}\log \det \mathbf{K}
        }, \\
        \, \\
        \, \\
        K_{a b}^{c c'}=\frac{1}{P}\operatorname{Tr}
        \left(
        \boldsymbol{\gamma}_c\boldsymbol{\gamma}_{c'}^{\intercal}
        \odot
        \boldsymbol{\psi}_a
        \boldsymbol{\psi}_b^\intercal
        \right).
    \end{gathered}
\end{equation}
Then, the integration over the $\widehat{\boldsymbol{\mu}}_c$ yields the probability density function (PDF) of the $\boldsymbol{\mu}_c$
\begin{equation}
\begin{gathered}
    \int \frac{d\widehat{\boldsymbol{\mu}_c}}{(2\pi)^{P/2}}
    e^{
    -\frac{1}{2}\sum_{c c'}r_{c c'}\widehat{\boldsymbol{\mu}}_c^\intercal
    \widehat{\boldsymbol{\mu}}_{c'}+i\sum_{c}\widehat{\boldsymbol{\mu}}_c^\intercal\boldsymbol{\mu}_c
    }
    =
    \frac{
    e^{
    -\frac{1}{2}\sum_{c c'}
    (\mathbf{r}^{-1})_{c c'}
    \boldsymbol{\mu}_c^\intercal
    \boldsymbol{\mu}_{c'}
    }}{(2\pi \det \mathbf{r})^{P/2}}, \\
    \, \\
    \, \\
    r_{c c'}=\frac{1}{N}\mathbf{m}_c^\intercal \mathbf{m}_{c'}.
\end{gathered}
\end{equation}

Thus, after these integrations the partition function reads

\begin{equation}
\label{eq:pluppa}
    \begin{gathered}
        \overline{\mathcal{Z}^n}=
        \int 
        \prod_{c} \frac{d\boldsymbol{\mu}_c}{(2\pi)^{P/2}}
        \,e^{-\frac{1}{2}\sum_{c, c'} (\mathbf{r}^{-1})_{c c'}\, \boldsymbol{\mu}_c^\intercal \boldsymbol{\mu}_{c'}
        -\frac{1}{2}\log\det \mathbf{r}
        } \\
        \times
        \int
        \prod_{c} d\boldsymbol{\Omega}_c e^{-\frac{\chi_p \Delta}{2 e^{-2t}}\sum_c \sum_a \frac{b_c}{\sigma_c^2}\boldsymbol{\omega}_{a; c}^\intercal \boldsymbol{\omega}_{a; c}
        } \\
        \, \\
        \times 
        \int d\boldsymbol{\Psi} e^{-\frac{1}{2}\sum_c b_c \sum_a \boldsymbol{\psi}_a^\intercal (\operatorname{diag}(\boldsymbol{\varsigma}_c)-z\boldsymbol{I}_P) \boldsymbol{\psi}_a
        -\frac{1}{2}\sum_{a b}\sum_{c c'}(\mathbf{K}(\boldsymbol{\Psi})^{-1})_{a b}^{c c'}\boldsymbol{\omega}_{a; c}^\intercal \boldsymbol{\omega}_{b; c'}  
        }
        \\
        \, \\
        \times e^{
        -\frac{N}{2}\log \det \mathbf{K}(\boldsymbol{\Psi})
        -\frac{M}{2}\sum_c b_c \log \det (\mathbf{I}_N+\frac{\chi_p}{M}\boldsymbol{J}_c')
        -\frac{M}{2}\sum_c b_c \log\det\left(\mathbf{I}_P+\frac{1}{M}\operatorname{diag}(\boldsymbol{h}_c)\boldsymbol{\mathcal{J}}_c\operatorname{diag}(\boldsymbol{h}_c)\right).
        }
    \end{gathered}
\end{equation}
Note that all the dependence on the projected centroids $\boldsymbol{\mu}_c$ is in the GEP Hermite coefficients, which features in the quantities $\bm{\varsigma}_c, \bm{h}_c, \mathbf{J}_c, \mathbf{\mathcal{J}}_c$.

\paragraph{The replicated order parameters.} We now introduce the order parameters of the problem: they are the following resolvent functions
\begin{equation}
\label{eq:gs_order_parameters}
\begin{gathered}
    g_{a b}^{(\Psi)}=\frac{1}{P}\operatorname{Tr}\bm{\psi}_a\bm{\psi}_b^\intercal\qquad g_{a; c}^{(\Psi, \varsigma)}=\frac{1}{P}\operatorname{Tr}\operatorname{diag}(\boldsymbol{\varsigma}_c)\bm{\psi}_a\bm{\psi}_a^\intercal\qquad g_{a b; c c'}^{(\Psi, \gamma)}=\frac{1}{P}\operatorname{Tr}\boldsymbol{\gamma}_c\boldsymbol{\gamma}_{c'}^\intercal \odot \bm{\psi}_a\bm{\psi}_b^\intercal \\
    \, \\
    g_{a; c}^{(\Psi, h)}=\frac{1}{P}\operatorname{Tr}\boldsymbol{h}_c \boldsymbol{h}_{c}^\intercal\odot\bm{\psi}_a\bm{\psi}_a^\intercal\qquad
    \qquad g_{a b; c c'}^{(\Omega)}=\frac{1}{N}\operatorname{Tr}\bm{\omega}_{a; c}\bm{\omega}_{b, c'}^\intercal.
\end{gathered}
\end{equation}
We enforce them in \eqref{eq:pluppa} through delta functions: 
\begin{equation*}
    1=\int d\{g_{ab}^{(\Psi)}\}\frac{d\{\widehat{g}_{a b}^{(\Psi)}\}}{(2\pi)^{n^2}}e^
    {
    i P\sum_{a b}\widehat{g}_{a b}^{(\Psi)}g_{a b}^{(\Psi)}
    -i\sum_{a b}\widehat{g}_{a b}^{(\Psi)}\boldsymbol{\psi}_a^\intercal \boldsymbol{\psi}_b
    }
\end{equation*}
\begin{equation*}
    1=\int d\{g_{a; c}^{(\Psi, \varsigma)}\}\frac{d\{\widehat{g}_{a; c}^{(\Psi, \varsigma)}\}}{(2\pi)^{n C}}e^
    {
    i P\sum_{a}\sum_c\widehat{g}_{a; c}^{(\Psi, \varsigma)}g_{a; c}^{(\Psi, \varsigma)}
    -i\sum_{a}\sum_c\widehat{g}_{a; c}^{(\Psi, \varsigma)}\boldsymbol{\psi}_a^\intercal\operatorname{diag}(\bm{\varsigma}_c)\boldsymbol{\psi}_a
    }
\end{equation*}
\begin{equation*}
    1=\int d\{g_{a b; c c'}^{(\Psi, \gamma)}\}\frac{d\{\widehat{g}_{a b,c c'}^{(\Psi, \gamma)}\}}{(2\pi)^{n^2 C^2}}e^
    {
    i P\sum_{a b}\sum_{c c'}\widehat{g}_{a b; c c'}^{(\Psi, \gamma)}g_{a b; c c'}^{(\Psi, \gamma)}
    -i\sum_{a b}\sum_{c c'}\widehat{g}_{a b; c c'}^{(\Psi, \gamma)}\boldsymbol{\psi}_a^\intercal\boldsymbol{\gamma}_c\boldsymbol{\gamma}_{c'}^\intercal\boldsymbol{\psi}_b
    }
\end{equation*}
\begin{equation*}
    1=\int d\{g_{a; c}^{(\Psi, h)}\}\frac{d\{\widehat{g}_{a; c}^{(\Psi, h)}\}}{(2\pi)^{n C}}e^
    {
    i P\sum_{a}\sum_c\widehat{g}_{a; c}^{(\Psi, h)}g_{a; c}^{(\Psi, h)}
    -i\sum_{a}\sum_c\widehat{g}_{a; c}^{(\Psi, h)}\boldsymbol{\psi}_a^\intercal\boldsymbol{h}_c\boldsymbol{h}_{c}^\intercal\boldsymbol{\psi}_a
    }
\end{equation*}
\begin{equation*}
    1=\int d\{g_{a b; c c'}^{(\Psi, \Omega)}\}\frac{d\{\widehat{g}_{a b,c c'}^{(\Psi, \Omega)}\}}{(2\pi)^{n^2 C^2}}e^
    {
    i N\sum_{a b}\sum_{c c'}\widehat{g}_{a b; c c'}^{(\Psi, \Omega)}g_{a b; c c'}^{(\Psi, \Omega)}
    -i\sum_{a b}\sum_{c c'}\widehat{g}_{a b; c c'}^{(\Psi, \Omega)}\boldsymbol{\omega}_{a; c}^\intercal \boldsymbol{\omega}_{b, c'}
    }
\end{equation*}
\;\\
Then, we can integrate over the $\bm{\Psi}$ and the $\mathbf{\Omega}_c$ \footnote{We ignore factors that are not exponential in system size.}: 
\;\\
\begin{equation*}
\begin{gathered}
    \int d\boldsymbol{\Psi}e^{
    -i\sum_{a b}\boldsymbol{\psi}_a^\intercal 
    \left[
    \widehat{g}_{a b}^{(\Psi)}\mathbf{I}_P
    +\delta_{a b} \sum_c \widehat{g}_{a; c}^{(\Psi, \varsigma)}\operatorname{diag}(\bm{\varsigma}_c)
    +\sum_{c c'}\widehat{g}_{a b; c c'}^{(\Psi, \gamma)}\operatorname{diag}(\boldsymbol{\gamma}_c\odot\boldsymbol{\gamma}_{c'})
    +\delta_{a b}\sum_c \widehat{g}_{a; c}^{(\Psi, h)}\operatorname{diag}(\boldsymbol{h}_{c}^{\odot 2})
    \right]
    \boldsymbol{\psi}_b
    }\\
    \\
    \propto
    e
    ^{
    -\frac{1}{2}\log\det \left(
    i\widehat{\mathbf{G}}^{(\Psi)}\otimes\mathbf{I}_P
    +\sum_c i\widehat{\bm{G}}_{c}^{(\Psi, \varsigma)}\otimes\,\operatorname{diag}(\bm{\varsigma}_c)
    +\sum_{c c'}i\widehat{\bm{G}}_{c c'}^{(\Psi, \gamma)}\otimes\,\operatorname{diag}(\boldsymbol{\gamma}_c\odot\boldsymbol{\gamma}_{c'})
    +\sum_c i\widehat{\bm{G}}_{c}^{(\Psi, h)}\otimes\,\operatorname{diag}(\boldsymbol{h}_{c}^{\odot 2})
    \right)
    }
\end{gathered}
\end{equation*}
\\
\begin{equation*}
    \int d\{\boldsymbol{\Omega}_c\}\,e^{-i\sum_{a b}\sum_{c c'}\widehat{g}_{a b; c c'}^{(\Omega)}\boldsymbol{\omega}_{a; b}^\intercal\boldsymbol{\omega}_{b; c'}}\propto
    e
    ^{
    -\frac{N}{2}\log\det \left(
    i\widehat{\boldsymbol{G}}^{(\Omega)}
    \right)
    }
\end{equation*}
\paragraph{The Replica-Symmetric ansatz and interpretation of the order parameters.} We impose a Replica-Symmetric (RS) ansatz \footnote{Random matrix theories are notoriously RS stable \cite{potters2020first}. The RS diagonal ansatz for the replicas is usually used ab initio, but it can be easily verified starting from a more generic RS ansatz.}
\begin{equation}
    g_{a b}^{(\Psi)}=
    \begin{cases}
        g^{(\Psi)}\qquad a=b
        \\
        \, \\
        0\qquad a\neq b
    \end{cases}
    \qquad
    \widehat{g}_{a b}^{(\Psi)}=
    \begin{cases}
        \widehat{g}^{(\Psi)}\qquad a=b
        \\
        \, \\
        0\qquad a\neq b
    \end{cases}
\end{equation}
\\
\begin{equation}
    g_{a;c}^{(\Psi, \varsigma)}=
    g_c^{(\Psi,\varsigma)}   
    \qquad
    \widehat{g}_{a; c}^{(\Psi,\varsigma)}=
    \widehat{g}_{c}^{(\Psi,\varsigma)}  
\end{equation}
\\
\begin{equation}
    g_{a b;c c'}^{(\Psi, \gamma)}=
    \begin{cases}
        g_{c c'}^{(\Psi,\gamma)}\qquad a=b
        \\
        \, \\
        0\qquad a\neq b
    \end{cases}
    \qquad
    \widehat{g}_{a b; c c'}^{(\Psi, \gamma)}=
    \begin{cases}
        \widehat{g}_{c c'}^{(\Psi,\gamma)}\qquad a=b
        \\
        \, \\
        0\qquad a\neq b
    \end{cases}
\end{equation}
\\
\begin{equation}
    g_{a;c}^{(\Psi, h)}=
    g_c^{(\Psi,h)}
    \qquad
    \widehat{g}_{a; c}^{(h)}=
    \widehat{g}_{c}^{(\Psi, h)}
\end{equation}
\\
\begin{equation}
    g_{a b; c c'}^{(\Omega)}=
    \begin{cases}
        g^{(\Omega)}_{c c'}\qquad a=b
        \\
        \, \\
        0\qquad a\neq b
    \end{cases}
    \qquad
    \widehat{g}_{a b; c c'}^{(\Omega)}=
    \begin{cases}
        \widehat{g}^{(\Omega)}_{c c'}\qquad a=b
        \\
        \, \\
        0\qquad a\neq b
    \end{cases}
\end{equation}
The order parameters \eqref{eq:gs_order_parameters} represent the statistics sufficient to characterize the spectral properties of the matrix $\widehat{\mathbf{U}}$: the $g^{(\Psi)}$ represents the resolvent function of $\widehat{\mathbf{U}}$ itself (see Eq.~\eqref{eq:meaning_of_g_Psi} later in the text), while the other order parameters capture the influence of different elements of the RF model on the spectrum of $\widehat{U}$. The $g_{cc'}^{(\Omega)}$ are resolvent functions of the projection of $\widehat{\mathbf{U}}$ in data space: one can verify that
\begin{equation}
    \label{eq:meaning_g_Omegas}
    g_{cc'}^{(\Omega)} = \frac{1}{N P}\operatorname{Tr}\left[\mathbf{W}^\intercal\operatorname{diag}(\bm{\gamma}_c)(\widehat{\mathbf{U}}-z)^{-1}\operatorname{diag}(\bm{\gamma}_{c'}) \mathbf{W}\right].
\end{equation}
They capture the correlations of the classes observed by the model, projected back into the data space. 
The resolvents $g_c^{(\Psi, h)}, g_c^{(\Psi, \varsigma)}$ respectively capture the effect of GEP noise variances ($\bm{h}_c^{\odot 2}$) and feature variances ($\bm{\varsigma}_c$): one can check that
\begin{equation}
    \label{eq:meaning_g_Psi_h}
    g_c^{(\Psi, h)}=\frac{1}{P}\operatorname{Tr}\left[\operatorname{diag}(\bm{h}_c)(\widehat{\mathbf{U}}-z)^{-1}\operatorname{diag}(\bm{h}_c)\right]
\end{equation}
\begin{equation}
    \label{eq:meaning_g_Psi_varsigma}
    g_c^{(\Psi, \varsigma)}=\frac{1}{P}\operatorname{Tr}\left[\operatorname{diag}(\bm{\varsigma}_c^{\odot 1/2})(\widehat{\mathbf{U}}-z)^{-1}\operatorname{diag}(\bm{\varsigma}_c^{\odot 1/2})\right].
\end{equation}
Finally, the $g_{c c'}^{(\Psi, \gamma)}$ express class correlations observed by the model in feature space:
\begin{equation}
    \label{eq:meaning_g_gammas}
    g_{cc'}^{(\Psi, \gamma)} = \frac{1}{P}\operatorname{Tr}\left[\operatorname{diag}(\bm{\gamma}_c)(\widehat{\mathbf{U}}-z)^{-1}\operatorname{diag}(\bm{\gamma}_{c'})\right].
\end{equation}
When centroids are extensive, these effects are heterogeneous in feature space: the model displays a different expressivity among the different hidden neurons. Essentially, centroids make neuron activations statistically correlated. In contrast, in the limit of subextensive centroids studied later in Appendix~\ref{sec:spectrum_subextensive_centroids}, the model acts homogeneously on the different hidden neurons: in this limit, one has $\bm{\gamma}_c\rightarrow \gamma_c \bm{1}_P$ and similarly for the other GEP vectors.

\paragraph{Applying the RS diagonal ansatz.} After clarifying the meaning of the order parameters, we proceed by simplifying all the terms in Eq.~\eqref{eq:pluppa}, taking into account the previous integrations over the $\mathbf{\Psi}, \mathbf{\Omega}$:
\begin{equation*}
\begin{gathered}
    \log\det \Biggl(
    i\widehat{\mathbf{G}}^{(\Psi)}\otimes\mathbf{I}_P
    +\sum_c i\widehat{\bm{G}}_{c}^{(\Psi, \varsigma)}\otimes\,\operatorname{diag}(\bm{\varsigma}_c) \\
    +\sum_{c c'}i\widehat{\bm{G}}_{c c'}^{(\Psi, \gamma)}\otimes\,\operatorname{diag}(\boldsymbol{\gamma}_c\odot\boldsymbol{\gamma}_{c'})
    +\sum_c i\widehat{\bm{G}}_{c}^{(\Psi, h)}\otimes\,\operatorname{diag}(\boldsymbol{h}_{c}^{\odot 2})
    \Biggr) \simeq \\
    \, \\
    n\log\det\left(
    i \widehat{g}_{\Psi}\mathbf{I}_P
    + \sum_c i\widehat{g}_{c}^{(\Psi, \varsigma)}\operatorname{diag}(\bm{\varsigma}_c)
    +\sum_{c c'} i\widehat{g}_{c c'}^{(\Psi, \gamma)}\operatorname{diag}(\boldsymbol{\gamma}_c\odot\boldsymbol{\gamma}_{c'})
    +\sum_c i\widehat{g}_{c}^{(\Psi, h)}\operatorname{diag}(\boldsymbol{h}_{c}^{\odot 2})
    \right) = \\
    \, \\
    n \sum_{p=1}^P \log\left(
    i \widehat{g}_{\Psi}
    + \sum_c i\widehat{g}_{c}^{(\Psi, \varsigma)}\varsigma_c^{(p)}
    +\sum_{c c'} i\widehat{g}_{c c'}^{(\Psi, \gamma)}\gamma_c^{(p)}\gamma_{c'}^{(p)}
    +\sum_c i\widehat{g}_{c}^{(\Psi, h)}(h_{c}^{(p)})^2
    \right)
\end{gathered}
\end{equation*}
\\
\begin{equation*}
    \log\det \left(
    i\widehat{\boldsymbol{G}}^{(\Omega)}
    \right)\simeq n \log \det i\{\widehat{g}_{c c'}^{(\Omega)}\}_{c c'}
\end{equation*}
\\
\begin{equation*}
    \log \det \left(\boldsymbol{I}_N+\frac{\chi_p}{\chi_m}\boldsymbol{J}_c\right) \simeq
    n\log\left(1+\frac{\chi_p}{\chi_m}g_{cc}^{(\Omega)}\right)
\end{equation*}
\; \\
\begin{equation*}
    \log\det\left(\mathbf{I}_P+\frac{1}{\chi_m}\operatorname{diag}(\boldsymbol{h}_c^{\odot 2})\boldsymbol{\mathcal{J}}_c\right)\simeq
    n\log\left[1+\frac{\chi_p}{\chi_m}\left(1-\frac{\chi_p}{\chi_m}\frac{g_{cc}^{(\Omega)}}{1+\frac{\chi_p}{\chi_m}g_{cc}^{(\Omega)}}\right)g_{a; c}^{(\Psi, h)}\right]
\end{equation*}
\;\\
\begin{equation*}
    \mathrm{K}_{ab}^{c c'}=\delta_{a b}g_{c c'}^{(\Psi, \gamma)}\Longrightarrow (\boldsymbol{\mathrm{K}}^{-1})_{a b}^{c c'}=\delta_{a b}\,(\boldsymbol{g}_{(\Psi, \gamma)}^{-1})_{c c'}
\end{equation*}
\, \\
\begin{equation*}
    \frac{1}{2\chi_p}\sum_{a b}\sum_{c c'}(\mathbf{K}(\bm{g}^{(\Psi, \gamma)})^{-1})_{a b}^{c c'}g_{a b; c c'}^{(\Omega)}\simeq \frac{n}{2\chi_p}\sum_{c c'}(\boldsymbol{g}_{(\Psi, \gamma)}^{-1})_{c c'}g_{c c'}^{(\Omega)}
\end{equation*}
\;\\
\begin{equation*}
    \frac{1}{2\chi_p}\log\det \boldsymbol{\mathrm{K}} \simeq \frac{n}{2 \chi_p}\log\det \boldsymbol{g}^{(\Psi, \gamma)}
\end{equation*}
\;\\
Here and in the following lines, the symbols $\boldsymbol{g}^{(\Psi, \gamma)}$ and $\boldsymbol{g}^{(\Omega)}$ indicate the matrices $g_{c c'}^{(\Psi, \gamma)}$, $g_{c c'}^{(\Omega)}$ for $c,c'=1,\dots, C$.
Collecting everything, we can finally write
\begin{equation}
    \label{eq:saddle_point_integral}
    \overline{\mathcal{Z}^n} \propto
    \Expected{
    \int d\{\bm{g}^{(\mathrm{all})}\}d\{\widehat{\bm{g}}^{(\mathrm{all})}\}e^{n P \mathcal{F}\left(
    \{\bm{g}^{(\mathrm{all}))},\widehat{\bm{g}}^{(\mathrm{all})}\}
    \right)}
    }{\{\mu_c\}}\propto \Expected{
    e^{n P \mathcal{F}\left(
    \{\bm{g}_*^{(\mathrm{all}))}\text{,}\,\widehat{\bm{g}}_*^{(\mathrm{all})}\}
    \right)}
    }{\{\mu_c\}}
\end{equation}
The compact notation $\{\bm{g}^{(\mathrm{all})}\}, \{\widehat{\bm{g}}^{(\mathrm{all})}\}$ identifies all the order parameters and their conjugate parameters. The action $\mathcal{F}$, in its final form, reads
\begin{equation}
\label{eq:action}
\begin{gathered}
    \mathcal{F}
    \left(\{\bm{g}^{(\mathrm{all})}\text{,}\,\widehat{\bm{g}}^{(\mathrm{all})}\}
    \right) = 
    i \widehat{g}_{\Psi} g_{\Psi}
    +\sum_c i\widehat{g}_{c}^{(\Psi, \varsigma)}g_{c}^{(\Psi, \varsigma)} 
    +\sum_{c c'}i\widehat{g}_{c c'}^{(\Psi, \gamma)} g_{c c'}^{(\Psi, \gamma)} \\
    \, \\
    +\sum_c i\widehat{g}_{c}^{(\Psi, h)}g_{c}^{(\Psi, h)}
    +\frac{1}{\chi_p}\sum_{c c'}i\widehat{g}_{c c'}^{(\Omega)} g_{c c'}^{(\Omega)}
    -\frac{1}{2\chi_p}\log\det i\widehat{\bm{g}}_{\Omega}
    \, \\
    \\
    -\frac{1}{2 P}\sum_{p=1}^P \log\left(
    i \widehat{g}_{\Psi}
    + \sum_c i\widehat{g}_{c}^{(\Psi, \varsigma)}\varsigma_c^{(p)}
    +\sum_{c c'} i\widehat{g}_{c c'}^{(\Psi, \gamma)}\gamma_c^{(p)}\gamma_{c'}^{(p)}
    +\sum_c i\widehat{g}_{c}^{(\Psi, h)}(h_{c}^{(p)})^2
    \right)
     \\
    \; \\
    -\frac{\Delta}{2 e^{-2t}}\sum_c \frac{b_c}{\sigma_c^2}g_{cc}^{(\Omega)}
    -\frac{1}{2\chi_p}\sum_{c c'}(\boldsymbol{g}_{(\Psi, \gamma)}^{-1})_{c c'}g_{c c'}^{(\Omega)} \\
    \, \\
    -\frac{1}{2\chi_p}\log\det \boldsymbol{g}^{(\Psi, \gamma)}
    -\frac{1}{2}\sum_c b_c\left(g_{c}^{(\Psi, \varsigma)}-z\;g^{(\Psi)}\right) \\
    \, \\
    -\frac{\chi_m}{2\chi_p}\sum_c b_c\log\left(1+\frac{\chi_p}{\chi_m}g_{cc}^{(\Omega)}+\frac{\chi_p}{\chi_m}g_{c}^{(\Psi,h)}\right)
\end{gathered}
\end{equation}
We finally use \eqref{eq:Edwards_Jones} to get
\begin{equation}
\label{eq:meaning_of_g_Psi}
\begin{gathered}
    g(z)=
    2\frac{\partial}{\partial z}\lim_{P\rightarrow\infty}\frac{1}{P}\Expected{\log \mathcal{Z}_U}{\mathbf{X}, \mathbf{W}}
    = \\
    2\frac{\partial}{\partial z}\lim_{n\rightarrow 0}\lim_{P\rightarrow\infty}\frac{1}{P}\frac{\Expected{e^{n P \mathcal{F}}}{\{\bm{\mu_c}\}}-1}{n}
    =2\frac{\partial}{\partial z}\Expected{\mathcal{F}}{\{\mu_c\}}= g_{\Psi}^*(z).
\end{gathered}
\end{equation}
\paragraph{Derivation of the spectral equations.} The order parameter $g_{\Psi}^*$ maximizing the action \eqref{eq:action} is the resolvent function that we wish to determine. This quantity depends on all the other resolvents appearing in \eqref{eq:action}: to understand how, we need to write the saddle point equations for \eqref{eq:action}, whose solutions are the resolvents yielding the maximum of \eqref{eq:action}.
For brevity of notation, we omit the symbol $*$ on the resolvents, previously used to indicate solutions of saddle point equations.
We start with those for the conjugate order parameters:
\\
\begin{equation}
\label{eq:sp_g_psi_hat}
    \frac{\partial \mathcal{F}}{\partial g_{\Psi}}=0\Longrightarrow i\widehat{g}_{\Psi}=-\frac{z}{2}
\end{equation}
\\
\begin{equation}
\label{eq:sp_g_psi_varsigma_hat}
    \frac{\partial \mathcal{F}}{\partial g_{c}^{(\Psi, \varsigma)}}=0\Longrightarrow i\widehat{g}_{c}^{(\Psi, \varsigma)}=\frac{b_c}{2}
\end{equation}
\\
\begin{equation}
\label{eq:sp_g_psi_h_hat}
    \frac{\partial \mathcal{F}}{\partial g_{c}^{(\Psi, h)}}=0\Longrightarrow i\widehat{g}_{c}^{(\Psi, h)}=\frac{b_c}{2}\frac{1}{1+\frac{\chi_p}{\chi_m}g_{cc}^{(\Omega)}+\frac{\chi_p}{\chi_m}g_{c}^{(\Psi,h)}}
\end{equation}
\\
\begin{equation}
\label{eq:sp_g_psi_gamma_hat}
\begin{gathered}
    \frac{\partial \mathcal{F}}{\partial \bm{g}_{(\Psi, \gamma)}}=0\Longrightarrow i\widehat{\bm{g}}_{(\Psi, \gamma)}=
    \frac{1}{2\chi_p}\frac{\partial}{\partial \bm{g}_{(\Psi, \gamma)}}\operatorname{Tr}\left(\bm{g}_{(\Psi, \gamma)}^{-1}\bm{g}_{(\Omega)}^\intercal\right)
    +\frac{1}{2\chi_p}\frac{1}{\det \boldsymbol{g}^{(\Psi, \gamma)}}\frac{\partial \det \boldsymbol{g}^{(\Psi, \gamma)}}{\partial \bm{g}_{(\Psi, \gamma)}} \\
    \, \\
    \qquad\qquad\qquad=
    -\frac{1}{2\chi_p}\bm{g}_{(\Psi, \gamma)}^{-1}\bm{g}_\Omega\bm{g}_{(\Psi,\gamma)}^{-1}
    +\frac{1}{2\chi_p}\bm{g}_{(\Psi, \gamma)}^{-1}
\end{gathered}
\end{equation}
\\
\begin{equation}
\label{eq:sp_g_omega_diag_hat}
   \frac{\partial \mathcal{F}}{\partial g_{c c}^{(\Omega)}}=0\Longrightarrow i\widehat{g}_{cc}^{(\Omega)}=\frac{1}{2}(\boldsymbol{g}_{(\Psi, \gamma)}^{-1})_{cc}+\frac{b_c}{2}\frac{\chi_p}{1+\frac{\chi_p}{\chi_m}g_{cc}^{(\Omega)}+\frac{\chi_p}{\chi_m}g_{c}^{(\Psi,h)}}+\frac{\Delta \chi_p}{2 e^{-2t}}\frac{b_c}{\sigma_c^2}
\end{equation}
\\
\begin{equation}
\label{eq:sp_g_omega_off_hat}
   \frac{\partial \mathcal{F}}{\partial g_{c c'}^{(\Omega)}}=0\Longrightarrow i\widehat{g}_{c c'}^{(\Omega)}=\frac{1}{2}(\boldsymbol{g}_{(\Psi, \gamma)}^{-1})_{c c'}.
\end{equation}
Second, we write the saddle point equations for the order parameters:
\begin{equation}
\label{eq:sp_g_Psi}
\begin{gathered}
    \frac{\partial \mathcal{F}}{\partial (i \widehat{g}_{\Psi})}=0 \\
    \, \\
    \Longrightarrow 
    g_{\Psi}=\frac{1}{2 P}\sum_{p=1}^P 
    \frac{1}{
    i \widehat{g}_{\Psi}
    + \sum_c i\widehat{g}_{c}^{(\Psi, \varsigma)}\varsigma_c^{(p)}
    +\sum_{c c'} i\widehat{g}_{c c'}^{(\Psi, \gamma)}\gamma_c^{(p)}\gamma_{c'}^{(p)}
    +\sum_c i\widehat{g}_{c}^{(\Psi, h)}(h_{c}^{(p)})^2
    }
\end{gathered}
\end{equation}
\\
\begin{equation}
\label{eq:sp_g_Psi_varsigma}
\begin{gathered}
    \frac{\partial \mathcal{F}}{\partial (i \widehat{g}_{c}^{(\Psi,\varsigma)})}=0 \\
    \, \\
    \Longrightarrow 
    g_c^{(\Psi, \varsigma)}=\frac{1}{2 P}\sum_{p=1}^P 
    \frac{\varsigma_c^{(p)}}{
    i \widehat{g}_{\Psi}
    + \sum_c i\widehat{g}_{c}^{(\Psi, \varsigma)}\varsigma_c^{(p)}
    +\sum_{c c'} i\widehat{g}_{c c'}^{(\Psi, \gamma)}\gamma_c^{(p)}\gamma_{c'}^{(p)}
    +\sum_c i\widehat{g}_{c}^{(\Psi, h)}(h_{c}^{(p)})^2
    }
\end{gathered}
\end{equation}
\\
\begin{equation}
\label{eq:sp_g_Psi_h}
\begin{gathered}
    \frac{\partial \mathcal{F}}{\partial (i \widehat{g}_{c}^{(\Psi,h)})}=0 \\
    \, \\
    \Longrightarrow 
    g_c^{(\Psi, h)}=\frac{1}{2 P}\sum_{p=1}^P 
    \frac{(h_c^{(p)})^2}{
    i \widehat{g}_{\Psi}
    + \sum_c i\widehat{g}_{c}^{(\Psi, \varsigma)}\varsigma_c^{(p)}
    +\sum_{c c'} i\widehat{g}_{c c'}^{(\Psi, \gamma)}\gamma_c^{(p)}\gamma_{c'}^{(p)}
    +\sum_c i\widehat{g}_{c}^{(\Psi, h)}(h_{c}^{(p)})^2
    }
\end{gathered}
\end{equation}
\\
\begin{equation}
\label{eq:sp_g_Psi_gamma}
\begin{gathered}
    \frac{\partial \mathcal{F}}{\partial (i \widehat{g}_{c c'}^{(\Psi,\gamma)})}=0 \\
    \, \\
    \Longrightarrow 
    g_{c c'}^{(\Psi, \gamma)}=\frac{1}{2 P}\sum_{p=1}^P 
    \frac{\gamma_c^{(p)} \gamma_{c'}^{(p)}}{
    i \widehat{g}_{\Psi}
    + \sum_c i\widehat{g}_{c}^{(\Psi, \varsigma)}\varsigma_c^{(p)}
    +\sum_{c c'} i\widehat{g}_{c c'}^{(\Psi, \gamma)}\gamma_c^{(p)}\gamma_{c'}^{(p)}
    +\sum_c i\widehat{g}_{c}^{(\Psi, h)}(h_{c}^{(p)})^2
    }
\end{gathered}
\end{equation}
\\
\begin{equation}
\label{eq:sp_g_Omega}
    \frac{\partial \mathcal{F}}{\partial (i \widehat{g}_{c c'}^{(\Omega)})}=0 \\
    \, \\
    \Longrightarrow 
    g_{c c'}^{(\Omega)}=\frac{1}{2}((i \widehat{\bm{g}}_{\Omega})^{-1})_{c c'}.
\end{equation}

\subsection{Spectral equations for extensive centroids}

We can finally write the equations necessary to solve for the spectrum of $\widehat{\mathbf{U}}$ by combining Eqs.~\eqref{eq:sp_g_psi_hat}, \eqref{eq:sp_g_psi_h_hat}, \eqref{eq:sp_g_psi_gamma_hat}, \eqref{eq:sp_g_omega_diag_hat}, \eqref{eq:sp_g_omega_off_hat} with Eqs.~\eqref{eq:sp_g_Psi}, \eqref{eq:sp_g_Psi_h}, \eqref{eq:sp_g_Psi_gamma}, \eqref{eq:sp_g_Omega}. Note that Eqs.~\eqref{eq:sp_g_psi_varsigma_hat} and \eqref{eq:sp_g_Psi_varsigma} are not necessary to determine the resolvent function $g(z)\equiv g_{\Psi}(z)$, since $i\widehat{g}_{c}^{(\Psi, \varsigma)}$ is a constant (Eq.~\eqref{eq:sp_g_psi_varsigma_hat}), independent of all other resolvents.
For the sake of compactness, we simplify the notation of the auxiliary resolvents and define the functions 
\begin{equation}
    \label{eq:D_call}
    \mathcal{D}(\{\mu_c\})=\sum_c b_c \varsigma_c(\mu_c)
\end{equation}
\begin{equation}
    \label{eq:R_call}
\begin{split}
    \mathcal{R}_z(\{g_{c}^{(h)}(z), g_{c c'}^{(\gamma)}(z), g_{c c'}^{(\Omega)}(z)\},\{\mu_c\}) &=
    -\frac{2}{\chi_p}\sum_{cc'}\mathrm{y}_{c c'}(\boldsymbol{g}^{(\Psi, \gamma)}, \boldsymbol{g}^{(\Omega)})\;\gamma_c(\mu_c)\gamma_{c'}(\mu_{c'}) \\
    &-\sum_c \frac{b_c h_c^{2}(\mu_c)}{1+\frac{\chi_p}{\chi_m}g_{cc}^{(\Omega)}+\frac{\chi_p}{\chi_m}g_{c}^{(\Psi, h)}}
\end{split}
\end{equation}
for $\mu_c\in \Real,c=1,\dots, C$. In Eq.~\eqref{eq:R_call}, we renamed $\mathrm{y}_{c c'} = i\chi_p\widehat{g}_{c c'}^{(\Psi, \gamma)},\forall c, c'=1,\dots, C$. Moreover, in \eqref{eq:D_call} and \eqref{eq:R_call} we introduced the GEP functions
\begin{equation}
    \label{eq:GEP_functions}
    \begin{gathered}
    \varsigma_c(\mu)=\widetilde{\beta}_{c}(\mu)-\beta_{c}(\mu)-\frac{\Delta}{\sigma_c^2 e^{-2 t}}\gamma_{c}^{2}(\mu) \\
    \,\\
    h_c^2(\mu) = \beta_c(\mu)-\alpha_c^2(\mu)-\gamma_c^2(\mu)
    \\
    \\
    \text{with}
    \\
    \, \\
    \alpha_c(\mu) = \Expected{\phi\left(\Gamma_c(t) x+e^{-t}\mu\right)}{x\sim \mathcal{N}(0, 1)} \\
    \\
    \widetilde{\beta}_c(\mu) = \Expected{\phi^2\left(\Gamma_c(t) x+e^{-t}\mu\right)}{x\sim \mathcal{N}(0, 1)} \\
    \\
    \beta_c(\mu) = \Expected{\phi\left(\Gamma_c(t) x_1+e^{-t}\mu\right)\phi\left(\Gamma_c(t) x_2+e^{-t}\mu\right)}{x_1, x_2\sim \mathcal{N}(0, \boldsymbol{\Sigma}_c(t))} \\
    \\
    \gamma_c(\mu) = \frac{\sigma_c e^{-t}}{\Gamma_c(t)}\Expected{x\;\phi(\Gamma_c(t)x+e^{-t}\mu)}{x\sim \mathcal{N}(0, 1)}.
    \end{gathered}
\end{equation}
Finally, let us use the notation
\begin{equation}
    \Expected{\cdot}{\{\mu_c\}}\equiv \int \frac{d\mu_1\dots d\mu_C}{(2\pi)^{C/2}\sqrt{\det \bm{r}}}e^{-\frac{1}{2}\sum_{c c'}\mu_c (\bm{r}^{-1})_{c c'}\mu_{c'}}(\cdot),
\end{equation}
with $r_{c c'}=\frac{1}{N}\bm{m}_c\cdot \bm{m}_{c'}$.
Then, in the limit $P\rightarrow\infty$ the resolvent $g(z)$ can be determined by solving the following system of equations for $k, k'=1,\dots, C$:
\begin{equation}
\label{eq:saddle_point_extensive_centroid}
\boxed{
\begin{gathered}
    \\
    \qquad g_k^{(h)}(z) = \Expected{\frac{h_k^2(\mu_k)}{\mathcal{D}(\{\mu_{c}\})-z-\mathcal{R}(\{g_{c}^{(h)}(z), g_{c c'}^{(\gamma)}(z), g_{c c'}^{(\Omega)}(z)\},\{\mu_{c}\})}}{\{\mu_{c}\}}\qquad \\
    \, \\
    \,g_{k k'}^{(\gamma)}(z) = \Expected{\frac{\gamma_k(\mu_k)\gamma_{k'}(\mu_{k'})}{\mathcal{D}(\{\mu_{c}\})-z-\mathcal{R}(\{g_c^{(h)}(z), g_{c c'}^{(\gamma)}(z), g_{c c'}^{(\Omega)}(z)\},\{\mu_c\})}}{\{\mu_c\}}\, \\
    \, \\
    i\widehat{g}_{k k}^{(\Omega)}=\frac{1}{2}(\boldsymbol{g}_{(\gamma)}^{-1})_{kk}+\frac{b_k}{2}\frac{\chi_p}{1+\frac{\chi_p}{\chi_m}g_{kk}^{(\Omega)}+\frac{\chi_p}{\chi_m}g_{k}^{(h)}}+\frac{\Delta \chi_p}{2 e^{-2t}}\frac{b_k}{\sigma_k^2} \\
    \, \\
    i\widehat{g}_{k k'}^{(\Omega)}=\frac{1}{2}(\boldsymbol{g}_{(\gamma)}^{-1})_{k k'}\qquad k\neq k' \\
    \, \\
    \bm{g}_{(\Omega)}=\frac{1}{2}(i \widehat{\bm{g}}_{\Omega})^{-1}.
    \\ \\
\end{gathered}
}
\end{equation}
The resolvent $g(z)$ can be computed from the solution of \eqref{eq:saddle_point_extensive_centroid} through the formula
\begin{equation}
    \label{eq:g_star}
    \boxed{
\begin{gathered}
    \\
    \qquad g(z) = \Expected{\frac{1}{\mathcal{D}(\{\mu_c\})-z-\mathcal{R}(\{g_c^{(h)}(z), g_{c c'}^{(\gamma)}(z), g_{c c'}^{(\Omega)}(z)\},\{\mu_c\})}}{\{\mu_c\}}.\qquad \\
    \, \\
\end{gathered}
}
\end{equation}
Equations like \eqref{eq:saddle_point_extensive_centroid}, usually in a much simpler form, appear whenever a random matrix ensemble is non-rotationally invariant: by inspection of \eqref{eq:sp_g_Psi} and \eqref{eq:g_star}, \eqref{eq:R_call}, \eqref{eq:D_call}, one can deduce that the diagonal entries of the resolvent matrix $\mathbf{G}(z)$ yielded by the replica computation in the limit $P\rightarrow\infty$ are
\begin{equation}
    G_{pp}(z) = \frac{1}{
    \mathcal{D}(\mu_1^{(p)},\dots,\mu_C^{(p)})-z-\mathcal{R}(\{g_{c}^{(h)}(z), g_{c c'}^{(\gamma)}(z), g_{c c'}^{(\Omega)}(z)\},\mu_1^{(p)},\dots,\mu_C^{(p)}),
    }
\end{equation}
where $p=1,\dots, P$.
The diagonal entries of the resolvent are fluctuating quantities also in the high-dimensional limit, correlated to the statistics of the projected class centroids. Conversely, in rotationally invariant ensembles, like the Wishart ensemble, the diagonal entries of the resolvent concentrate around their mean in the high-dimensional limit.

\eqref{eq:saddle_point_extensive_centroid} is a system of $C(C+2)$ complex equations\footnote{The total number of equations is actually $2C(C+2)$, since the system is solved for the real and imaginary parts of each of the resolvents in \eqref{eq:saddle_point_extensive_centroid}.}, for given $z\in \Complex$ \footnote{In the high-dimensional limit, in regions of the spectrum with absolute continuous measure, the resolvent develops a branch cut singularity for $\operatorname{Im}z=0$, rather than having first order poles on the eigenvalues, as happens for discrete spectra. In this situation, the equations that define the spectrum are well defined in the whole complex plane.}.
In the multiclass case $C>2$, it is a rather difficult system to solve. In the special binary case, $C=2$, it is possible to simplify some terms in \eqref{eq:saddle_point_extensive_centroid}: first,  the explicit expression of the matrix inverse
\begin{equation}
    \boldsymbol{g}_{(\Psi, \gamma)}^{-1}=
    \frac{1}{\det \boldsymbol{g}_{(\Psi, \gamma)}}
    \begin{pmatrix}
        g_{22}^{(\Psi, \gamma)}\qquad - g_{12}^{(\Psi, \gamma)}\\
        - g_{12}^{(\Psi, \gamma)} \qquad g_{11}^{(\Psi, \gamma)}
    \end{pmatrix}
    \qquad 
    \det \boldsymbol{g}_{(\Psi, \gamma)}=
    g_{11}^{(\Psi, \gamma)}g_{22}^{(\Psi, \gamma)}
    -(g_{12}^{(\Psi, \gamma)})^2
\end{equation}
we can write for $\mathbf{y}$: 
\begin{equation}
\begin{gathered}
    \mathbf{y} = 
    \frac{1}{2\chi_p}\frac{1}{g_{11}^{(\Psi,\gamma)}g_{22}^{(\Psi,\gamma)}-(g_{12}^{(\Psi,\gamma)})^2}
    \begin{pmatrix}
        g_{22}^{(\Psi,\gamma)}\qquad -g_{12}^{(\Psi,\gamma)} \\
        -g_{12}^{(\Psi,\gamma)}\qquad g_{11}^{(\Psi,\gamma)}
    \end{pmatrix} \\
    \, \\
    \times \left[\mathbf{I}_2-\frac{1}{g_{11}^{(\Psi,\gamma)}g_{22}^{(\Psi,\gamma)}-(g_{12}^{(\Psi,\gamma)})^2}
    \begin{pmatrix}
        g_{22}^{(\Psi,\gamma)}\qquad -g_{12}^{(\Psi,\gamma)} \\
        -g_{12}^{(\Psi,\gamma)}\qquad g_{11}^{(\Psi,\gamma)}
    \end{pmatrix}
    \begin{pmatrix}
        g_{11}^{(\Omega)}\qquad g_{12}^{(\Omega)} \\
        g_{12}^{(\Omega)}\qquad g_{22}^{(\Omega)}
    \end{pmatrix}
    \right]
\end{gathered}
\end{equation}
and unveil the equations for the entries of $\mathbf{g}_{\Omega}$, by explicitly writing the matrix inverse in \eqref{eq:sp_g_Omega} and using \eqref{eq:sp_g_omega_diag_hat}, \eqref{eq:sp_g_omega_off_hat}
\begin{equation}
\label{eq:sp_eqs_gomega}
    \begin{gathered}
        g_{11}^{(\Omega)}=\frac{l_{22}}{l_{11}l_{22}-[(\boldsymbol{g}_{(\Psi, \gamma)}^{-1})_{12}]^2} \\
        \, \\
        g_{22}^{(\Omega)}=\frac{l_{11}}{l_{11}l_{22}-[(\boldsymbol{g}_{(\Psi, \gamma)}^{-1})_{12}]^2} \\
        \, \\
        g_{12}^{(\Omega)}=-\frac{(\boldsymbol{g}_{(\Psi, \gamma)}^{-1})_{12}}{l_{11}l_{22}-[(\boldsymbol{g}_{(\Psi, \gamma)}^{-1})_{12}]^2} \\
        \, \\
        \text{with} \\
        \, \\
        l_{cc}=\chi_p \left[\frac{\Delta b_c}{ e^{-2t}\sigma_c^2}+\frac{1}{\chi_p}(\boldsymbol{g}_{(\Psi, \gamma)}^{-1})_{cc}
    +\frac{b_c}{1+\frac{\chi_p}{\chi_m}g_{cc}^{(\Omega)}+\frac{\chi_p}{\chi_m}g_{c}^{(\Psi, h)}}\right].
    \end{gathered}
\end{equation}
In the binary case, the system \eqref{eq:saddle_point_extensive_centroid} counts $16$ self-consistent real equations. We reserve the study of Eqs.~\eqref{eq:saddle_point_extensive_centroid} for future work.

\subsection{Spectral equations for subextensive centroids}
\label{sec:spectrum_subextensive_centroids}

We consider the case of subextensive centroids, i.e. for any class $c$ one has $\lVert\bm{\mu}_c\lVert=o_P(P)$: this setting represents datasets with centered classes.
In this case, $P(\mu_1,\dots,\mu_C)\simeq \delta(\mu_1)\cdots\delta(\mu_C)$ and the problem can be considered  relatively straightforward: defining for $c=1,\dots,C$ the scalars
\begin{equation}
\label{eq:scalar_Hermite_coefficients}
\begin{gathered}
    \widetilde{\beta}_c=\Expected{\phi^2(\Gamma_c z)}{z}\qquad \qquad \gamma_c=\Expected{z\,\phi(\Gamma_c z)}{z} \\
    \, \\
    \beta_c=\Expected{\phi(\Gamma_c z_1)\phi(\Gamma_c z_2)}{(z_1, z_2)\sim \mathcal{N}(0,\bm{\Sigma}_c(t))} \\
    \, \\
    \varsigma_c = \widetilde{\beta}_c-\beta_c-\frac{\Delta \gamma_c^2}{\sigma_c^2 e^{-2t}}\qquad h_c^2 = \beta_c-\gamma_c^2
\end{gathered}
\end{equation}
the resolvent function $g(z)\equiv g_{\Psi}(z)$ is determined from the solution of the following system of saddle point equations:
\begin{equation}
\label{eq:saddle_point_equations_finite_centroids}
\boxed{
\begin{gathered}
    \\
    \,\left(1-\frac{1}{\chi_p}\right)\frac{1}{g_{\Psi}}=-\frac{1}{\chi_p}\frac{g_\Omega}{g_{\Psi}^2}+\sum_c b_c\,(\varsigma_c-z)+\sum_c \frac{b_c h_c^2}{1+\frac{\chi_p}{\chi_m}[\gamma_c^2 g_{\Omega}+h_c^2 g_{\Psi}]}\, \\
    \, \\
    \frac{1}{\chi_p}\frac{1}{g_{\Omega}}=\frac{\Delta}{e^{-2t}}\sum_c \frac{b_c}{\sigma_c^2}\gamma_c^2+\frac{1}{\chi_p}\frac{1}{g_{\Psi}}+\sum_c \frac{b_c \gamma_c^2}{1+\frac{\chi_p}{\chi_m}[\gamma_c^2 g_{\Omega}+h_c^2g_{\Psi}]} \\\\
\end{gathered}
}
\end{equation}
Eqs.~\eqref{eq:saddle_point_equations_finite_centroids} can be obtained as follows: first, write the action \eqref{eq:action} in the limit of subextensive centroids, thus replacing the vector Hermite coefficients with their homogeneous counterparts in Eqs.~\eqref{eq:scalar_Hermite_coefficients}, $\widetilde{\bm{\beta}}_c\rightarrow \widetilde{\beta}_c\bm{1}_P,\,\bm{\beta}_c\rightarrow \beta_c\bm{1}_P,\, \bm{\gamma}_c\rightarrow \gamma_c \bm{1}_P$ (note that, for an odd activation, the coefficient $\bm{\alpha}_c$ vanishes in this limit). Second, note that $g_{c}^{(\Psi, h)}\rightarrow h_c^2\,g_{\Psi}$, $g_{c}^{(\Psi, \varsigma)}\rightarrow \varsigma_c\,g_{\Psi}$, $g_{c c'}^{(\Psi, \gamma)}\rightarrow \gamma_c\gamma_{c'}\,g_{\Psi}$, $g_{c c'}^{(\Omega)}\rightarrow \gamma_c \gamma_{c'}g_{\Omega}$; to handle singular terms such as determinants and inverses of these last two resolvents appearing in \eqref{eq:action}, set $g_{c c}^{(\Psi, \gamma)}=\gamma_c^2\,g_{\Psi}+\epsilon$, $g_{c c}^{(\Omega)}=\gamma_c^2\,g_{\Omega}+\epsilon$, expand and simplify. Then, derive the saddle point equations of the resulting action and obtain \eqref{eq:saddle_point_equations_finite_centroids}.

We remark that, while for extensive centroids $\widehat{\mathbf{U}}_{gep}$ in \eqref{eq:U_GEP} is non-rotationally invariant, in the limit of subextensive centroids $\widehat{\mathbf{U}}_{gep}$ consists of the sum of different Wishart matrices and therefore, the resulting matrix ensemble is rotationally invariant: usually, in this case the spectral equations are algebraic, and this is also the case for  \eqref{eq:saddle_point_equations_finite_centroids}.
Finally, note that, in contrast with equations \eqref{eq:saddle_point_extensive_centroid}, in the case of subextensive centroids the total number of real equations to solve is always $4$, for any number of classes $C$.

For $\sigma_1=\dots=\sigma_C=\sigma$, Eqs.~\eqref{eq:saddle_point_equations_finite_centroids} reduce to the result obtained in \cite{bonnaire2025diffusion} for data sampled from a single gaussian.

\begin{figure}
    \centering
    \includegraphics[width=0.475\linewidth]{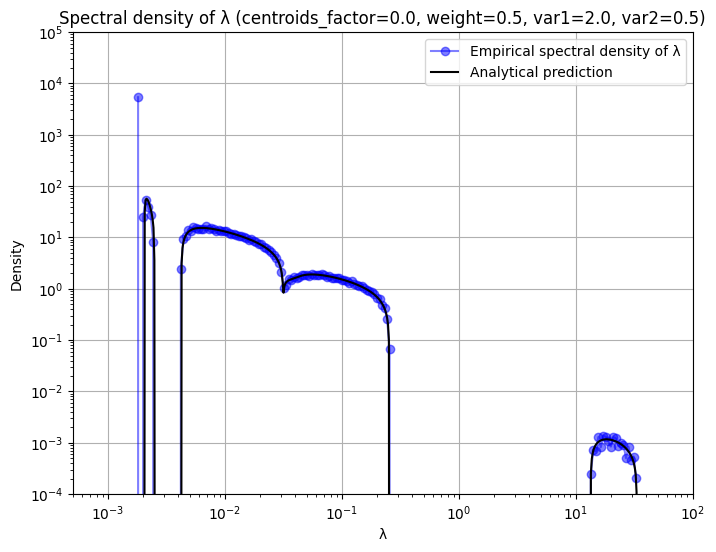}
    \includegraphics[width=0.475\linewidth]{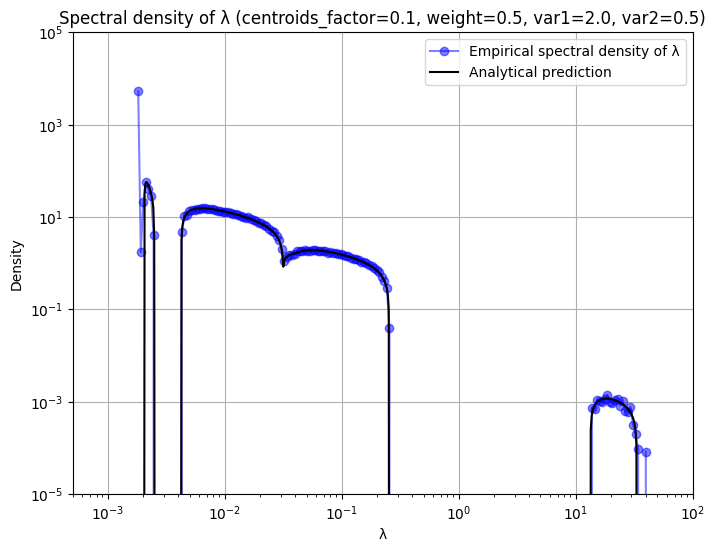}
    \caption{Test of Eqs.~\eqref{eq:saddle_point_equations_finite_centroids} for $t=0.01$.}
    \label{fig:empirical_VS_theory_rho_finite_centroids}
\end{figure}

In figure \ref{fig:empirical_VS_theory_rho_finite_centroids} we compare the solution of \eqref{eq:saddle_point_equations_finite_centroids} in the $C=2$ case with empirical histograms of the eigenvalues of $\widehat{\mathbf{U}}$, testing the case of zero centroid norm (left) and $\lVert\bm{m}_1\lVert=\lVert \bm{m}_2\lVert=1$ (right): we find a perfect agreement between theory and simulations. We compute the theoretical spectral density from the solution of \eqref{eq:saddle_point_equations_finite_centroids} using \eqref{eq:pdf_eigenvalues_theor}.

\subsubsection{Equations for the edges}
\label{sec:equations_for_the_edges}

The equations for the spectral edges $\lambda_*$ can be obtained from the spectral equations \eqref{eq:saddle_point_equations_finite_centroids} by performing the following joint limits
\begin{equation}
\label{eq:condition_for_the_edges} \lim_{\lambda\rightarrow\lambda_*}\operatorname{Im}g_{\Psi}(\lambda)=\lim_{\lambda\rightarrow\lambda_*}\operatorname{Im}g_{\Omega}(\lambda)=0,\qquad \lim_{\lambda\rightarrow\lambda_*}\frac{\operatorname{Im}g_{\Omega}(\lambda)}{\operatorname{Im}g_{\Psi}(\lambda)}\equiv \omega_*\in \Real/\{0\}.
\end{equation}
For any bulk of the spectrum, the lower and upper edges are defined, respectively, as the smallest and the largest eigenvalues for which the spectral equations admit a real solution:
conditions \eqref{eq:condition_for_the_edges} ensure that we find such points since they guaranty that we are approaching the edges from within the spectrum of $\widehat{\mathbf{U}}$ ---otherwise the quantity $\omega_*$ would be undefined. By applying them to the real and imaginary parts of Eqs.~\eqref{eq:saddle_point_equations_finite_centroids}, and setting $\operatorname{Re}g_{\Psi}(\lambda_*)=g_\Psi^*$ and $\operatorname{Re}g_{\Omega}(\lambda_*)=g_\Omega^*$, we obtain the following system of three algebraic equations

\begin{equation}
\label{eq:eqs_for_edges_finite_centroids}
\begin{gathered}
    -\left(1-\frac{1}{\chi_p}\right)\frac{1}{(g_{\Psi}^*)^2}=
    -\frac{\omega_*}{\chi_p (g_{\Psi}^*)^2}
    +\frac{2 g_{\Omega}^*}{\chi_p (g_{\Psi}^*)^3}
    -\frac{\chi_p}{\chi_m}\sum_c \frac{b_c h_c^2(h_c^2+\omega_*\gamma_c^2)}{\{1+\frac{\chi_p}{\chi_m}[\gamma_c^2 g_\Omega^*+h_c^2g_{\Psi}^*]\}^2} \\
    \, \\
    \frac{\omega_*}{\chi_p (g_{\Omega}^*)^2}=
    \frac{1}{\chi_p}\frac{1}{(g_{\Psi}^*)^2}
    +\frac{\chi_p}{\chi_m}\sum_c \frac{b_c \gamma_c^2(h_c^2+\omega_*\gamma_c^2)}{\{1+\frac{\chi_p}{\chi_m}[\gamma_c^2 g_\Omega^*+h_c^2g_{\Psi}^*]\}^2} \\
    \, \\
    \frac{1}{\chi_p g_{\Omega}^*}
    =
    \frac{\Delta}{e^{-2t}}\sum_c \frac{b_c}{\sigma_c^2}\gamma_c^2
    +\frac{1}{\chi_p}\frac{1}{g_{\Psi}^*}
    +\sum_c \frac{b_c \gamma_c^2}{1+\frac{\chi_p}{\chi_m}[\gamma_c^2 g_\Omega^*+h_c^2 g_{\Psi}^*]}. \\
\end{gathered}
\end{equation}
The spectral edges are computed from the solutions of \eqref{eq:eqs_for_edges_finite_centroids}:
\begin{equation}
\label{eq:spectral_edges_finite_centroids}
\boxed{
    \lambda_* = \sum_c b_c \varsigma_c
    +\sum_c \frac{b_c h_c^2}{1+\frac{\chi_p}{\chi_m}[\gamma_c^2 g_\Omega^*+h_c^2g_{\Psi}^*]}
    -\left(1-\frac{1}{\chi_p}\right)\frac{1}{g_{\Psi}^*}
    -\frac{1}{\chi_p}\frac{g_\Omega^*}{(g_\Psi^*)^2}.
}
\end{equation}
Eqs.~\eqref{eq:eqs_for_edges_finite_centroids} and \eqref{eq:spectral_edges_finite_centroids} are used to compute the curves in fig. \ref{fig:timescale_second} in the limit $t\rightarrow 0$, for the case of $C=2$ classes.

One can show that Eqs.~\eqref{eq:eqs_for_edges_finite_centroids} admit at most $2C+4$ solutions. In the following, we connect the solutions of \eqref{eq:eqs_for_edges_finite_centroids} to the spectral edges $\lambda_{gen}, \lambda_{mem}^{(1)}, \lambda_{mem}^{(2)}$ defined in section~\ref{sec:timescale_separation}. Similarly to \cite{bonnaire2025diffusion}, we take $\chi_p>\chi_m\gg 1$ and consider different scaling regimes for the resolvent function. We set $\chi_{pm}=\chi_p/\chi_m$.

\paragraph{First memorization bulk.} We consider the scaling $g_{\Psi}=\mathcal{O}_{\chi_p}(1)$ and $g_{\Omega}=\mathcal{O}_{\chi_p}\left(\frac{1}{\chi_p}\right)$:
let us set $\widetilde{g}_{\Omega}^* = \chi_p g_{\Omega}^*$.
At leading order in $1/\chi_p$, the equations for the edges become 
\begin{equation}
    \label{eq:edge_equations_scaling_I}
    \begin{gathered}
        \frac{1}{(g_{\Psi}^*)^2}\approx\chi_{pm}\sum_c \frac{b_c h_c^4}{\left(1+\chi_{pm}h_c^2g_{\Psi}^*\right)^2} \\
        \, \\
        \omega_* \approx \frac{1}{\chi_m}\sum_c \frac{b_c \gamma_c^2 h_c^2 \widetilde{g}_{\Omega}^*}{\left(1+\chi_{pm}h_c^2g_{\Psi}^*\right)^2} = \mathcal{O}_{\chi_m}\left(\frac{1}{\chi_m}\right) \\
        \, \\
        \frac{1}{\widetilde{g}_{\Omega}^*} \approx \frac{\Delta}{e^{-2t}}\sum_c \frac{b_c}{\sigma_c^2}\gamma_c^2 + \sum_c \frac{b_c \gamma_c^2}{1+\chi_{pm}h_c^2g_{\Psi}^*} \\
        \, \\
        \lambda_* \approx \sum_c b_c \varsigma_c
        +\sum_c \frac{b_c h_c^2}{1+\chi_{pm}h_c^2g_{\Psi}^*}
        -\frac{1}{g_{\Psi}^*}
        -\frac{\widetilde{g}_\Omega^*}{(g_\Psi^*)^2}
    \end{gathered}
\end{equation}

In this limit, the equation for $g_{\Psi}^*$ is decoupled from that for $g_{\Omega}^*$. We observe that the equation for $g_{\Psi}^*$ can be rearranged as an algebraic equation of order $2C$, thus yielding at most $2C$ values of spectral edges for $C$ bulks. This regime identifies the $t$-independent memorization phase: the eigenvalue $\lambda_{mem}^{(1)}$ of section~\ref{sec:timescale_separation} is obtained from the solution $g_{\Psi}^*$ that returns the largest spectral edge. This bulk has $M$ eigenvalues, since it originates from the memorization term $\mathbf{\delta U}_{gep}$ in Eq.~\eqref{eq:isolate_gen_mem}. Thus, it carries a weight $\chi_m/\chi_p$.

Note that from the first of \eqref{eq:edge_equations_scaling_I} the scaling $g_{\Psi}^*\propto 1/\chi_{pm}$ emerges. Then, for $t\ll 1$ one finds that all the edges in this scaling regime satisfy $\lambda_*\propto \chi_{pm}$ and
\begin{equation}
\label{eq:scaling_tau_mem_1}
    \tau_{mem}^{(1)}=\frac{1}{2\lambda_{mem}^{(1)}}\propto \frac{\chi_m}{\chi_p}
\end{equation}
as proven in \cite{bonnaire2025diffusion} for the unimodal case.

\paragraph{Generalization bulk.} Scaling $g_{\Psi}=\mathcal{O}_{\chi_p}(\frac{1}{\chi_p})$ and $g_{\Omega}=\mathcal{O}_{\chi_p}\left(\frac{1}{\chi_p}\right)$: we set $\widetilde{g}_{\Psi}^* = \chi_p g_{\Psi}$. We can find the spectral edges more easily from the spectral equations \eqref{eq:saddle_point_equations_finite_centroids}, which for this scaling read
\begin{equation}
\begin{gathered}
    \frac{\chi_p}{\widetilde{g}_{\Psi}^*}=-\frac{\widetilde{g}_{\Omega}^*}{(\widetilde{g}_{\Psi}^*)^2}+(\varsigma-z)+\sum_c b_c h_c^2 \\
    \, \\
    \frac{1}{\widetilde{g}_{\Omega}^*} = 
    \frac{\Delta}{e^{-2t}}\sum_c \frac{b_c}{\sigma_c^2}\gamma_c^2
    + \frac{1}{\widetilde{g}_{\Psi}^*}+\sum_c b_c \gamma_c^2
\end{gathered}
\end{equation}
where we set $\varsigma = \sum_c b_c \varsigma_c$.
Combining the two equations above, one finds the spectral equation of a Wishart matrix, and therefore the edges are those of a Marchenko--Pastur distribution \cite{potters2020first}:
\begin{equation}
    \lambda_{\pm}^{(g)} \approx \varsigma+\sum_c b_c h_c^2 + \left(\frac{\Delta}{e^{-2t}}\sum_c \frac{b_c}{\sigma_c^2}\gamma_c^2+\sum_c b_c \gamma_c^2\right)(1\pm \sqrt{\chi_p})^2
\end{equation}
These are the edges of the rightmost bulk shown in fig. \ref{fig:timescale_first}, which counts $N$ eigenvalues and carries a weight $1/\chi_p$, since it is related to the population correlation $\widetilde{\mathbf{U}}_{gep}$ (Eq.~\eqref{eq:U_tilde_GEP} and see also Eq.~\eqref{eq:isolate_gen_mem}). It represents the generalization phase of the training: thus, we identify $\lambda_{gen}=\lambda_{-}^{(g)}$. We then have for $t\ll 1$
\begin{equation}
\label{eq:scaling_tau_gen}
    \tau_{gen} = \frac{1}{2\lambda_{gen}}\propto \frac{1}{\chi_p}
\end{equation}
Combining \eqref{eq:scaling_tau_mem_1}, \eqref{eq:scaling_tau_gen}, we find the result of \cite{favero2025bigger, bonnaire2025diffusion, montanari2025dynamical} for the generalization window
\begin{equation}
    w_g \equiv \frac{\tau_{mem}^{(1)}}{\tau_{gen}} \propto \chi_m,
\end{equation}
i.e., the generalization window grows linearly with the size of the dataset. Having verified the sample size scaling, in section~\ref{sec:timescale_separation} we study the dependence of the prefactor of this scaling law on the variances of the classes in a binary setting, to show how class diversity affects the generalization window.

\begin{figure}[H]
    \centering
    \includegraphics[width=0.9\linewidth]{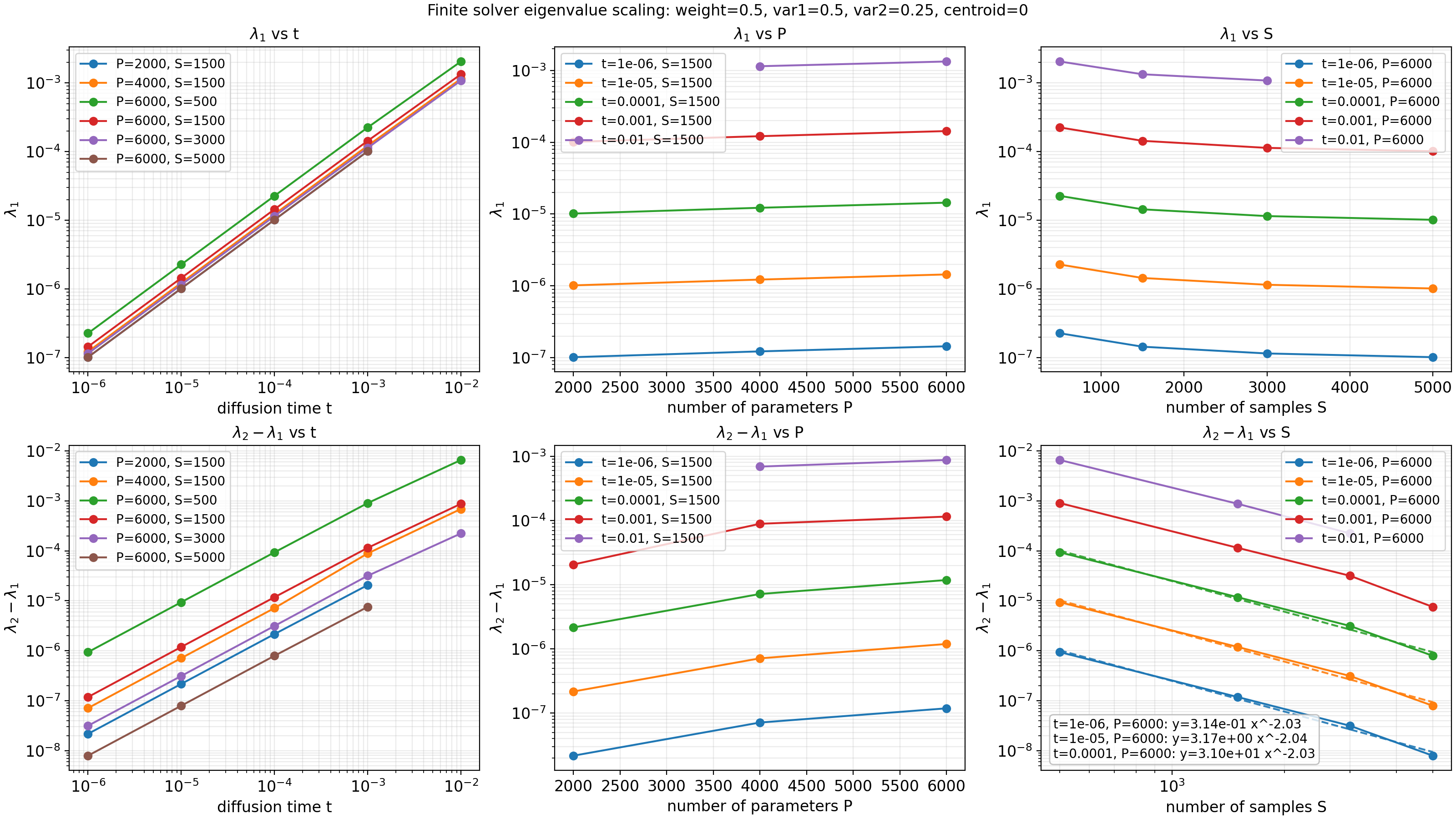}
    \caption{Check of the scaling of $\lambda_{mem}^{(2)}$.}
    \label{fig:scaling_lambda_mem_2}
\end{figure}

\paragraph{Second memorization bulk.} Scaling $g_{\Psi}=\mathcal{O}_{\chi_p}(\frac{\chi_m^2}{t \chi_p})$ and $g_{\Omega}=\mathcal{O}_{\chi_p}\left(\frac{\chi_m^2}{t\chi_p}\right)$: the choice of the scalings is motivated by fig. \ref{fig:scaling_lambda_mem_2}. 
We show that the eigenvalue $\lambda_{mem}^{(2)}$ scales linearly with $t$ and converges to a constant value for a large number of hidden neurons and number of samples. We also show the width $\Delta\lambda$ of the leftmost bulk in fig. \ref{fig:timescale_first} as a function of the same quantities: we find that it scales linearly with diffusion time and that it decays as the inverse square of the number of samples. We also find empirically that the weight of the leftmost bulk is $1/\chi_p$: then, considering our findings in fig. \ref{fig:scaling_lambda_mem_2}, we choose to scale the resolvent functions as above. 
We now proceed to expand the spectral equations \eqref{eq:saddle_point_equations_finite_centroids} for $t\ll 1$ and $\chi_m\gg 1$: we get the following
\begin{equation}
\begin{gathered}
    \frac{t \chi_{pm}}{\chi_m \widetilde{g}_{\Psi}^*} = 
    -\frac{t \widetilde{g}_{\Omega}^*}{\chi_m^2 (\widetilde{g}_{\Psi}^*)^2}
    +\varsigma-z
    +\frac{t}{\chi_m}\sum_c \frac{b_c h_c^2}{\gamma_c^2 \widetilde{g}_{\Omega}^* +h_c^2 \widetilde{g}_{\Psi}^*} \\
    \, \\
    \frac{t}{\chi_m^2 \widetilde{g}_{\Omega}^*} = 
    \frac{\Delta}{e^{-2t}}\sum_c \frac{b_c}{\sigma_c^2}\gamma_c^2
    +\frac{t}{\chi_m^2 \widetilde{g}_{\Psi}^*}
    +\frac{t}{\chi_m}\sum_c \frac{b_c \gamma_c^2}{\gamma_c^2 \widetilde{g}_{\Omega}^* +h_c^2 \widetilde{g}_{\Psi}^*}
\end{gathered}
\end{equation}
By summing these two equations together, we find for $t\ll 1$ and $\chi_m\gg 1$
\begin{equation}
    \widetilde{g}_{\Omega}^*\approx \frac{z \widetilde{g}_{\Psi}^*}{2\sum_c b_c \frac{\gamma_c^2}{\sigma_c^2}\, t}
\end{equation}
After some algebra, one finds that the edges are ($\widetilde{\varsigma}=t\,\varsigma$)
\begin{equation}
    \lambda_{\pm}^{(m)} = \widetilde{\varsigma}\,t+\frac{\chi_{pm}\, t}{4\chi_m^3\sum_c b_c \frac{\gamma_c^2}{\sigma_c^2}}\pm t\,\sqrt{\frac{\widetilde{\varsigma} \chi_{pm}}{2\chi_m^3\, \sum_c b_c \frac{\gamma_c^2}{\sigma_c^2}}
    -\frac{2\sum_c b_c \frac{\gamma_c^2}{\sigma_c^2}}{\chi_{m}^3}\sum_c b_c \frac{h_c^2}{\gamma_c^2}
    }
\end{equation}
In this last equation, we identify $\lambda_{mem}^{(2)}=\lambda_{-}^{(m)}$: we observe that it scales linearly in $t$. The bulk width $\lambda_{+}^{(m)}-\lambda_{-}^{(m)}$ scales linearly with $t$ and decays as $\chi_{pm}^{1/2}/\chi_{m}^{3/2}\propto \frac{1}{\chi_m^2}$.

We finally find that the scaling of the final memorization time is
\begin{equation}
    \tau_{mem}^{(2)}=\frac{1}{2\lambda_{mem}^{(2)}}\propto \frac{1}{t}
\end{equation}
and that the memorization window scales as
\begin{equation}
    w_m = \frac{\tau_{mem}^{(2)}}{\tau_{mem}^{(1)}}\propto \frac{\chi_p}{\chi_m\, t}.
\end{equation}
In section~\ref{sec:timescale_separation} we study the dependence of the prefactor of $w_m$ on the variances of the classes and their sampling weights.

The remaining weight is $1-\chi_m/\chi_p-2/\chi_p$ and is carried by a delta peak: it represents eigenvectors that are orthogonal to all rows of the random features and GEP noise matrices. These modes do not affect training. The delta is located at $\lambda=\varsigma \equiv \sum_c b_c (\widetilde{\beta}_c-\beta_c-\frac{\Delta}{\sigma_c^2 e^{-2 t}}\gamma_c^2)\propto t$.

\section{Theoretical curves of train and test errors}
\label{sec:analytical_prediction_errors}

For the reader's convenience, we rewrite the equations of the train and test errors in Eqs.~\eqref{eq:empirical_risk_score_denoising}, \eqref{eq:metrics}:

\begin{equation}
\label{eq:train_loss_appendix}
\begin{gathered}
    \mathcal{E}_{train}=\frac{1}{N M}\sum_{\nu=1}^{M} \Expected{\left\lVert \sqrt{\Delta_t} \mathbf{s}\left(\bm{x}_{\nu}e^{-t}
    +\sqrt{\Delta_t}\boldsymbol{\xi}\Bigl|\mathbf{A},\mathbf{W}
    \right) 
    +\boldsymbol{\xi}\right\lVert^2}{\boldsymbol{\xi}} \\
    \, \\
\end{gathered}
\end{equation}

\begin{equation}
\label{eq:test_loss_appendix}
\begin{gathered}
    \mathcal{E}_{test}=\sum_c b_c\, \mathcal{E}_{test}^{(c)} \\
    \, \\
    \mathcal{E}_{test}^{(c)}=\frac{1}{N}\Expected{\lVert \sqrt{\Delta_t} \mathbf{s}\left(\bm{x} e^{-t}
    +\sqrt{\Delta_t}\boldsymbol{\xi}|\mathbf{A},\mathbf{W}
    \right) 
    +\boldsymbol{\xi}\lVert^2}{\bm{x}\sim\mathcal{N}(\mathbf{m}_c, \sigma_c^2\mathbf{I}_N),\,\boldsymbol{\xi}}
\end{gathered}
\end{equation}

Training and test errors can be rewritten in the following form:
\begin{equation}
\label{eq:train_loss_mod}
    \mathcal{E}_{train}=
    1+
    \frac{\Delta_t}{N}\operatorname{Tr}(\mathbf{A}\mathbf{U}\mathbf{A}^{\intercal})
    +\frac{2 \sqrt{\Delta_t}}{N}\operatorname{Tr}(\mathbf{V}\mathbf{A}),
\end{equation}
\begin{equation}
\label{eq:test_loss_mod}
\begin{gathered}
    \mathcal{E}_{test}^{(c)}=
    1+
    \frac{\Delta_t}{N}\operatorname{Tr}(\mathbf{A}\widetilde{\mathbf{U}}_c\mathbf{A}^{\intercal})
    +\frac{2 \sqrt{\Delta_t}}{N}\operatorname{Tr}(\mathbf{V}_c\mathbf{A}) \\
    \\
    \simeq  \mathcal{E}_{train}+\frac{\Delta_t}{N}\operatorname{Tr}(\mathbf{A}(\widetilde{\mathbf{U}}_c-\mathbf{U})\mathbf{A}^{\intercal})+\frac{2\sqrt{\Delta_t}}{N}\operatorname{Tr}((\mathbf{V}_c-\mathbf{V})\mathbf{A}), \\
    \, \\
    \mathcal{E}_{test} = \mathcal{E}_{train} + \frac{\Delta_t}{N}\operatorname{Tr}(\mathbf{A}(\widetilde{\mathbf{U}}-\mathbf{U})\mathbf{A}^{\intercal}).
\end{gathered}
\end{equation}
By substituting into Eqs.~\eqref{eq:train_loss_mod}, \eqref{eq:test_loss_mod} the solution of the dynamics \eqref{eq:dynamics_readouts_solution} and by expanding everything in the basis of eigenvectors of $\mathbf{U}$, we derived equations for the train and test errors:
\begin{equation}
\label{eq:errors_semi_analytical}
\boxed{
\begin{gathered}
    \mathcal{E}_{train}(\tau)=1-\frac{\Delta}{N}\sum_{q=1}^P \frac{1-e^{-4\lambda_q\,\tau}}{\lambda_q}
    \left\lVert \sum_{c=1}^C \frac{b_c}{\Gamma_c}\boldsymbol{\omega}_c^{(q)}\right\lVert^2, \\
    \, \\
    \mathcal{E}_{test}(\tau)= \mathcal{E}_{train}(\tau)+\frac{\Delta}{N}\sum_{q, p} \frac{
    \left(
    1-e^{-2\lambda_q\,\tau}
    \right)
    \left(
    1-e^{-2\lambda_p\,\tau}
    \right)
    }{\lambda_q\lambda_p} \\
    \, \\
    \times\left( \sum_c b_c\frac{\boldsymbol{\omega}_c^{(q)}}{\Gamma_c} \right)^\intercal
    \left( \sum_c b_c\frac{\boldsymbol{\omega}_c^{(p)}}{\Gamma_c} \right)
    \left(\boldsymbol{\psi}_q^\intercal \widetilde{\mathbf{U}}\boldsymbol{\psi}_p-\lambda_p\delta_{q p}\right), \\
    \, \\
    \, \\
    \mathcal{E}_{test}^{(c)}(\tau)=\mathcal{E}_{train}(\tau)+\frac{\Delta}{N}\sum_{q, p} \frac{
    \left(
    1-e^{-2\lambda_q\,\tau}
    \right)
    \left(
    1-e^{-2\lambda_p\,\tau}
    \right)
    }{\lambda_q\lambda_p} \\
    \, \\
    \times\left( \sum_c b_c\frac{\boldsymbol{\omega}_c^{(q)}}{\Gamma_c} \right)^\intercal
    \left( \sum_c b_c\frac{\boldsymbol{\omega}_c^{(p)}}{\Gamma_c} \right)
    \left(\boldsymbol{\psi}_q^\intercal \widetilde{\mathbf{U}}_c\boldsymbol{\psi}_p-\lambda_p\delta_{q p}\right)
    \, \\
    \, \\
    -\frac{2 \Delta}{N}\sum_q \frac{
    1-e^{-2\,\lambda_q\,\tau}
    }{\lambda_q}
    \left(\sum_{c'} b_{c'}\frac{\boldsymbol{\omega}_{c'}^{(q)}}{\Gamma_{c'}} \right)^\intercal
    \left(
    \frac{\boldsymbol{\omega}_{c}^{(q)}}{\Gamma_{c}}
    -\left(\sum_{c'} b_{c'}\frac{\boldsymbol{\omega}_{c'}^{(q)}}{\Gamma_{c'}} \right)
    \right),
\end{gathered}
}
\end{equation}

where $\{\lambda_q, \boldsymbol{\psi}_q\}$ are eigenvalues and eigenvectors of $\mathbf{U}$, and we introduced $\boldsymbol{\omega}_c^{(q)}=\frac{1}{\sqrt{N}}\mathbf{W}^\intercal (\boldsymbol{\psi}_q\odot \widetilde{\boldsymbol{\gamma}}_c) $.
The formulas in Eqs.~\eqref{eq:errors_semi_analytical}, with the feature correlation matrices replaced by their GEP expressions, are our \emph{semi-analytical} curves for the train and test errors. We name them so because we need to numerically diagonalize the matrix $\mathbf{U}_{gep}$ to use them.
The curves for the test errors were used in sections~\ref{sec:timescale_separation} and~\ref{sec:class_specific_gen_mem} to represent our theoretical predictions.

For test errors, in general we cannot write asymptotic expressions in terms of the solution of the saddle point equations \eqref{eq:saddle_point_extensive_centroid}; we can only do that for the train error. We derived the following asymptotic formula for the train error, in the joint limits $N, P, M\rightarrow \infty$ with fixed ratios $\chi_p=P/N$ and $\chi_m=M/N$:
\begin{equation}
\label{eq:train_error_fully_analytical}
\boxed{
    \,\mathcal{E}_{train}(\tau)=1-\frac{\Delta \chi_p}{e^{-2t}}\sum_{c c'}\frac{b_c b_{c'}}{\sigma_c \sigma_{c'}} \int \frac{d\lambda}{\lambda}\left(1-e^{-4\lambda \tau}\right)\rho_{c c'}^{({\Omega})}(\lambda)
}
\end{equation}
where the spectral densities $\rho_{c c'}^{(\Omega)}$ are computed from the resolvent functions $g_{c c'}^{(\Omega)}$ in Eq.~\eqref{eq:saddle_point_extensive_centroid}:
\begin{equation*}
    \rho_{c c'}^{(\Omega)}(\lambda)=\frac{1}{\pi}
    \lim_{\epsilon\rightarrow 0_+}\operatorname{Im}
    g_{c c'}^{(\Omega)}(\lambda+i \epsilon)
\end{equation*}
We refer to Eq.~\eqref{eq:train_error_fully_analytical} as our \emph{fully analytical} formula for the train error.
We derived \eqref{eq:train_error_fully_analytical} from the first equation in \eqref{eq:errors_semi_analytical}, noting that the dot products $\omega_{c}(\lambda)\cdot \omega_{c'}(\lambda)$ featuring in \eqref{eq:errors_semi_analytical} can be computed from the imaginary part of $g_{\Omega}^{c c'}$. Indeed, since by definition one has
\begin{equation*}
\begin{gathered}
    g_{\Omega}^{c c'}(z) = \frac{1}{N^2}\sum_{q=1}^P \frac{\bm{\psi}(\lambda_q)^\intercal (\mathbf{W}^\intercal \odot \bm{1}_N\bm{\gamma}_{c'})(\bm{\gamma}_c\bm{1}_N^\intercal \odot \mathbf{W})\bm{\psi}(\lambda_q)}{\lambda_q-z} \\
    \equiv \frac{\sigma_c \sigma_{c'}e^{-2 t}}{\Gamma_c \Gamma_{c'}}\int d\lambda \frac{\rho(\lambda) \,\bm{\omega}_c(\lambda)\cdot \bm{\omega}_{c'}(\lambda)}{\lambda-z}
\end{gathered}
\end{equation*}
then it follows that
\begin{equation}
\label{eq:omega_dots}
   \boldsymbol{\omega}_c(\lambda)\cdot \boldsymbol{\omega}_{c'}(\lambda) = \frac{\Gamma_c \Gamma_{c'}}{\sigma_c \sigma_{c'}e^{-2 t}}\frac{\rho_{\Omega}^{c c'}(\lambda)}{\rho(\lambda)}.
\end{equation}
We indicate with $\bm{\omega}(\lambda)$ the continuous limit of the $\bm{\omega}_q$ defined above.
In the limit of subextensive centroids, $\lVert \bm{m}_c \lVert=\mathcal{O}_N(1)$, the vector Hermite coefficients become homogeneous and thus \eqref{eq:omega_dots} turns into
\begin{equation}
    \boldsymbol{\omega}_c(\lambda)\cdot \boldsymbol{\omega}_{c'}(\lambda) = \frac{\Gamma_c \Gamma_{c'}\gamma_c \gamma_{c'}}{\sigma_c \sigma_{c'}e^{-2 t}}\frac{\rho_{\Omega}(\lambda)}{\rho(\lambda)}
\end{equation}
and the analytical train error formula simplifies to
\begin{equation}
    \mathcal{E}_{train}(\tau)=1-\frac{\Delta \chi_p}{e^{-2t}}\left(\sum_{c=1}^C b_c \frac{\gamma_c}{\sigma_c}\right)^2 \int \frac{d\lambda}{\lambda}\left(1-e^{-4\lambda \tau}\right)\rho_{\Omega}(\lambda).
\end{equation}

\section{Score MSE}
\label{sec:Exact_score_and_score_MSE}

We write the formula for the MSE of the score:
\begin{equation}
    \label{eq:score_MSE_appendix}
    \mathcal{E}_{score}=\sum_c b_c \mathcal{E}_{score}^{(c)},\qquad \mathcal{E}_{score}^{(c)}=\frac{1}{N}\Expected{\norm{
    \mathbf{A}\,\phi\left(\frac{1}{\sqrt{N}}\mathbf{W}\bm{x}\right)
    -\bm{s}_{true}(\bm{x}, t)
    }^2}{\bm{x}\sim \mathcal{N}_c(t)}
\end{equation}
The MSE on the score is a proxy for the quality of the generative model: indeed, the Kullback-Leibler (KL) divergence between the prior distribution $\mathcal{P}^*$ and the sampling distribution learned by the diffusion model $\widetilde{\mathcal{P}}$ is upper-bounded by the integral over diffusion time of the score error \cite{song2021maximum}:
\begin{equation}
\label{eq:D_KL_upper_bounding}
    D_{KL}(\widetilde{\mathcal{P}}|\mathcal{P}^*)\equiv -\int d\bm{x}\mathcal{P}^*(\bm{x})\log(\widetilde{\mathcal{P}}(\bm{x})/\mathcal{P}^*(\bm{x}))\leq \frac{N}{2}\int dt \mathcal{E}_{score}(t)
\end{equation}
Given the hypothesis of well-separated classes, the condition in \eqref{eq:D_KL_upper_bounding} holds also for the single components of the mixture:
\begin{equation}
\label{eq:D_KL_upper_bounding_class_specific}
    D_{KL}(\widetilde{\mathcal{P}}_c|\mathcal{P}_c^*)\equiv -\int d\bm{x}\mathcal{P}_c^*(\bm{x})\log(\widetilde{\mathcal{P}}_c(\bm{x})/\mathcal{P}_c^*(\bm{x}))\leq \frac{N}{2}\int dt \mathcal{E}_{score}^{(c)}(t).
\end{equation}
We can compute semi-analytical formulas analogous to those in \eqref{eq:errors_semi_analytical} for the score MSEs.
The score of a Gaussian mixture such as that in \eqref{eq:pdf_data} has the following expression:
\begin{equation}
\label{eq:true_score_formula}
\begin{gathered}
    \bm{s}_{true}(\bm{x}, t)=\sum_{c=1}^C b_c \bm{s}_{true}^{(c)}(\bm{x}, t)\qquad  \bm{s}_{true}^{(c)}(\bm{x}, t)=-\frac{\bm{x}-e^{-t}\bm{m}_c}{\Gamma_c^2(t)}\,\zeta_c(\bm{x}, t)\\
    \, \\
    \zeta_c(\bm{x}, t)=\frac{\mathcal{P}_c(\bm{x}, t)}{\sum_{c'=1}^C b_{c'}\mathcal{P}_{c'}(\bm{x}, t)}
    ,\qquad \mathcal{P}_c(\bm{x}, t)=\frac{e^{-\frac{\norm{\bm{x}-e^{-t}\bm{m}_c}^2}{2\Gamma_c^2(t)}}}{(2\pi \Gamma_c^2(t))^{N/2}}
\end{gathered}
\end{equation}
For generic mixtures, the components of the true score interfere with each other through the functions $\zeta_c$. In the case of well-separated mixtures,  one gets $\zeta_c=1/b_c$ for any datapoint sampled from that class: then, the scores are the scores that the components of the mixture would have in isolation,
\begin{equation}
    b_c\mathbf{s}_{true}^{(c)}(\bm{x}, t)\approx -\frac{\bm{x}-\mathbf{m}_c e^{-t}}{\Gamma_c^2(t)}\equiv \bm{s}_c(\bm{x}, t)\qquad \bm{x}\text{ belonging to class }c.
\end{equation}
We can define the effective domain for classes by looking at Eqs.~\eqref{eq:true_score_formula}: all exponential terms should be vanishing for classes $c'\neq c$ compared to class $c$, and this is achieved for values of $\bm{x}$ such that
\begin{equation}
\label{eq:well_separated_mixtures_inequalities_Gaussian}
    \frac{1}{\Gamma_{c'}^2(t)N}\lVert \bm{x} -\mathbf{m}_{c'} e^{-t}\lVert^2-\frac{1}{\Gamma_c^2(t)N}\lVert \bm{x} -\mathbf{m}_c e^{-t}\lVert^2+\log(\Gamma_{c'}/\Gamma_c)>0,\qquad \forall c'\neq c, N\rightarrow \infty
\end{equation}
The boundaries of the effective domains are regions such that
\begin{equation}
    \left|\frac{1}{\Gamma_{c'}^2(t)}\lVert \bm{x} -\mathbf{m}_{c'} e^{-t}\lVert^2-\frac{1}{\Gamma_c^2(t)}\lVert \bm{x} -\mathbf{m}_c e^{-t}\lVert^2+N\log(\Gamma_{c'}/\Gamma_c)\right|=\mathcal{O}_N(1)\qquad N\rightarrow \infty
\end{equation}
In the high-dimensional limit $N\gg 1$, samples $\bm{x}$ belong with high probability to the bulks of the effective domains, the boundaries are atypical.
We consider the case of well-separated mixtures, where the inequalities \eqref{eq:well_separated_mixtures_inequalities_Gaussian} hold for all classes: then the score MSE in \eqref{eq:score_MSE_appendix} reduces to
\begin{equation}
    \mathcal{E}_{score}=\sum_c b_c \mathcal{E}_{score}^{(c)},\qquad \mathcal{E}_{score}^{(c)}\simeq \frac{1}{N}\Expected{\norm{
    \mathbf{A}\,\phi\left(\frac{1}{\sqrt{N}}\mathbf{W}\bm{x}\right)
    +\frac{\bm{x}-e^{-t}\bm{m}_c}{\Gamma_c^2(t)}
    }^2}{\bm{x}\sim \mathcal{N}_c(t)}
\end{equation}
\begin{figure}[htbp]
    \centering
    \includegraphics[width=\linewidth]{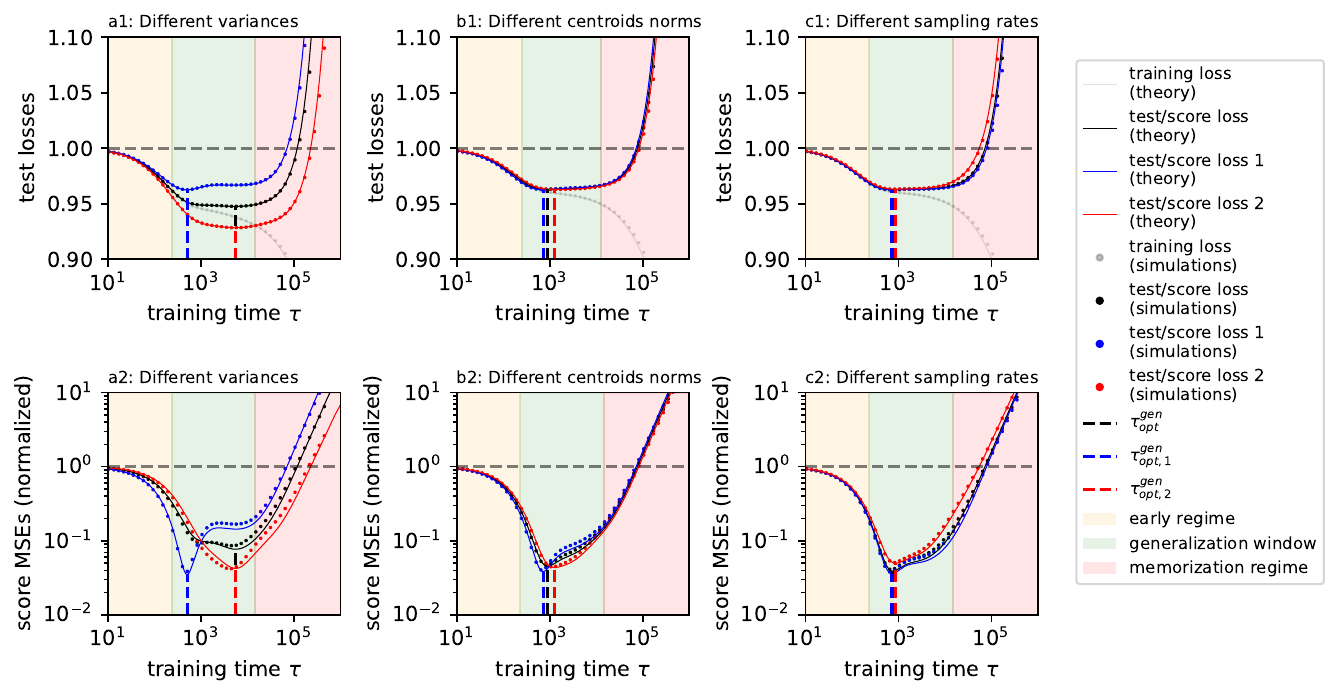}
    \caption{Comparison between the test losses and the score MSEs. Parameters as in figure \ref{fig:test_error_ablations}.}
    \label{fig:scoreMSEs_AND_test_losses}
\end{figure}

Finally, by following the steps of Appendix~\ref{sec:analytical_prediction_errors}, we obtain
\begin{equation}
\label{eq:semi_analytical_score_errors}
\boxed{
    \begin{gathered}
        \, \\
        \mathcal{E}_{score}^{(c)}(\tau) =
        \frac{1}{\Gamma_c^2(t)}
        -\frac{2}{\Gamma_c N}\sum_q \frac{1-e^{-2\lambda_q \tau}}{\lambda_q}\left(\sum_{c''}\frac{b_{c''}}{\Gamma_{c''}}\boldsymbol{\omega}_{c''}^{q}\right)^\intercal \boldsymbol{\omega}_{c}^{q}
         \\
        \, \\
       \,+ \frac{1}{N}\sum_{p q} \frac{(1-e^{-2\lambda_p\tau})(1-e^{-2\lambda_q\tau})}{\lambda_p \lambda_q}
        \left(
         \sum_{c''} \frac{b_{c''}}{\Gamma_{c''}}
        \boldsymbol{\omega}_{c''}^{p}
        \cdot
        \sum_{c'''} \frac{b_{c'''}}{\Gamma_{c'''}}
        \boldsymbol{\omega}_{c'''}^{q}
        \right)
        \boldsymbol{\psi}_p^\intercal \mathbf{\widetilde{U}}_{c}\boldsymbol{\psi}_q\, \\
        \, \\
        \equiv \frac{1}{\Gamma_c^2(t)}+\frac{1}{\Delta_t}(\mathcal{E}_{test}^{(c)}(\tau)-1) \\
        \,
    \end{gathered}
}
\end{equation}
We verified that in the limit of well-separated classes, the score MSE of any class $c$ is just the test loss of that class plus a diffusion time and a class-dependent constant shift, along with a rescaling multiplicative factor dependent on diffusion time, in agreement with Tweedie's identity. This implies that the class-specific generalization times and memorization times defined in \eqref{eq:class_specific_gen_mem_time} for the test losses are the same for the score MSEs.
In figure \ref{fig:scoreMSEs_AND_test_losses} we compare the test loss shown in figure \ref{fig:test_error_ablations} with the score MSE.

\clearpage
\phantomsection
\addcontentsline{toc}{part}{Appendices for experiments}
{\Large \bfseries Appendices for experiments}

\vspace{0.5em}

\section{Numerical experiments on Random Features}
\label{sec:RF_experiments}

\subsection{Training of RF model}

Training of the RF model was conducted using gradient descent: for $k=1,\dots, N_{epoch}$, we used
\begin{equation*}
    \mathbf{A}_{k+1} = \mathbf{A}_{k} - \eta \frac{\partial \mathcal{E}_{train}}{\partial \mathbf{A}}\Biggl |_{\mathbf{A}=\mathbf{A}_k}
\end{equation*}
with $\mathcal{E}_{train}$ given by Eq.~\eqref{eq:empirical_risk_score_denoising}.
In all experiments, we set $\eta = 5\times 10^{-5} \frac{N}{\Delta_t}$, where $N$ is data dimension, and $\Delta_t=1-e^{-2 t}$ (compare with Eq.~\eqref{eq:dyn} to understand the origin of this scaling choice of the learning rate). We set $N_{epoch}=2\times 10^6$ and $600$ log-spaced sampling times. We fix $P=6000$, $N=100$ and $M=3000$ for the parameters in sections~\ref{sec:timescale_separation},~\ref{sec:class_specific_gen_mem}.

For each choice of data distribution parameters, we conducted several runs with different instantiations of the RF model and the dataset, namely different draws of the matrix $\mathbf{W}$ and the dataset $\mathbf{X}$, and then obtained curves for train, test and score errors versus training time by averaging over different runs. We applied a Gaussian smoothing to each error curve of each run, to smooth out fluctuations at small scales due to GD discrete steps.
Generalization and memorization times were computed using \eqref{eq:class_specific_gen_mem_time} on the curves averaged over different runs.
As the dynamics converge to a unique readout matrix, asymptotically independent of the initial condition (see Eq.~\eqref{eq:dyn}), we did not average over different initial conditions.

An experiment with 5 runs, with the parameters specified above, requires approximately four hours and a half on an $A40$ GPU.

\subsection{Numerical diagonalization of $\mathbf{U}_{gep}, \widetilde{\mathbf{U}}_{gep}$}

To evaluate the semi-analytical curves in Eqs.~\eqref{eq:errors_semi_analytical},\eqref{eq:semi_analytical_score_errors}, for each choice of Gaussian mixture, we obtained the eigenvalues and eigenvectors of $N_{diag}=10$ instances of the GEP matrices $\mathbf{U}_{gep}$ and $\widetilde{\mathbf{U}}_{gep}$. In sections~\ref{sec:timescale_separation},~\ref{sec:class_specific_gen_mem}, we set $P=6000$ for the number of hidden neurons, $N=100$ for data dimension and $M=3000$ for the number of samples. In section~\ref{sec:Numerical_Experiments_Fashion_MNIST}, we set $P=10000$, $N=100$, $M=500$.

The numerical diagonalization of instances of $\mathbf{U}_{gep}, \widetilde{\mathbf{U}}_{gep}$ requires roughly 10 minutes on a state-of-the-art CPU.

\subsection{Solution of the spectral equations}

\begin{algorithm}[H]
\caption{Solver of Eqs.~\eqref{eq:saddle_point_equations_finite_centroids}}
\label{alg:solver}
\begin{algorithmic}[1]
\Require
  Hermite coefficients $\{(\beta_k,\tilde\beta_k,\gamma_k)\}_{k=1}^C$,
  diagonal terms $d_k = \tilde\beta_k - \beta_k - \frac{\Delta}{2 e^{-2 t}}\gamma_k^2$,
  grid $\{\lambda_j\}_{j=1}^{N_{\mathrm{values}}}$,
  continuation schedule $\mathcal{E} = \operatorname{Concat}(\operatorname{Log-Spaced}([100, 10^{-9}], N_{eps}),\{0\}]$.
\Ensure
  real parts $\{\operatorname{Re}g_{\Psi}(\lambda_j),\,\operatorname{Re}g_{\Omega}(\lambda_j)\}$
  and spectral densities $\{\rho(\lambda_j),\,\rho_\Omega(\lambda_j)\}$

\vspace{3pt}
\For{$j = 1$ \textbf{to} $N_{\mathrm{values}}$}               \Comment{sweep over spectral grid}

  \State $\lambda \gets \lambda_j$
  \State $z \gets \lambda + i\,\varepsilon_0$  \Comment{initialize at largest imaginary shift}

  \State $g_{\Psi}^{(0)} \gets -1/z$, \quad $g_{\Omega}^{(0)} \gets -1/z$
  \Comment{free-theory initial guess}

  \For{$\varepsilon \in \mathcal{E}$}          \Comment{decrease imaginary part to 0}
    \State $z \gets \lambda + i\,\varepsilon$

    \State $(g_{\Psi},g_{\Omega}) \gets
      \textsc{Solve}\!\left(F_\Psi(\cdot,\cdot;\,z)=0,\;
                            F_\Omega(\cdot,\cdot;\,z)=0,\;
                            \text{init}=(g_{\Psi}^{(0)},g_{\Omega}^{(0)})\right)$
    \Comment{Levenberg–Marquardt}

    \State $g_{\Psi}^{(0)} \gets g_{\Psi}$, \quad $g_{\Omega}^{(0)} \gets g_{\Omega}$
    \Comment{warm start for next $\varepsilon$}
  \EndFor

  \State $\operatorname{Re}g_{\Psi}(\lambda_j) \gets \operatorname{Re}(g_{\Psi})$
  \State $\rho(\lambda_j) \gets \tfrac{1}{\pi}\,\operatorname{Im}(g_{\Psi})$
  \State $\operatorname{Re}g_{\Omega}(\lambda_j)  \gets \operatorname{Re}(g_{\Omega})$
  \State $\rho_\Omega(\lambda_j)  \gets \tfrac{1}{\pi}\,\operatorname{Im}(g_{\Omega})$

\EndFor
\end{algorithmic}
\end{algorithm}

To solve Eqs.~\eqref{eq:saddle_point_equations_finite_centroids}, we used the following algorithm: for each of $10^4$ log-spaced values of $\lambda\in [10^{-8}, 10^2]$, we took $z=\lambda+i\varepsilon$ and iteratively solved \eqref{eq:spectral_edges_finite_centroids} for decreasing values of $\varepsilon$ until $\varepsilon=0$, starting from $\varepsilon_{0}=100$ with initial conditions $g_{start}(z)=-1/z$ and $g_{\Omega}^{(start)}=-1/z$. 
The spectral densities $\rho(\lambda), \rho_{\Omega}(\lambda)$ are evaluated with formula \eqref{eq:pdf_eigenvalues_theor}. Spectral density values below $10^{-6}$ are truncated to zero.

The choice of $\varepsilon_{0}\gg 1$ ensures the selection of the correct branch of $g(z)$ (in random matrix theory, the resolvent function is a multivalued function featuring different branches, see \cite{potters2020first}), uniquely identified by $g(z)\approx -1/z$ for $|z|\gg 1$.

To solve Eqs.~\eqref{eq:eqs_for_edges_finite_centroids} to compute the edges through \eqref{eq:spectral_edges_finite_centroids}, we computed the initial conditions from the solution of \eqref{eq:spectral_edges_finite_centroids}, by selecting values of $\lambda$ in the grid where $\rho(\lambda)$ acquires a nonzero value.

In algorithm \ref{alg:solver} we report pseudocode for the solver algorithm. The solver takes around 3 minutes on a state-of-the-art CPU to solve the equations on a grid of $N_{values}=10^4$ values of $\lambda$. 

\section{Numerical experiments on Fashion MNIST}

\subsection{Additional simulations framework details}\label{sec:more_numerical_info}

\paragraph{Model.} A standard DDPM \cite{ho2020denoising} was employed on the FashionMNIST dataset \cite{xiao2017fashion}, with a U-Net \cite{ronneberger2015u} serving as the backbone. The network follows a U-shaped encoder–decoder structure connected by skip connections across four resolution levels (64, 128, 256, 256 channels, respectively).
Each level contains ResNet blocks \cite{he2016deep} with a dropout rate of 0.1 \cite{srivastava2014dropout}, conditioned on the diffusion time step through sinusoidal time embeddings \cite{vaswani2017attention, ho2020denoising} injected additively into each block. The encoder is composed of two such blocks per level, and the decoder of three. Four-head self-attention with GroupNorm pre-normalization is applied within the ResNet blocks at the two intermediate resolutions. At the deepest level, a two-block bottleneck bridges the encoder and the decoder, with self-attention enabled in the first block and disabled in the second.

\paragraph{Training.} Each model is trained on a binary subset of FashionMNIST: the Sneaker class, that exhibits the lowest pixel-wise variance, is paired with each of the other classes, yielding nine class pairs in total. Each pair is trained at three Sneaker proportions (25\%, 50\%, 75\%) on 8,000 images on a single core of an NVIDIA A100 GPU for approximately one week, reaching approximately $4\times 10^6$ gradient steps. In this study, a linear noise schedule is adopted, with the parameters ranging from $\beta_1=10^{-4}$ to $\beta_T=2\times 10^{-2}$, and $T=1000$ discrete steps. The Adam optimizer ($\eta=10^{-4}$) is utilized to minimize the mean squared error (MSE) between the predicted and the true noise \cite{kingma2014adam}. $32$ model checkpoints are saved at log-spaced intervals ($5$ per scale) per run to capture both early and late training dynamics.

\paragraph{Generation.} 500 samples per pair were generated with the DDPM ancestral sampler \cite{ho2020denoising}. All checkpoints in one to two hours with a single NVIDIA A100 or A40 GPU are included. Starting from $\bm{x}_T \sim \mathcal{N}(\bm{0}, \mathbf{I}_N)$, denoising through all $T$ reverse steps was iteratively performed by the model. At each step $t \in \{T, \ldots, 1\}$, the update is
 \begin{equation*}
     \bm{x}_{t-1} = \frac{1}{\sqrt{\alpha_t}}\!\left(\bm{x}_t - \frac{\beta_t}{\sqrt{1-\bar{\alpha}_t}}\,\bm{\varepsilon}_\theta(\bm{x}_t,t)\right) + \sqrt{\beta_t}\,\bm{z}_{t},
 \end{equation*}
 where $\bm{z}_t \sim \mathcal{N}(\bm{0}, \mathbf{I}_N)$ for $t > 1$ and $z_t = 0$ for $t = 1$, $\bar{\alpha}_t = \prod_{s=1}^{t} \alpha_{s} = \prod_{s=1}^{t}(1 - \beta_s)$ is the cumulative product, and $\bm{\varepsilon}_\theta(\bm{x}_t, t)$ is the U-Net's predicted noise. Examples of generated samples can be found in Appendix~\ref{sec:more_figures}.

\paragraph{Evaluation.} In accordance with the nearest-neighbor criterion outlined in~\cite{yoon2023diffusion,gu2023memorization,bonnaire2025diffusion}, a generated sample $\bm{x}_c$ from class $c$ is deemed memorized if
\begin{equation*}
    r(\bm{x}_c)
    =
    \frac{\|\bm{x}_c - \mathrm{nn}_1(\bm{x}_c)\|_2}
    {\|\bm{x}_c - \mathrm{nn}_2(\bm{x}_c)\|_2}
    < \frac{1}{3},
\end{equation*}
where $\mathrm{nn}_i(\bm{x}_c)$ denotes the $i$-th nearest neighbor of $\bm{x}_c$ in the training set under the $\ell_2$ norm. The per-class memorization fraction is defined as the expected memorization rate under the estimated data distribution $p_\theta^c(t)$ at time $t$,
\begin{equation*}
    f_\mathrm{mem}^c(t)
    =
    \mathbb{E}_{\bm{x}_c \sim p_\theta^c(t)}
    \!\left[
    \mathbb{I}\!\left(r(\bm{x}_c)<\tfrac{1}{3}\right)
    \right],
\end{equation*}
and is empirically estimated by its mean over $n_c$ generated samples. We define the threshold-crossing time $t_{\mathrm{mem}}^c = \inf\{t:f_{\mathrm{mem}}^c(t)\ge 1/3\},$ so the Sneaker-partner memorization gap
\begin{equation}
    \Delta t_{\mathrm{mem}}(c_p)
    =
    t_{\mathrm{mem}}^{\mathrm{snk}}
    -
    t_{\mathrm{mem}}^{c_p}
\end{equation}
is tracked across all log-spaced checkpoints and class ratios.

To assign each generated sample to its class, a dedicated binary ResNet-18 classifier~\cite{he2016deep} is trained for each of the nine Sneaker--partner pairs. The first convolutional layer is replaced with a single-channel counterpart to accommodate Fashion MNIST's grayscale input, and the network is trained with the same input preprocessing as the diffusion model (class-pair mean subtracted, unit standard deviation). Training uses AdamW~\cite{loshchilov2017decoupled} with cosine annealing~\cite{loshchilov2016sgdr} over 15 epochs, differential learning rates (backbone: $10^{-5}$, head: $10^{-4}$)~\cite{howard2018universal, devlin2019bert}, weight decay $0.01$~\cite{krogh1991simple}, and label smoothing $0.1$~\cite{szegedy2016rethinking}. The classifiers used for class assignment are trained on a 90/10 train--validation split; using an 80/20 split led to no material difference. Across all class proportions, all nine classifiers achieved above $96.0\%$ macro F1 and $97.3\%$ balanced accuracy. The reported lowest values were observed for the visually similar Sneaker–Ankle Boot pair, confirming reliable class attribution of generated samples under the primary imbalance conditions examined.

\subsection{Supplementary figures}\label{sec:more_figures}

\begin{figure}[H]
    \centering
    \includegraphics[width=0.95\linewidth]{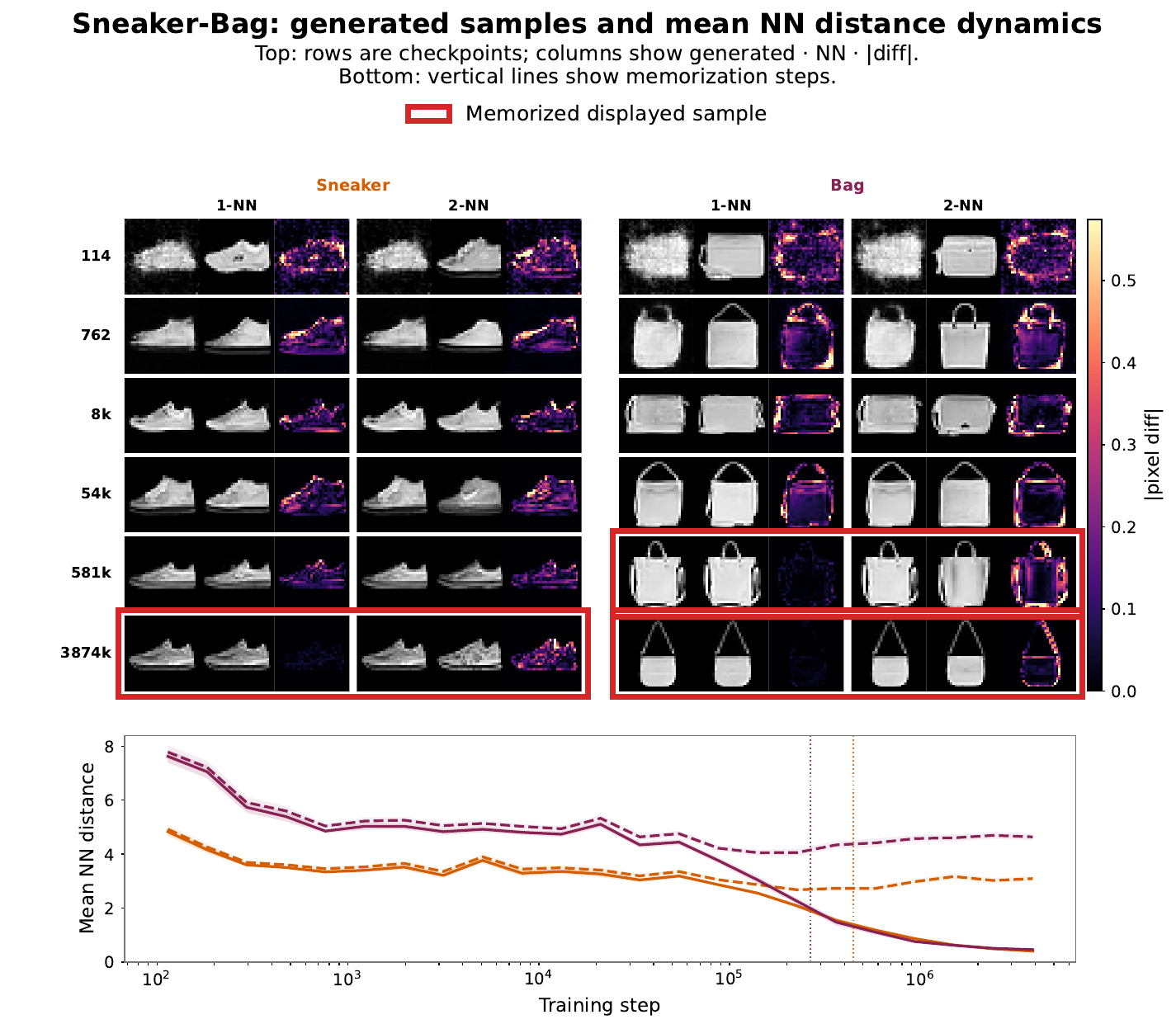}
    \caption{Overview examples for the Sneaker and Bag classes.}
    \label{fig:bag-overview}
\end{figure}

\paragraph{Generated samples and memorization dynamics.}
We first visualize a representative Sneaker--Bag pair in Fig.~\ref{fig:bag-overview} to illustrate the transition from generalization to memorization. The figure shows generated samples at selected checkpoints alongside their first and second nearest training examples, providing an intuitive view of the sample-level memorization criterion defined in section~\ref{sec:Numerical_Experiments_Fashion_MNIST}. Under this criterion, the memorized sample is highlighted in red. The bottom panel traces the corresponding nearest-neighbor dynamics: solid and dashed curves denote the mean distances to the first and second nearest neighbors, respectively, while vertical dotted lines mark the class-specific memorization times. Figs.~\ref{fig:tshirt-overview} to~\ref{fig:ankle-boot-overview} report the same analysis for the remaining eight Fashion MNIST class pairs.

\begin{figure}[H]
    \centering
    \begin{subfigure}[t]{\linewidth}
        \centering
        \includegraphics[width=\linewidth]{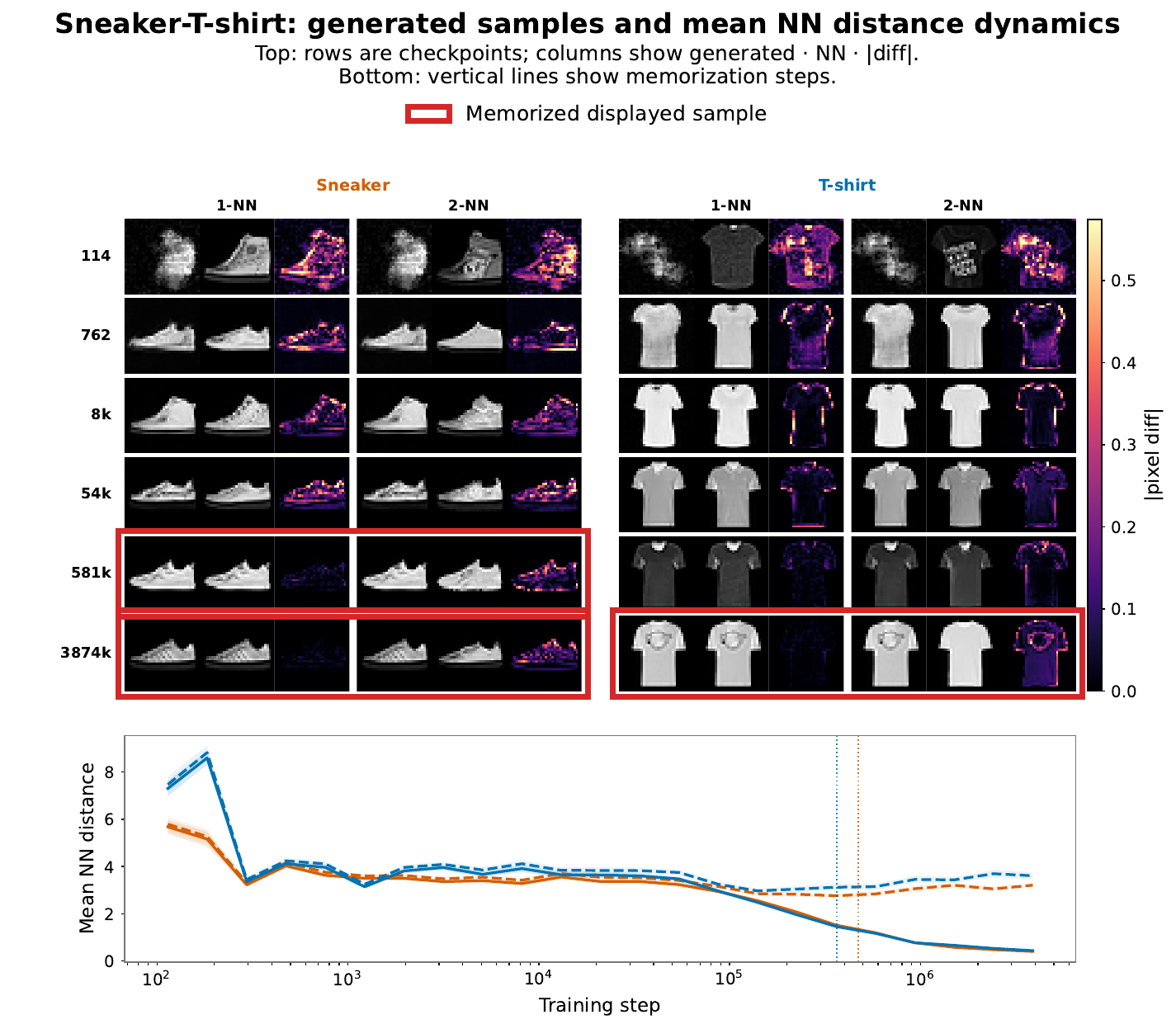}
        \caption{T-shirt/top}
        \label{fig:tshirt-overview}
    \end{subfigure}
    
    \caption{Overview examples for Fashion MNIST classes (Part 1).}
    \label{fig:fashion-mnist-overview-main}
\end{figure}
\begin{figure}[H]
    \ContinuedFloat 
    \centering
    \begin{subfigure}[t]{\linewidth}
        \centering
        \includegraphics[width=\linewidth]{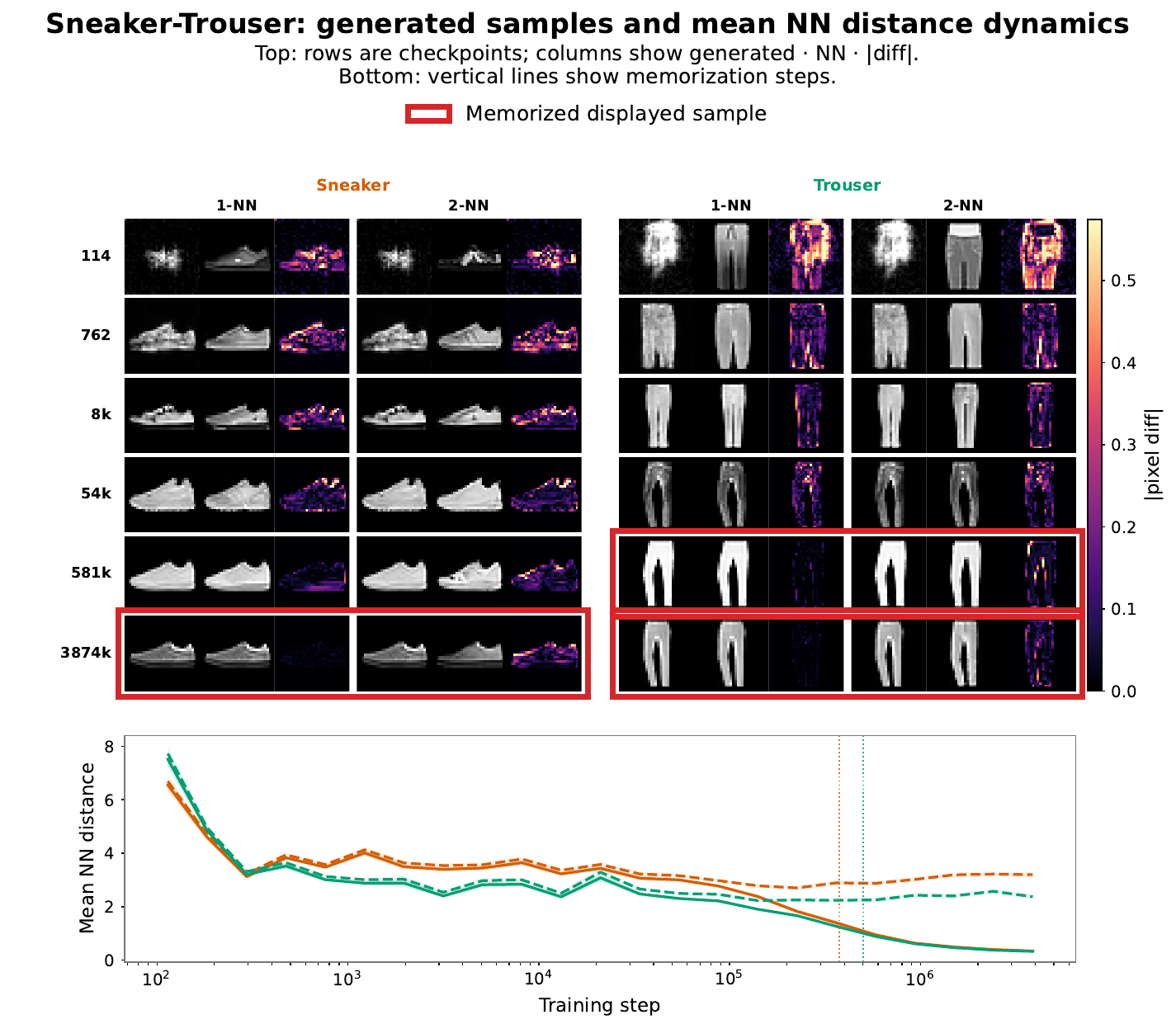}
        \caption{Trouser}
        \label{fig:trouser-overview}
    \end{subfigure}
    
    \caption{Overview examples for Fashion MNIST classes (Part 2).}
\end{figure}
\begin{figure}[H]
    \ContinuedFloat 
    \centering
    \begin{subfigure}[t]{\linewidth}
        \centering
        \includegraphics[width=\linewidth]{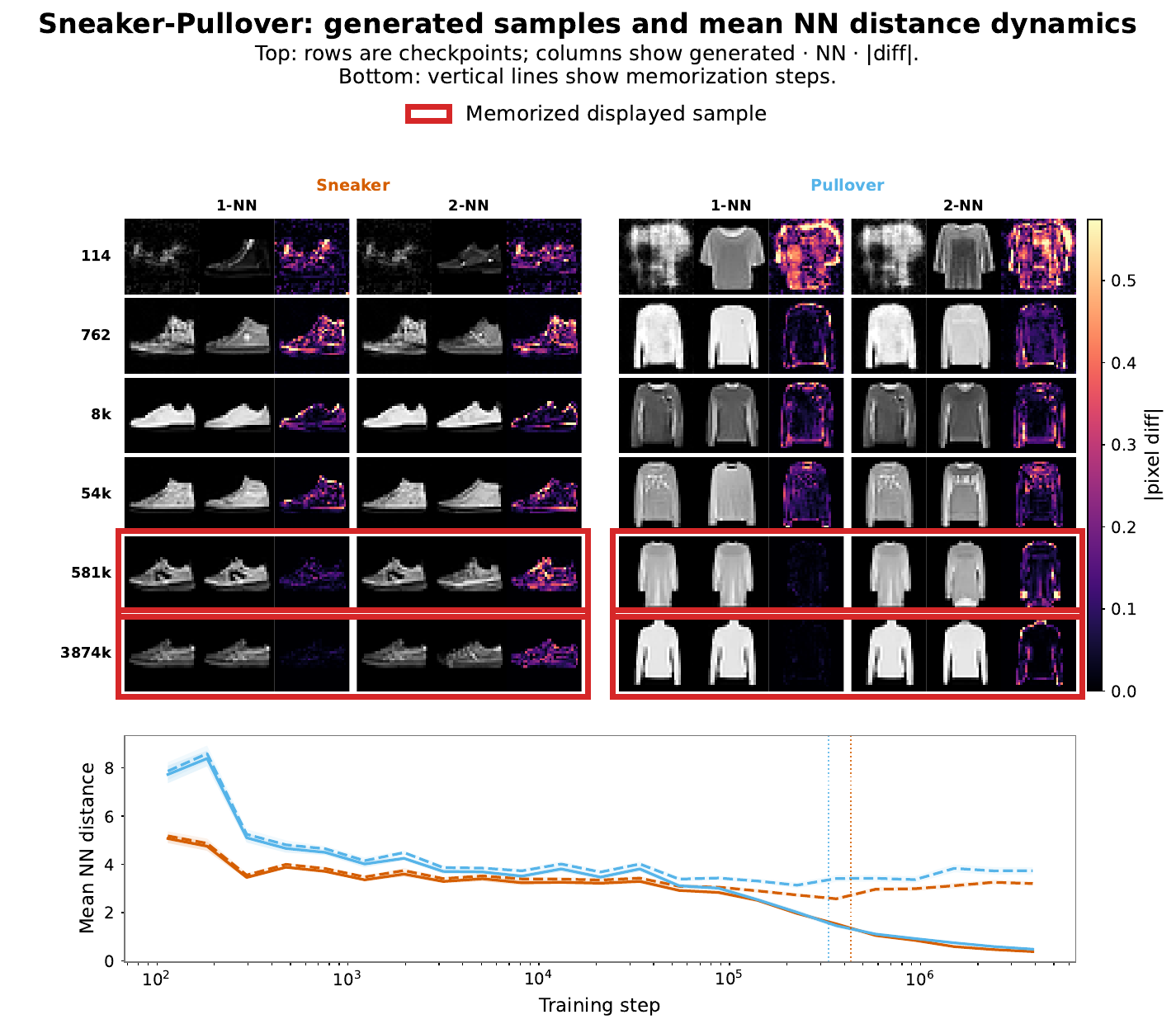}
        \caption{Pullover}
        \label{fig:pullover-overview}
    \end{subfigure}
    
    \caption{Overview examples for Fashion MNIST classes (Part 3).}
\end{figure}
\begin{figure}[H]
    \ContinuedFloat 
    \centering
    \begin{subfigure}[t]{\linewidth}
        \centering
        \includegraphics[width=\linewidth]{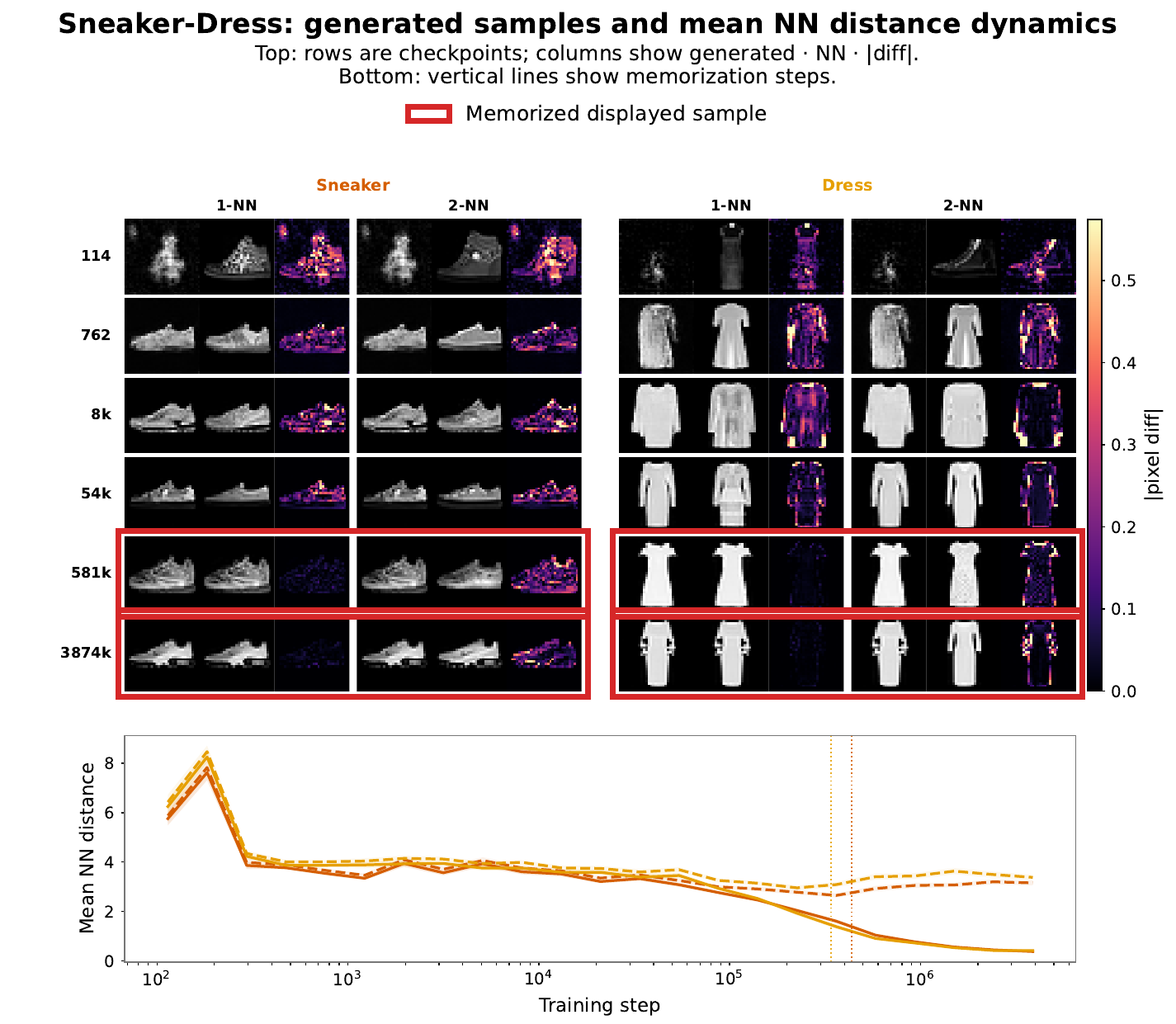}
        \caption{Dress}
        \label{fig:dress-overview}
    \end{subfigure}
    
    \caption{Overview examples for Fashion MNIST classes (Part 4).}
\end{figure}
\begin{figure}[H]
    \ContinuedFloat 
    \centering
    \begin{subfigure}[t]{\linewidth}
        \centering
        \includegraphics[width=\linewidth]{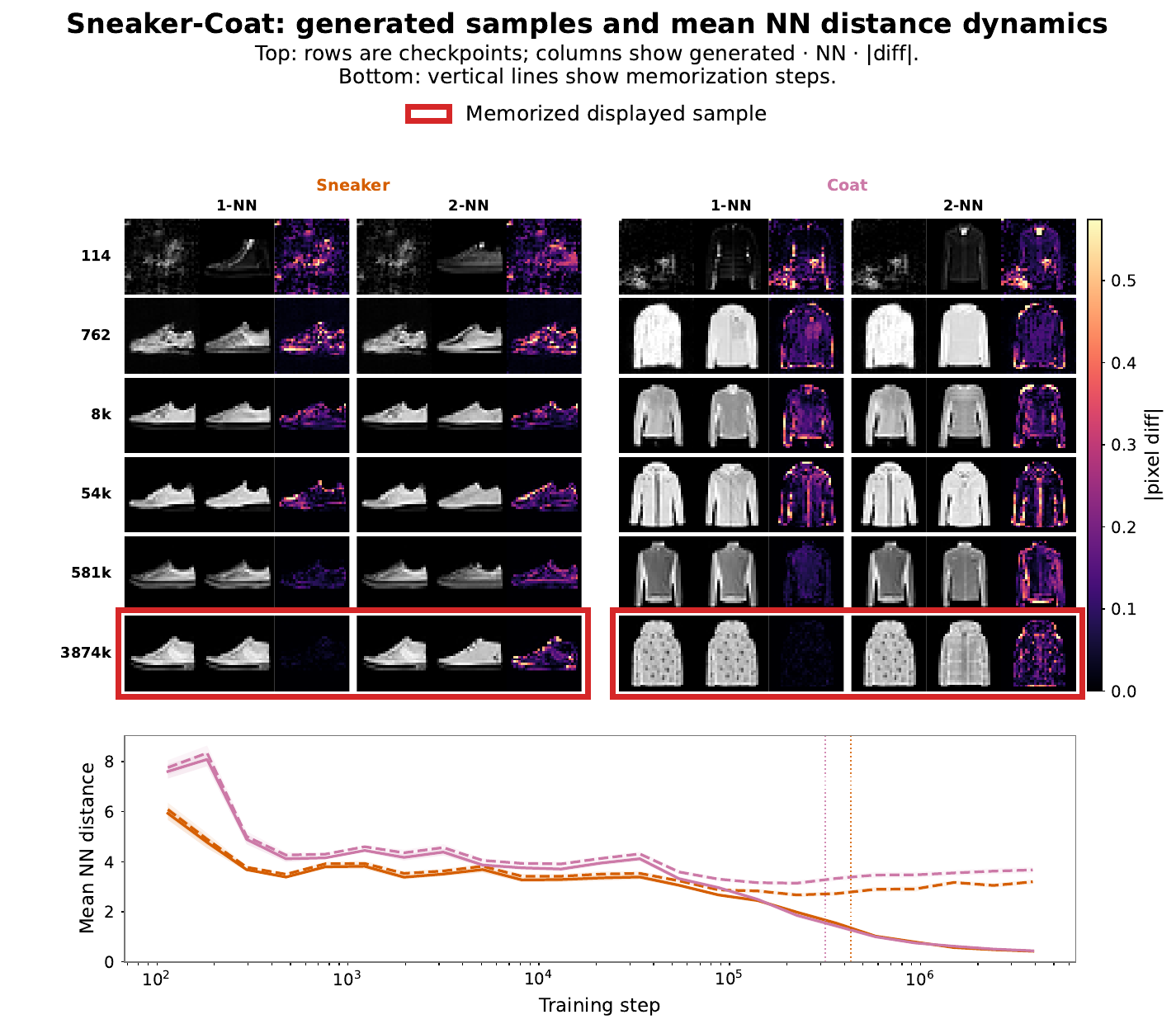}
        \caption{Coat}
        \label{fig:coat-overview}
    \end{subfigure}
    
    \caption{Overview examples for Fashion MNIST classes (Part 5).}
\end{figure}
\begin{figure}[H]
    \ContinuedFloat 
    \centering
    \begin{subfigure}[t]{\linewidth}
        \centering
        \includegraphics[width=\linewidth]{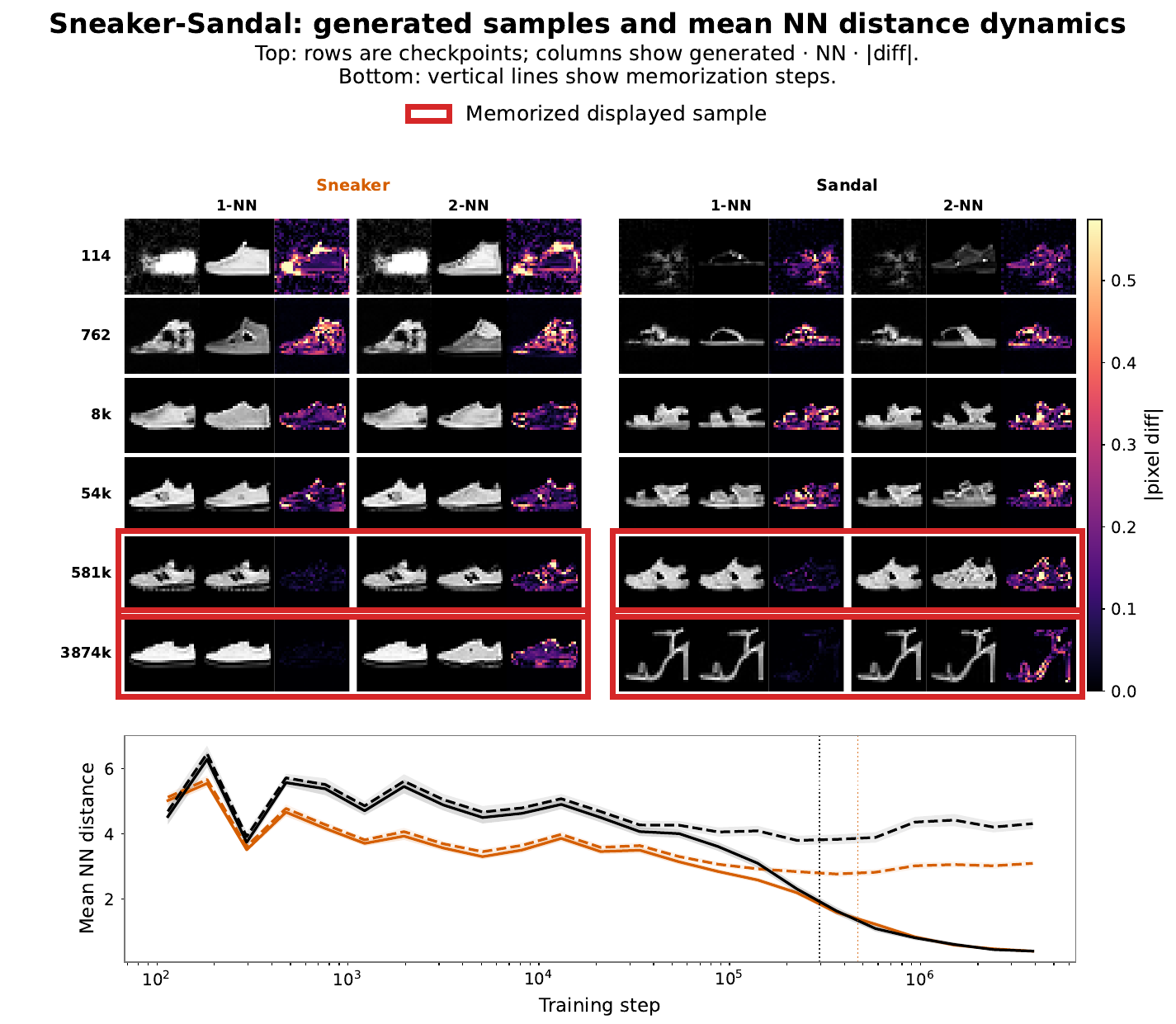}
        \caption{Sandal}
        \label{fig:sandal-overview}
    \end{subfigure}
    
    \caption{Overview examples for Fashion MNIST classes (Part 6).}
\end{figure}
\begin{figure}[H]
    \ContinuedFloat 
    \centering
    \begin{subfigure}[t]{\linewidth}
        \centering
        \includegraphics[width=\linewidth]{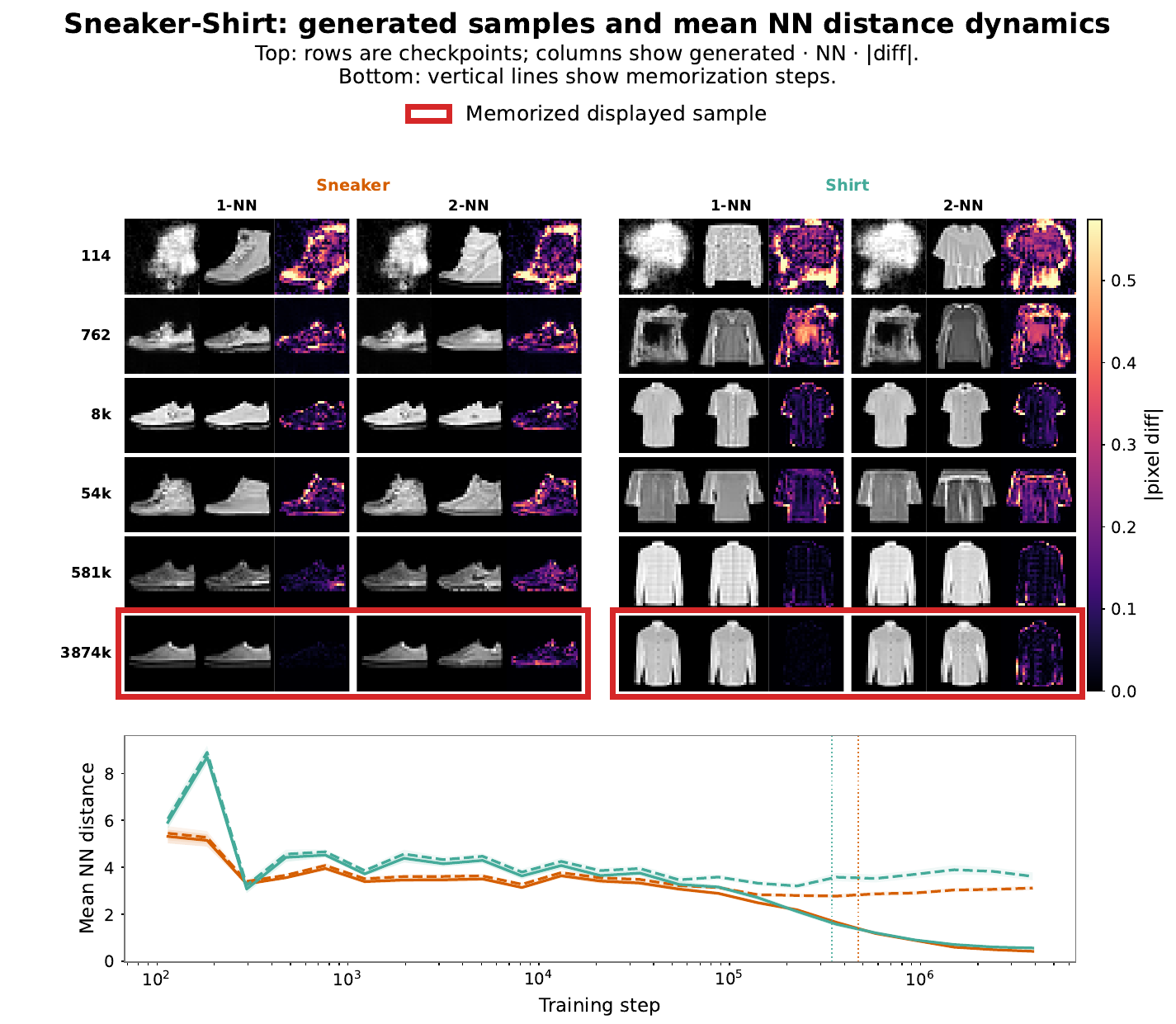}
        \caption{Shirt}
        \label{fig:shirt-overview}
    \end{subfigure}
    
    \caption{Overview examples for Fashion MNIST classes (Part 7).}
\end{figure}
\begin{figure}[H]
    \ContinuedFloat 
    \centering
    \begin{subfigure}[t]{\linewidth}
        \centering
        \includegraphics[width=\linewidth]{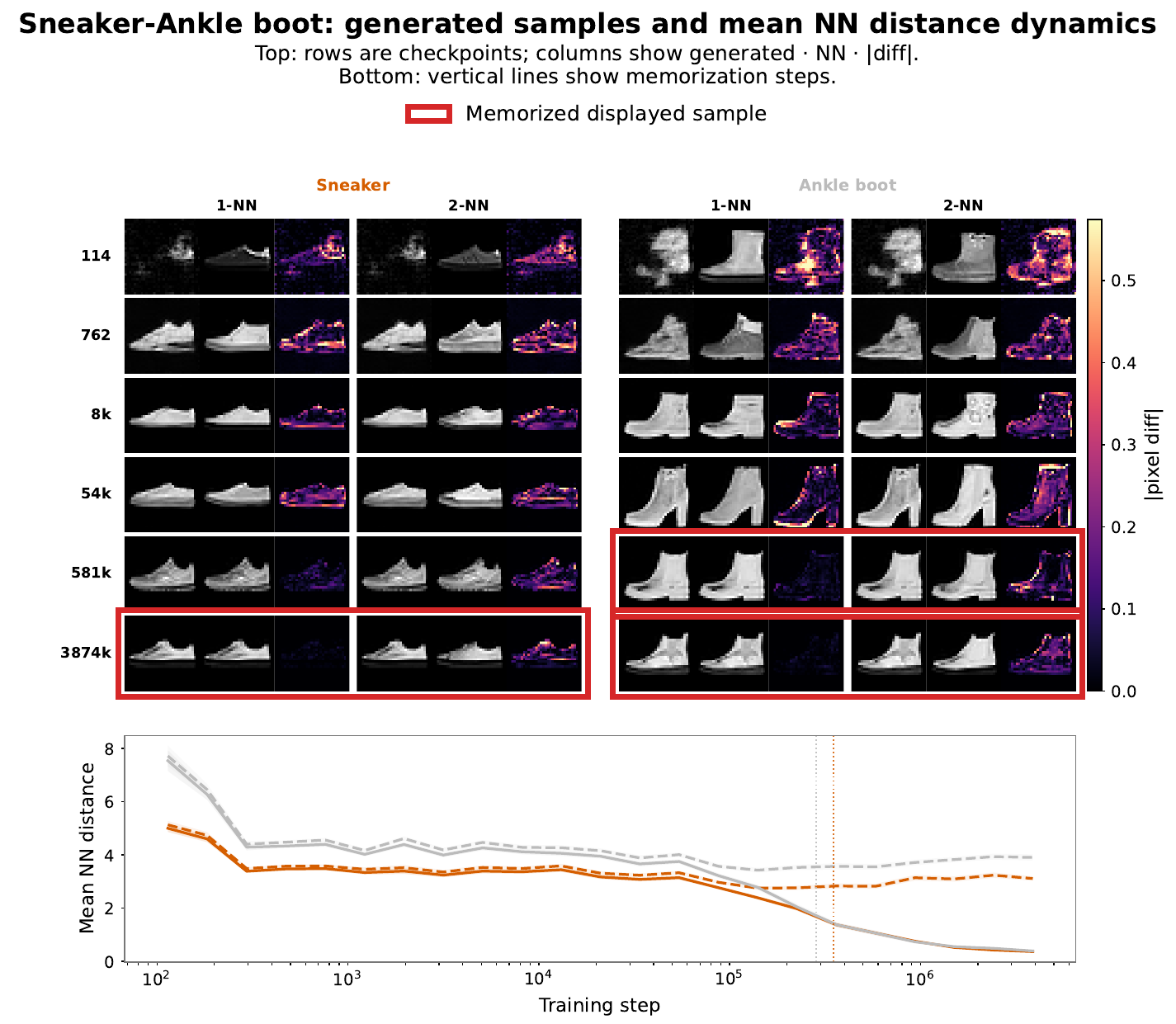}
        \caption{Ankle boot}
        \label{fig:ankle-boot-overview}
    \end{subfigure}
    
    \caption{Overview examples for Fashion MNIST classes (Part 8).}
\end{figure}

\paragraph{Memorization fraction curves.}
We next extend the illustration in Fig.~\ref{fig:Fashion MNIST}(a) by showing the memorization fraction curves for all Sneaker-paired Fashion MNIST classes in Fig.~\ref{fig:fashion-mnist-overview-main}. Including early checkpoints shows that the memorization fraction remains zero for all classes before the onset of memorization. Afterward, all paired classes except Trouser reach the memorization threshold before Sneaker.

\begin{figure}[H]
    \centering
    \includegraphics[width=\linewidth]{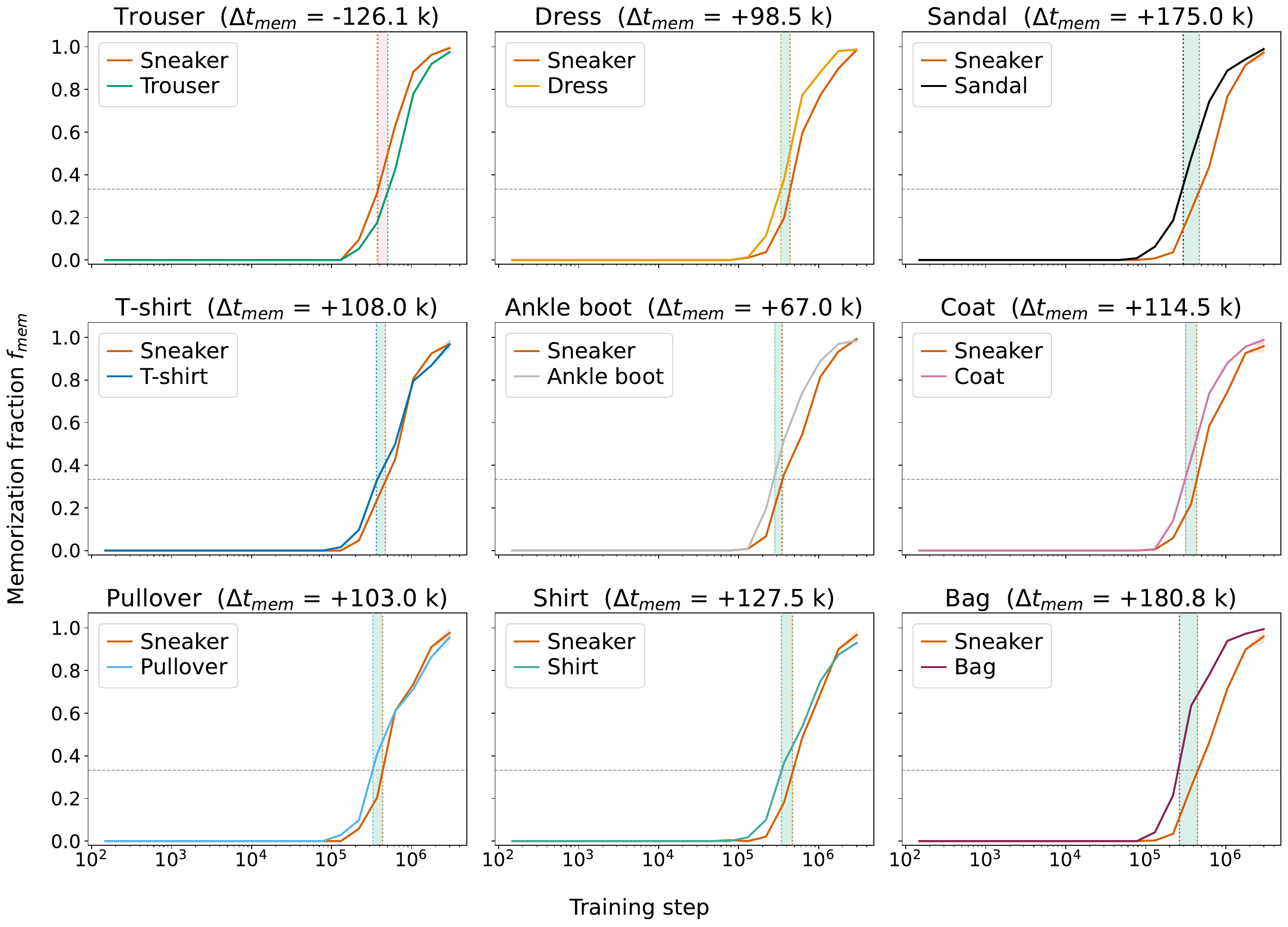}
    \caption{Memorization fraction curves for all Sneaker-paired Fashion MNIST classes.}
    \label{fig:all-fashionmnist-memorizations}
\end{figure}

\begin{figure}[H]
    \centering
    \includegraphics[width=0.95\linewidth]{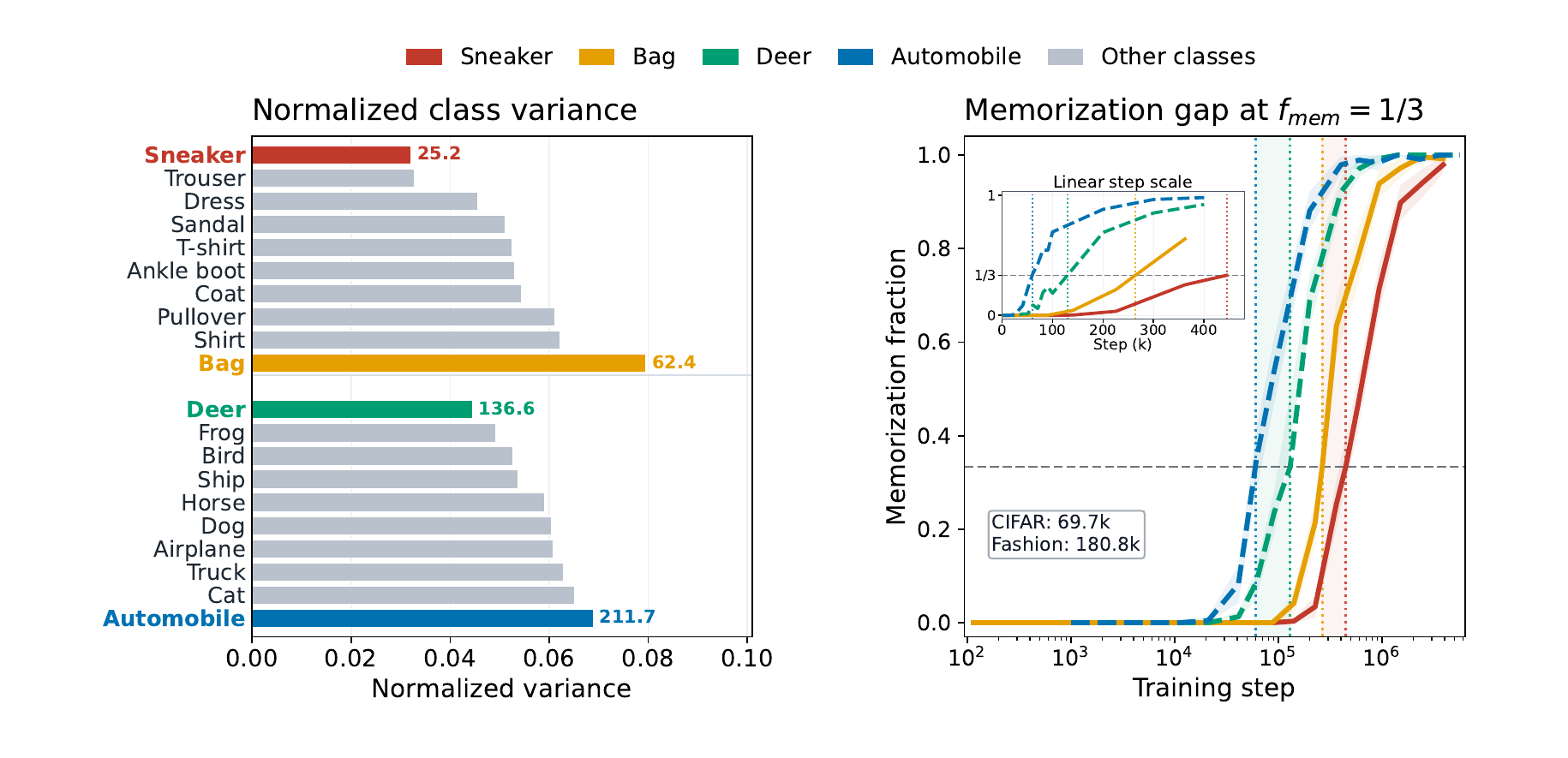}
    \caption{
    Left: normalized variance for each class on two datasets; the annotated numbers report the total class variance.
    Right: memorization fraction curves for the class pair with the largest variance gap in each dataset.
    The inset shows the same curves on a linear training-step scale, and the legend reports the corresponding memorization-time gaps.
    }
    \label{fig:cifar-comparison}
\end{figure}

\begin{figure}[H]
    \centering
    \includegraphics[width=\linewidth]{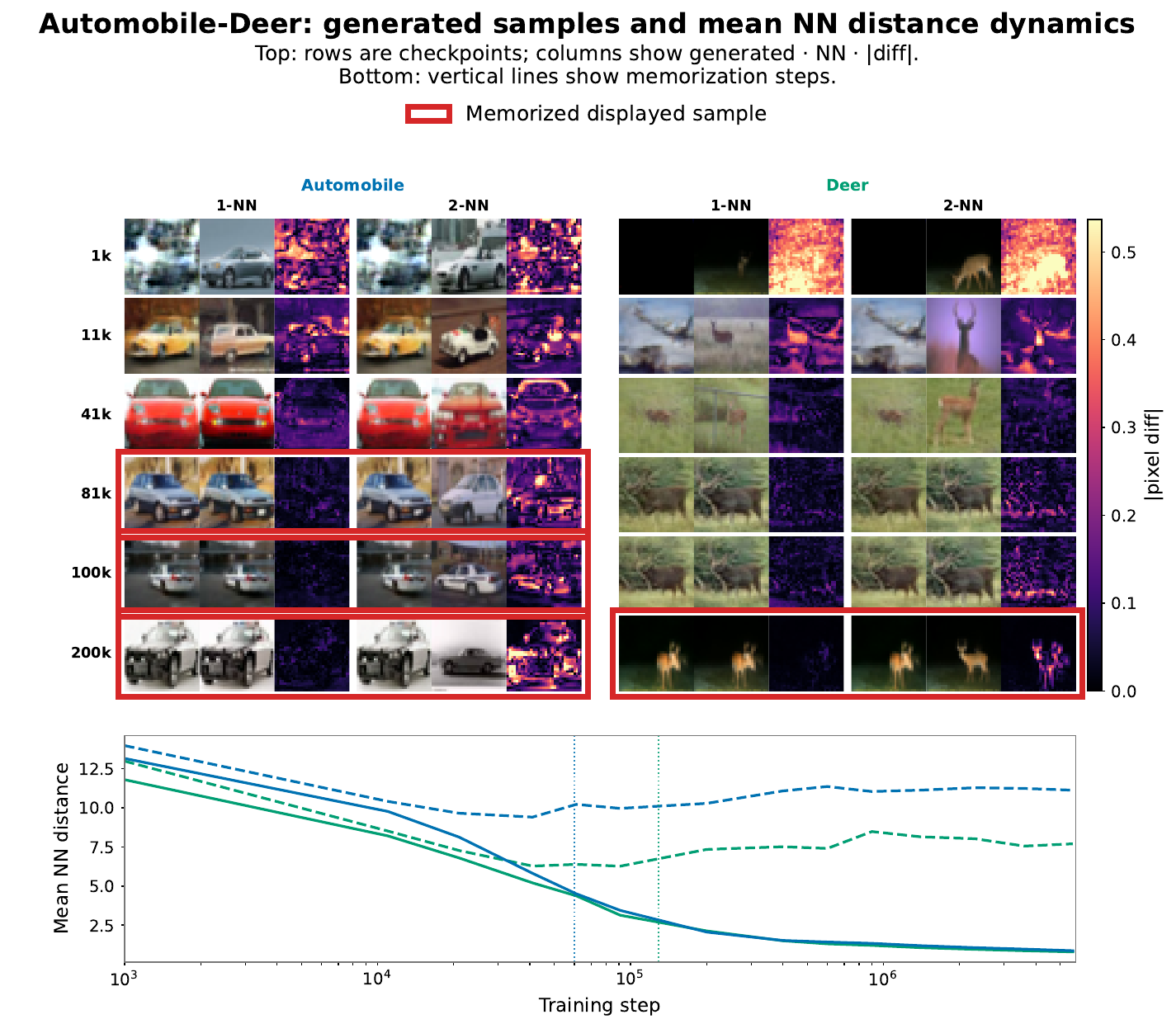}
    \caption{
    CIFAR generated samples and nearest-neighbor dynamics.
    For CIFAR, we restrict the second nearest neighbor to the same class.
    This avoids fluctuations in the average 2-NN distance caused by Automobile samples whose unconstrained second nearest neighbor may lie in Deer class, while preventing such fluctuations from shifting the Automobile memorization time earlier.
    }
    \label{fig:cifar-overview}
\end{figure}

\paragraph{Further validation on CIFAR.}
We further present the same phenomenon on CIFAR \cite{krizhevsky2009learning} using the same DDPM architecture and training settings, with $3000$ training samples. The left panel of Fig.~\ref{fig:cifar-comparison} shows that CIFAR classes have smaller normalized in-class variances than Fashion MNIST classes. Nevertheless, when zooming in on the lowest- and highest-variance CIFAR classes, Automobile is memorized before Deer, matching the variance-based prediction. Absolute memorization times differ from those on Fashion MNIST, which is expected because the CIFAR experiment uses a different total sample size; this is consistent with the theoretical prediction that sample number also affects memorization time~\cite{kadkhodaie_generalization_2023, bonnaire2025diffusion}. The corresponding generated samples and the dynamics of the nearest-neighbor distance are shown in Fig.~\ref{fig:cifar-overview}.

\begin{figure}[t]
    \centering
    \includegraphics[width=\linewidth]{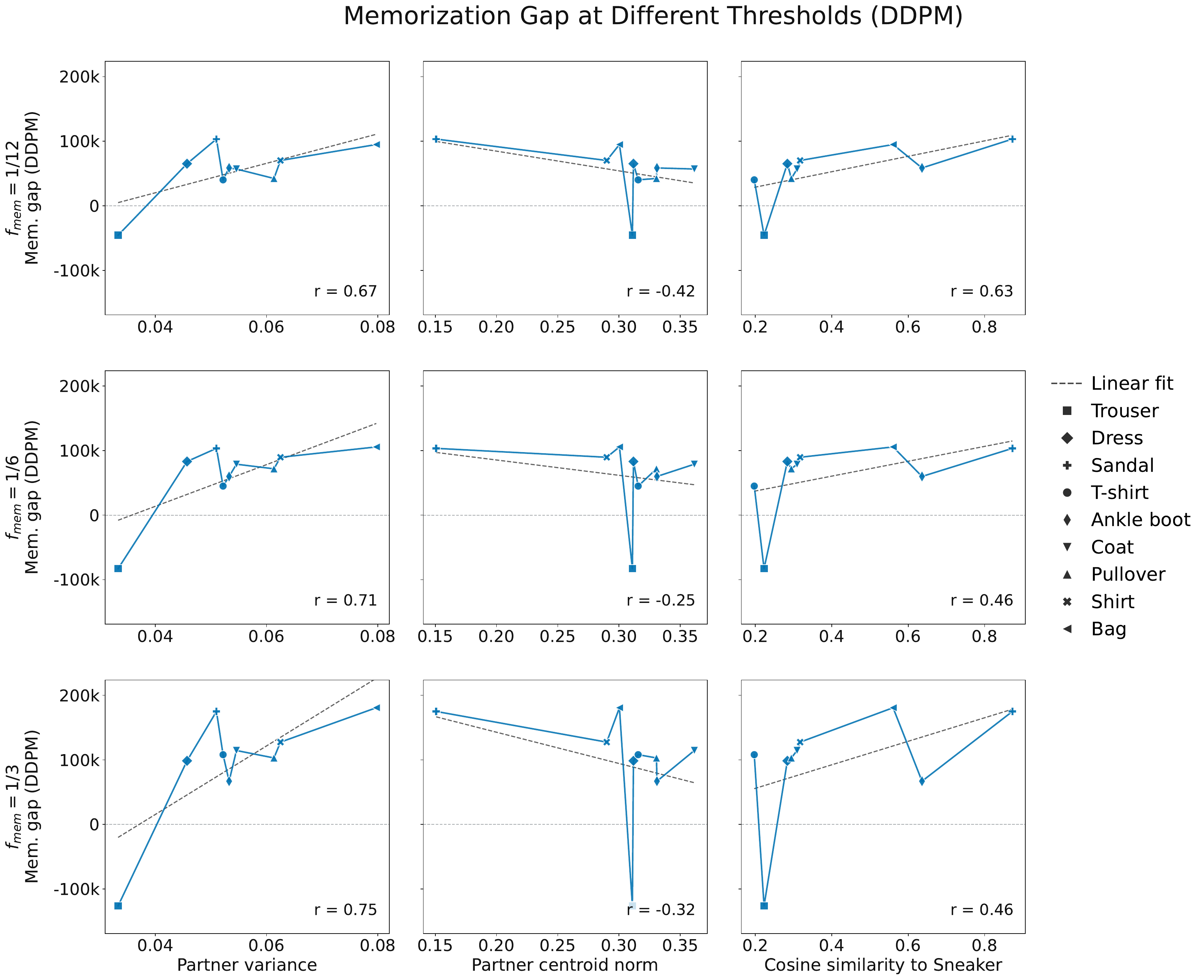}
    \caption{
    Same plot as Fig.~\ref{fig:ddpm-rf}, but using alternative thresholds to compute the DDPM memorization gap $\Delta t_{\mathrm{mem}}$.
    The three rows correspond to $f_{\mathrm{mem}}=1/12$, $1/6$, and $1/3$, with the definition of $\Delta t_{\mathrm{mem}}$ given in Section~\ref{sec:Numerical_Experiments_Fashion_MNIST}.
    The overall patterns are similar across these threshold choices.
    }
    \label{fig:ddpm-thresholds}
\end{figure}

\begin{figure}[t]
    \centering
    \includegraphics[width=\linewidth]{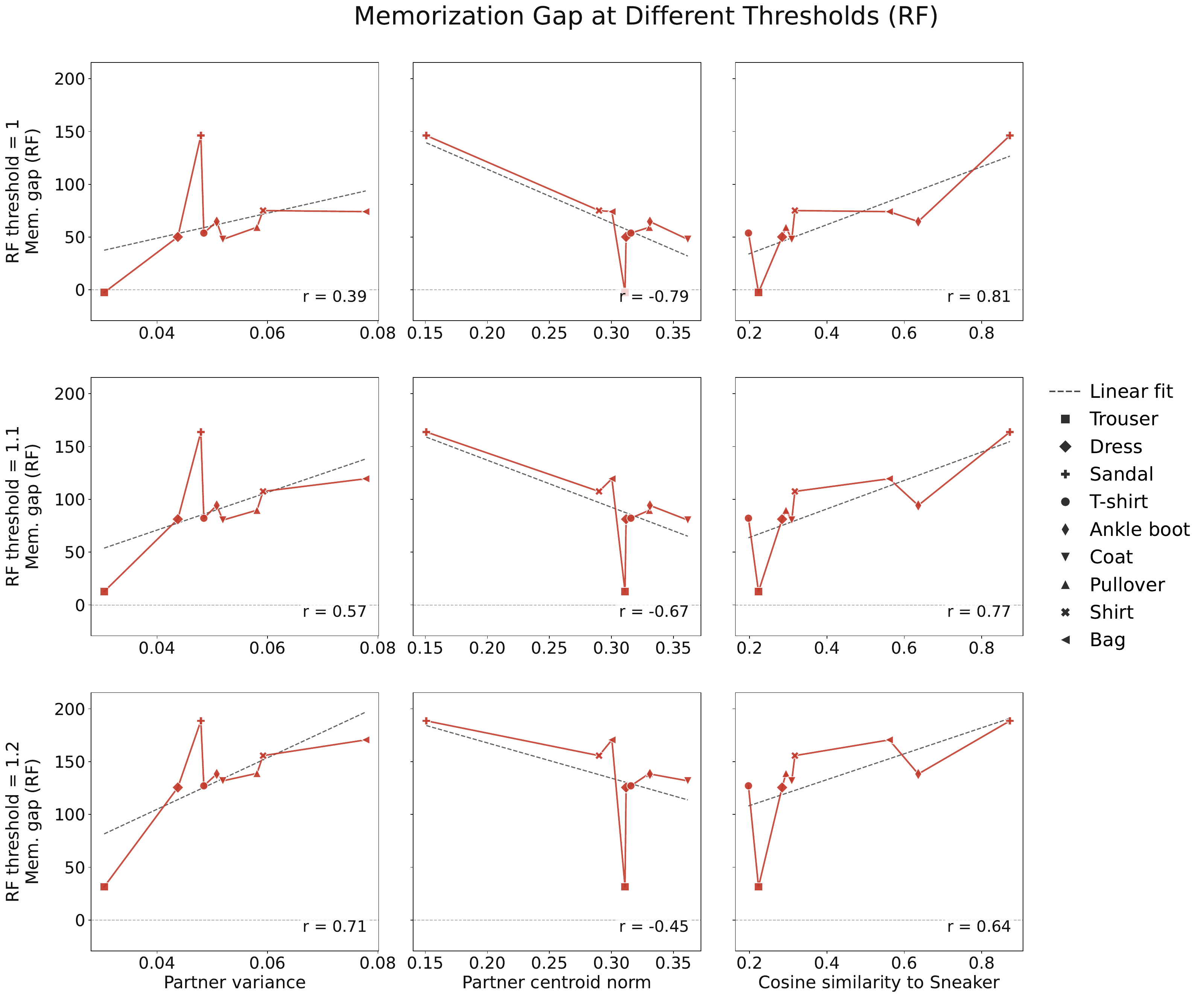}
    \caption{
    Same plot as Fig.~\ref{fig:ddpm-rf}, but using alternative RF memorization thresholds.
    The RF threshold is defined by how much the test loss is allowed to degrade relative to the initial loss, as described in Section~\ref{sec:class_specific_gen_mem}; for example, a threshold of $1.1$ means $10\%$ worse than the initial loss. Similar patterns persist across thresholds.
    }
    \label{fig:rf-thresholds}
\end{figure}

\paragraph{Robustness to memorization thresholds.}
Because memorization has no unique operational definition, choosing comparable thresholds across models is inherently difficult. We use $f_{\mathrm{mem}}=1/3$ for DDPM in the main text to capture a realistic regime where a substantial, but not exhaustive, fraction of samples is memorized. The same trends hold under stricter choices $f_{\mathrm{mem}}=1/6$ and $1/12$. For RF, whose criterion is tied to the initial loss, tightening the threshold from $20\%$ worse than the initial loss to $10\%$ worse, or equal to the initial loss, also preserves the qualitative picture (see Fig.~\ref{fig:ddpm-thresholds} for DDPM and Fig.~\ref{fig:rf-thresholds} for RF, respectively). Moreover, using RF threshold $1.3$ gives a correlation with the DDPM result comparable to that obtained with the main threshold $1.2$.

\end{document}